%% file: neurips_2026.tex
\documentclass{article}

\input{preamble}

\title{\shortname: Generative Plannable 4D Worlds}

\author{%
\makebox[\dimexpr\textwidth-2\tabcolsep\relax][c]{%
Hangwei Zhang\textsuperscript{1,2} \quad
Rujing Dang\textsuperscript{2} \quad
Ruimin Lin\textsuperscript{2}} \\ [0.5em]
\makebox[\dimexpr\textwidth-2\tabcolsep\relax][c]{\bfseries%
Jiahong Wu\textsuperscript{2,\dag} \quad
Xiangxiang Chu\textsuperscript{2} \quad
Chuanxia Zheng\textsuperscript{1,\dag}} \\ [1em]
\textsuperscript{1} PVG, Nanyang Technological University
\quad
\textsuperscript{2} AMAP, Alibaba Group
}

\begin{document}

\maketitle

\begingroup
\renewcommand{\thefootnote}{}
\footnotetext{\textsuperscript{\dag}Corresponding authors.}
\endgroup

\vspace{-2em}  
\begin{figure}[h]
    \centering
    \includegraphics[width=\linewidth, trim={19.5 268.4 75.3 17.7}, clip]{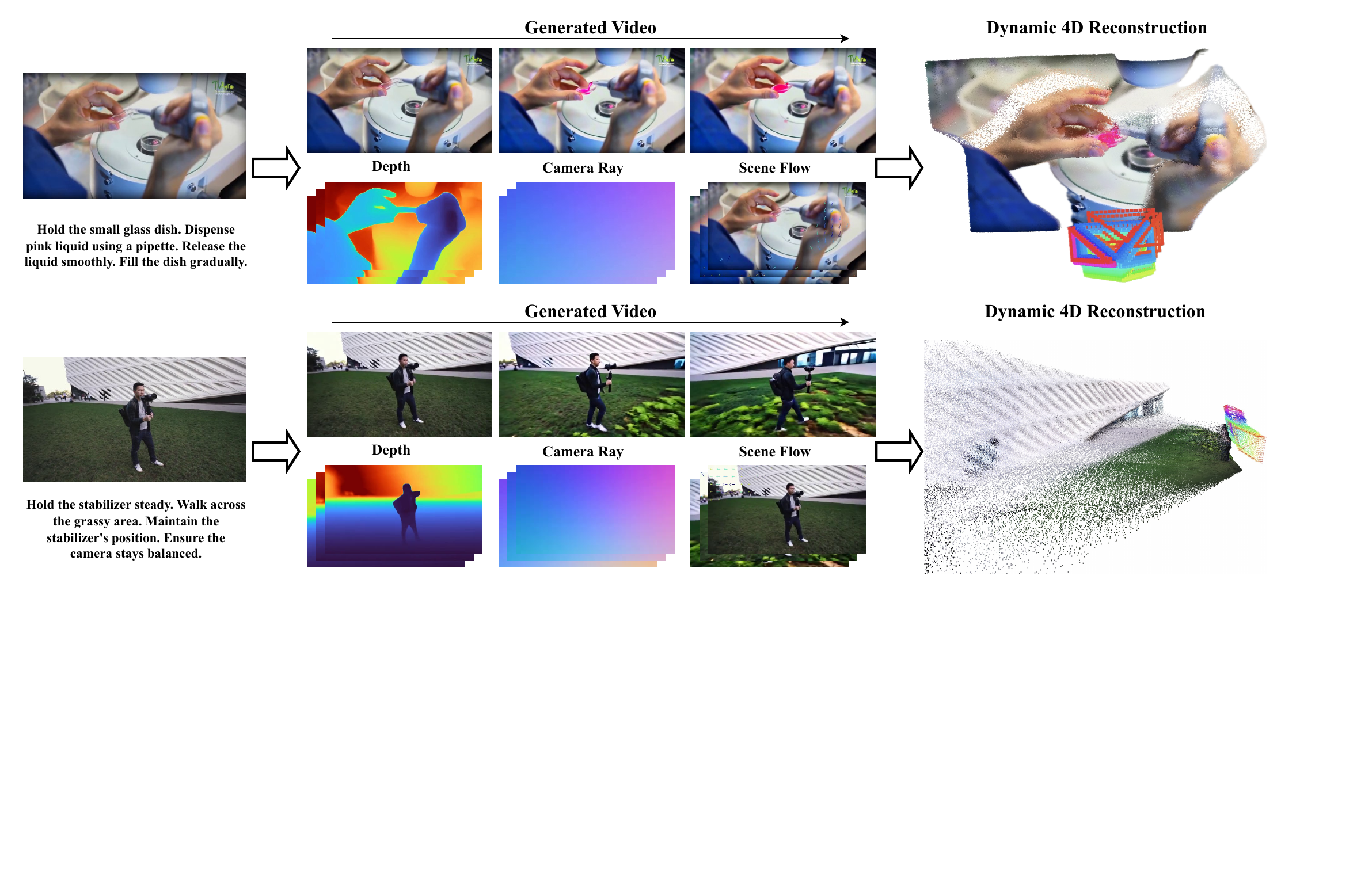}
    \vspace{-0.5em}  
    \caption{
        \textbf{Exampled future 4D world generated by \shortname.}
        Given an observed scene context and a text-specified action, \shortname generates a future video together with synchronized depth, camera rays, and scene flow (subset shown), which are further lifted into a dynamic 4D scene.
        Top: fine-grained hand-object interaction.
        Bottom: camera motion with scene-level geometry changes.
  }
    \label{fig:teaser}
\end{figure}
\vspace{-6pt}  

\input{sections/0_abstract}
\input{sections/1_intro}

\input{sections/2_related_works}
\input{sections/3_methodology}
\input{sections/4_experiments}
\input{sections/5_conclusion}

\bibliographystyle{plainnat}
\bibliography{main,chuanxia_general,chuanxia_specific}

\input{sections/appendix}

\newpage
\input{checklist.tex}

\end{document}

%% file: preamble.tex
\PassOptionsToPackage{numbers, compress}{natbib}
\PassOptionsToPackage{table}{xcolor}
\usepackage{comment}
\usepackage{multirow}
\usepackage{comment}
\usepackage{natbib}
\usepackage{caption}
\usepackage{wrapfig}

\usepackage[main, final]{neurips_2026}

\usepackage[utf8]{inputenc}
\usepackage[T1]{fontenc}
\usepackage{hyperref}
\usepackage{url}
\usepackage{graphicx}
\usepackage{booktabs}
\usepackage{amsmath}
\usepackage{amsfonts}
\usepackage{nicefrac}
\usepackage{microtype}
\usepackage{xcolor}
\IfFileExists{axessibility.sty}{\usepackage[accsupp]{axessibility}}{}
\usepackage{xspace}
\usepackage{multirow}
\usepackage{cleveref}
\usepackage{bm}

\newcommand{\shortname}{\textsc{Plan4D}\xspace}

\newcommand{\cxz}[1]{{\color{purple}[cxz: #1]}}

\newcommand{\heading}[1]{\vspace{2pt}\noindent\textbf{#1.}}

\definecolor{bestC}{HTML}{EAB6B6}
\definecolor{secondC}{HTML}{F4D6B4}
\definecolor{thirdC}{HTML}{FCF5C7}
\newcommand{\fst}[1]{\cellcolor{bestC}\textbf{#1}}
\newcommand{\snd}[1]{\cellcolor{secondC}#1}
\newcommand{\trd}[1]{\cellcolor{thirdC}#1}

%% file: sections/0_abstract.tex
\begin{abstract}

Most world models for future prediction prioritize the aesthetic quality of generated videos,
but offer only limited geometric consistency over time,
due to the lack of explicit 4D modeling.
In this paper,
we introduce \shortname,
a possible next milestone towards building a \emph{generative plannable 4D world model} that predicts future RGB videos together with dense 4D geometry,
including depth, cameras, point clouds, and scene flow,
from an observed scene context and text-specified actions.
Since training data with paired actions and dense 4D annotations is inherently scarce,
\shortname leverages the complementary priors of foundational video generators and 3D reconstructors.
Specifically,
we propose a model-stitching architecture that couples Depth Anything 3 with LongCat-Video and augments it with an auxiliary DPT head for scene-flow prediction.
The resulting model allows us to plan and generate complex, long-horizon, and fine-grained actions in dynamic 3D scenes,
while substantially improving both visual fidelity and geometric accuracy of the generated future videos.
Extensive experiments on both synthetic and real-world datasets demonstrate that \shortname achieves state-of-the-art performance in generative 4D modeling of environmental reactions.
Project Page: \url{https://fogradio.github.io/plan4d-project}.

\end{abstract}

%% file: sections/1_intro.tex
\section{Introduction}
\label{sec:intro}

Given an observed scene context and a set of text-specified actions,
we consider the task of \emph{generative plannable 4D worlds},
a new paradigm for predicting reliable futures for the physical world that jointly produces visually plausible RGB frames and physically grounded 4D geometry --including depth, camera dynamics, and scene flow --within a single feed-forward model.
Prior \emph{observation-prediction models} formulate future prediction largely as RGB synthesis~\citep{team2026dreamxworld,team2026dreamx}, often on a per-frame basis,
notably UniSim~\citep{yang2024learning}, Genie 1--3~\cite{bruce2024genie,parkerholder2024genie2,genie3}, NWM~\cite{bar2025navigation}, and V-JEPA 2~\citep{assran2025v}.
By contrast,
with our actionable 4D future world model,
humans and AI agents can plan and execute complex, long-horizon, and fine-grained actions in the explicit 3D environment,
such as picking up a cup, opening a bottle, or navigating through a cluttered room.
These dynamic 3D quantities are essential for accurate measurement, reliable simulation, and effective execution of downstream tasks~\citep{krishnan2025orchid,dharmarajan2025dream2flow, li2025unified,huang2026pointworld, gu2026right}.


A major bottleneck of this task is the lack of suitable datasets,
namely action-conditioned video sequences paired with rich text and \emph{comprehensive} 3D annotations,
including camera trajectories, accurate dense depth and motion.
For example,
large-scale video-caption datasets~\cite{bain2021frozen,wang2023internvid,chen2024panda} may still lack detailed captions for physical fine-grained actions;
the latest Action-100M~\cite{chen2026action100m} dataset provides detailed action annotations but without 3D annotations,
while most dynamic 3D datasets~\cite{greff2021kubric,grauman2024ego,kastingschafer2025seed4d} are relatively small in scale or lack natural actions.

Correspondingly,
powerful baselines have emerged in parallel on these separate data sources,
including video generators~\cite{ho2022video,blattmann2023align,blattmann2023stable,singer2023makeavideo,guo2024animatediff,kong2024hunyuanvideo,yang2024cogvideox,wan2025,liu2025infinitystar,team2025longcat},
action-conditioned world models~\cite{yang2024learning,bruce2024genie,parkerholder2024genie2,genie3,bar2025navigation,assran2025v,kanervisto2025world,li2025vmem},
and 3D reconstructors~\cite{wang2024dust3r,leroy2024grounding,wang2025continuous,wang2025vggt,keetha2025mapanything,wang2025pi,lin2025depth,zhang2025bokehdepth}.
Despite being trained to build the same physical world,
these models still operate as isolated systems for particular tasks.
Standard pixel-level generators excel at synthesizing visually realistic RGB frames~\citep{wu2026latent},
but typically lack physical consistency~\citep{wu2026omniworldbench,li2026phyparam};
and 3D models recover explicit geometry well~\citep{wang2025editor} but are largely reconstruction-centric and static.
Even advanced 4D frameworks~\cite{zhang24monst3r,jiang2025geo4d,sucar2026v,zhu2026motioncrafter,karhade2025any4d,luo20264rc,lu2026track4world,lan2026track4d} only infer motion from \emph{observed} frames and \emph{cannot generate action-conditioned future worlds}.
As a result, none of these families alone provides a \emph{generative plannable 4D world},
in which appearance, geometry, camera and action-conditioned future dynamics evolve consistently over time.

What if,
given these different dataset sources,
we stitch them together to model the same underlying 4D world?
To achieve such a \emph{generative plannable 4D world},
we propose \textbf{\shortname},
a unified framework for \emph{planning} and \emph{generating} dense 4D worlds from observed context and action descriptions.
Our \emph{core insight} is to decouple the acquisition of strong geometric priors from the generation of future dynamics,
then reconnect them through a stitched latent world model that jointly plans RGB appearance and dense 4D geometry,
unlike recent methods~\citep{zhu2025aether,zhen2025tesseract,dai2025fantasyworld,lu2026phys4d,fang2025worldreel} that attempt to learn a 4D world model's geometry from scratch.
Such a paradigm reduces the task's learning difficulty and data requirements,
while allowing us to leverage the powerful priors of pretrained video generators and 3D reconstructors to achieve complex, long-horizon and fine-grained dynamics.

In practice,
this paradigm requires solving several coupled challenges.
i) the model must extend a frozen 3D foundation model from static scenes to dynamic 4D prediction while preserving its geometric consistency.
ii) it must align geometry tokens with the latent space of a video generator without destroying the pretrained generative prior.
iii) it must learn from real, unlabeled action videos without dense 4D supervision~\citep{yuan2025self}.
iv) it must also ensure consistency across the generated RGB, depth, camera, point clouds, and scene flow.
Recent methods, including VIST3A~\citep{go2025vist3a}, Geometry Forcing~\citep{wu2025geometry}, and Gen3R~\citep{huang2026gen3r}, couple 3D reconstructors with video generators for static text-to-3D or scene-level 3D generation.
We extend this direction to \emph{full actionable 4D world generation}, jointly predicting full 4D geometry and appearance in future video sequences depicting fine-grained actions.

Unlike prior approaches that use geometry mainly as an auxiliary constraint~\citep{wu2025geometry,bai2025geovideo,yin2026vigor, du2026videogpa,an2026vggrpo},
or decode geometry only after RGB generation~\citep{dharmarajan2025dream2flow}, 
our \shortname turns a video generator into an actionable 4D world model,
which \emph{explicitly aligns future appearance and 4D structure inside the shared evolving video latent state.}
To preserve strong geometric priors while extending them beyond reconstruction, we augment Depth Anything 3~\citep{lin2025depth} (DA3) with a lightweight token-only scene-flow head that predicts dense forward 3D motion from intermediate geometry tokens,
while keeping its pretrained depth and camera-ray predictions intact.
To connect this geometric readout with future generation, we learn a \emph{latent-to-geometry stitching module} that maps video VAE latents into a selected DA3 block and trains it with token distillation and geometric supervision.
The resulting stitched decoder provides a stable, differentiable interface from video latents to full 4D geometry.
We then train the video generator against this frozen geometric interface, so unlabeled action videos can supervise not only visual prediction but also physically grounded 4D consistency.
At inference time, a single generated latent future yields RGB frames and dense 4D geometry in one feed-forward pass.

Our main contributions can be summarized as follows:
\vspace{-5pt}
\begin{itemize}
    \item We take a step towards the new paradigm of \emph{generative plannable 4D worlds}, which jointly generates future RGB appearance and corresponding full 4D geometry, including depth, camera rays, point clouds, and scene flow, from observed context and action descriptions.
    \item We propose \shortname, a unified framework with a scene-flow head and a geometry-video stitching mechanism that connect a frozen 3D foundation model with a pretrained video generator, preserving geometric priors while enabling action-conditioned dynamic 4D future prediction from unlabeled videos.
    \item We illustrate how \shortname can achieve state-of-the-art performance in dense 4D world generation, improving both visual realism and geometric accuracy across scene flow, 3D reconstruction, and action-conditioned video prediction through extensive experiments.
\end{itemize}
\vspace{-5pt}

%% file: sections/2_related_works.tex
\section{Related Works}
\label{sec:related_works}

\noindent\textbf{4D world models.}\label{sec:rel:4dwm}
Recent work has begun to model 4D worlds from two complementary directions.
Explicit dynamics methods such as ParticleFormer~\citep{huang2025particleformer} and PointWorld~\citep{huang2026pointworld} predict particle trajectories or point-level motion, but mainly operate in structured 3D state spaces rather than generating rich future 4D.
4D-native generative models instead learn spatiotemporal world evolution, ranging from dynamic reconstruction, visual planning, and image-to-4D generation in Aether~\citep{zhu2025aether} and 4DNeX~\citep{chen20254dnex}, to RGB-D or geometry-aware scene generation in TesserAct~\citep{zhen2025tesseract} and Robo4DGen~\citep{liu2025geometry}. GWM~\citep{lu2025gwm}, iMoWM~\citep{zhang2025imowm}, and Kinema4D~\citep{xu2026kinema4d} further explore Gaussian, tokenized, and kinematics-driven representations, while FantasyWorld~\citep{dai2025fantasyworld}, WorldReel~\citep{fang2025worldreel}, TeleWorld~\citep{chen2025teleworld}, HY-World 2.0~\citep{hy2026hy}, and Lyra 2.0~\citep{shen2026lyra} couple video generation with geometric prediction for more unified world modeling.
Despite this progress, existing systems typically emphasize either explicit geometric forecasting or large-scale generative simulation, and they rarely integrate the complementary priors of pretrained geometry foundation models into the latent dynamics of video generation.

\noindent\textbf{Aligning 3D Reconstruction and Video Generation.}\label{sec:rel:align}
Early uses of generative priors for 3D synthesis mostly follow a cascade, where geometric constraints are applied to generated outputs rather than to internal generative states~\citep{poole2022dreamfusion, gao2024cat3d, wang2026edit}. Later methods tighten this link through reconstruction-guided diffusion or distillation, as in GenFusion~\citep{wu2025genfusion}, VideoScene~\citep{wang2025videoscene}, and Lyra~\citep{bahmani2025lyra}. Recent work moves the alignment into feature or latent spaces. Geometry Forcing~\citep{wu2025geometry} aligns intermediate diffusion features with pretrained geometric representations~\citep{yu2024representation}, Gen3R~\citep{huang2026gen3r} aligns geometry and appearance latents for joint generation, and GLD~\citep{jang2026repurposing} repurposes geometric features as multi-view diffusion latents. From the model-stitching view~\citep{lenc2015understanding, bansal2021revisiting}, VIST3A~\citep{go2025vist3a} connects a pretrained video VAE with a 3D network at linearly matchable layers. Our method builds on this direction, but stitches a video diffusion model to a 3D foundation model so that RGB synthesis and dense 4D geometry can be optimized and decoded from a shared latent rollout.

%% file: sections/3_methodology.tex
\section{Plan4D}
\label{sec:methodology}

Given an observed context
$\mathcal{O}=\{\bm{I}_i\}_{i=1}^M$
and the planned action space
$\mathcal{A}$,
our goal is to learn a generative plannable 4D world model
$f_{\bm{\theta}}$
that is conditioned on $\mathcal{O}$ and $\mathcal{A}$ and predicts a set of future 4D representations:
\(
    \label{eq:problem}
    f_{\bm{\theta}}: \{\mathcal{O}, \mathcal{A}\} \mapsto \{\bm{I}_t, \bm{D}_t, \bm{r}_t, \bm{V}_t\}_{t=1}^T,
\)
where $\mathcal{A}$ is a sequence of textual action descriptions $\bm{p}$.
The model $f_{\bm{\theta}}$
learns a video latent trajectory that is conditioned on the observed context and the action description,
and reads out the future RGB video frames $\bm{I}_t \in \mathbb{R}^{H \times W \times 3}$;
and its corresponding depth maps $\bm{D}_t \in \mathbb{R}^{H \times W}$, camera ray maps $\bm{r}_t \in \mathbb{R}^{H \times W \times 6}$, and full 4D scene flow $\bm{V}_t \in \mathbb{R}^{H \times W \times 3}$.
The ray maps $\bm{r}_t$ encode the camera origin and the ray directions for each pixel,
which can be used to reconstruct the 3D point cloud of the scene.
The scene flow $\bm{V}_t$ encodes the 3D motion of each pixel from time $t$ to $t+1$\footnote{Since we only define the forward flow, we then simplify $\bm{V}_{t\to t+1}$ as $\bm{V}_t$.}.

We parametrize $f_{\bm{\theta}}$ with a model-stitching architecture that couples a video generator, LongCat-Video~\cite{team2025longcat} (which uses the Wan 2.1 VAE~\cite{wan2025}), with a foundational 3D reconstructor, Depth Anything 3~\cite{lin2025depth} (DA3).
The \emph{key insight} is that,
by building on the pretrained video generator and 3D reconstructor,
our model can leverage the complementary priors of these two foundational models to achieve both high visual fidelity and strong geometric consistency in the generated future videos,
unlike training it from scratch~\cite{zhu2025aether,zhen2025tesseract,dai2025fantasyworld,fang2025worldreel},
which is extremely challenging and data-hungry.
To improve the motion modeling capability of DA3,
we first augment it with an auxiliary DPT-style scene flow head in~\cref{sec:method_flow}.
In~\cref{sec:method_stitch},
we then stitch the video generator and the augmented DA3 together through a learnable latent-to-geometry stitching module,
followed by a distillation and geometric supervision process to calibrate the stitched model in~\cref{sec:method_finetuning}.
Finally, we train the joint model in~\cref{sec:method_joint}.

\begin{figure*}[tb!]
\centering
\includegraphics[width=\textwidth, trim={25.0 5.1 43.3 5.2}, clip]{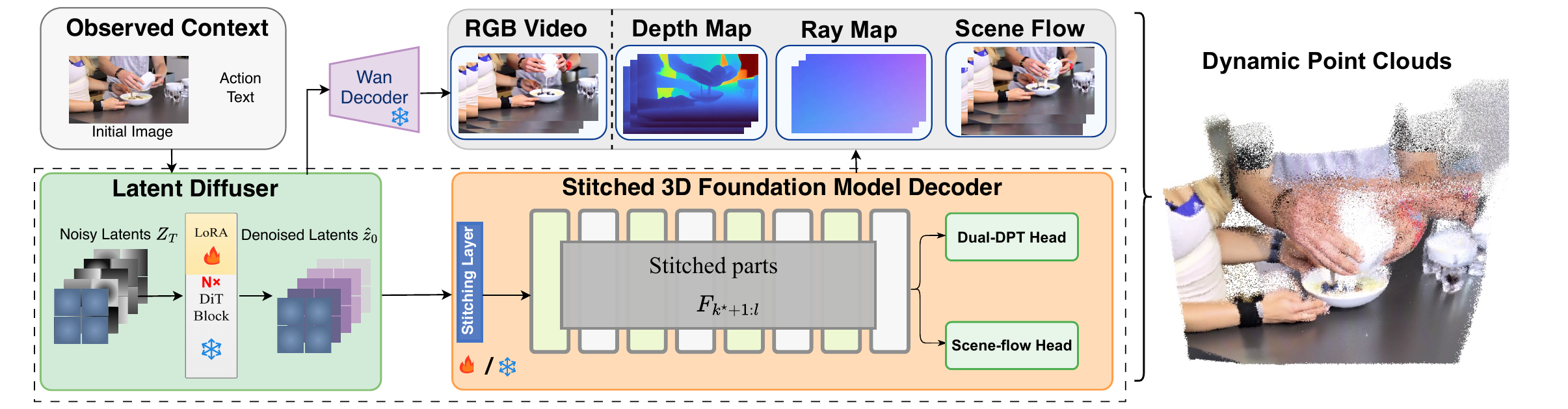}
\caption{\textbf{Overview of the \shortname pipeline.}
A video generator produces a denoised latent trajectory, while a stitched DA3 decoder reads the same latent state to recover synchronized depth, camera rays, camera parameters, point clouds, and scene flow.
During joint training, the stitched decoder is frozen and acts as a differentiable geometry critic for the video generator; at inference time, RGB video and dense 4D structure are decoded from the shared latent trajectory.}
\label{fig:plan4d-pipeline}
\end{figure*}

\subsection{Augmenting DA3 for Dynamic Scene Flow Prediction}
\label{sec:method_flow}

The physical world is \emph{not} a collection of independent per-frame geometries.
To model the dynamics and correspondence of the 4D world,
we augment DA3~\citep{lin2025depth} with a dense \emph{scene flow} head that predicts the 3D motion of each pixel.
Following the original DPT-style decoder in DA3,
the new head ingests the four intermediate token maps of the frozen encoder,
and outputs a dense three-channel motion field via a lightweight prediction layer.
Note that,
the original encoder, as well as the depth and camera pose branches, remains unchanged; we simply add a dedicated motion branch.
Our analysis in~\Cref{sec:appendix_frozen_da3_analysis} shows that frozen DA3 tokens already encode cross-frame motion cues, which supports using this branch as a lightweight readout rather than a full motion model.
In particular,
for each frame $t$, the new head predicts a scene flow on the pixel lattice of that frame.
Each valid pixel stores the forward 3D displacement from time $t$ to $t+1$. The raw prediction is produced in the local camera basis of frame $t$ and is mapped to a common reference frame with a lightweight adapter,
\begin{equation}
\label{eq:flow}
\bm{V}^{(0)}_t(u,v) = \mathbf{R}_{t \rightarrow 0}\,\bm{V}^{(\mathrm{cam})}_t(u,v),
\end{equation}
where $\mathbf{R}_{t \rightarrow 0}$ denotes the rotation from the camera basis of frame $t$ to the first-frame camera basis.
We apply only rotation because scene flow is a vector quantity rather than a 3D point.
This representation keeps spatial indexing local to each frame while expressing all motion vectors in a temporally stable basis.
As a result, the predicted dynamics remain consistent under camera motion and align naturally with the depth, ray, and point-based geometry produced by DA3.
Despite being a seemingly minor change,
the new head has a substantial effect in practice:
it consistently improves reconstruction, camera, and depth estimation quality on various dynamic scenes compared with stitched variants without motion modeling,
as shown in our experiments.

\subsection{Latent-to-Geometry Model Stitching}
\label{sec:method_stitch}

Following recent works in model stitching~\citep{pan2023stitchable,go2025vist3a,huang2026gen3r},
we design a simple learnable module to stitch the video generator and the augmented DA3 together,
allowing the video latent to be directly decoded into full 4D geometry,
including depth, camera poses, point clouds, and scene flow.
Let $\bm{I}^{1:T}$ be a video clip and let
\(
\bm{z}_0^{1:T} = \mathcal{E}_{\mathrm{vae}}(\bm{I}^{1:T})
\)
be its Wan 2.1 VAE latent~\citep{wan2025}.
We first upsample the latent to the spatial and temporal resolution required by DA3:
$\tilde{\bm{z}}_0^{1:T} = \operatorname{Up}_{(s_t,s_h,s_w)}(\bm{z}_0^{1:T})
\in \mathbb{R}^{(s_t t)\times (s_h h)\times (s_w w)\times C}.$
A learnable 3D stitching module $\mathcal{S}_{\bm{\theta}}$ then maps the upsampled latent tensor into the channel dimension of a DA3 encoder block,
$\mathbf{h}^{(k)} = \mathcal{S}_{\bm{\theta}}\bigl(\tilde{\bm{z}}_0^{1:T}\bigr).$
In our implementation $\mathcal{S}_{\bm{\theta}}$ is a compact spatiotemporal projection, instantiated as a 3D convolution. Its temporal kernel aggregates neighboring latent frames, and its spatial stride matches the VAE grid to the DA3 patch grid. 

The injection layer is not chosen heuristically. For a set of paired clips, we extract Wan 2.1 VAE latents and frozen DA3 candidate features from several early encoder blocks. For each candidate block $k$, we solve a regularized linear fitting problem,
\begin{equation}
\label{equ:linear_fitting}
(k^\star,\bm{\theta}_0)
= \arg\min_{k,\bm{\theta}}
\sum_i
\left\|
\mathcal{S}_{\bm{\theta}}\bigl(\tilde{\bm{z}}_0^{1:T}\bigr)
- \mathbf{g}^{(k)}_i
\right\|_2^2
+ \lambda \|\bm{\theta}\|_2^2 ,
\end{equation}
where $\mathbf{g}^{(k)}_i$ is the frozen DA3 feature grid at block $k$ for clip $i$. This procedure selects the block whose representation is most linearly reachable from the video VAE latent~\citep{go2025vist3a, an2026vggrpo}.
It also provides a stable initialization $\bm{\theta}_0$ for the stitching module. The selected block acts as the interface between the generative video prior and the geometric reconstruction prior.

After stitching, the original RGB patch embedding and the encoder blocks before $k^\star$ in DA3 are bypassed.
The projected latent tokens are combined with DA3 special tokens and positional encodings, then processed by the remaining DA3 encoder blocks and the native dense heads,
\(
\hat{\mathcal{G}}
=
\mathcal{H}^{\mathrm{DA3}}_{k^\star:L}
\left(
\mathbf{h}^{(k^\star)}
\right).
\)
The output $\hat{\mathcal{G}}$ contains depth, camera rays, confidence maps, and scene flow. This design is important because the stitched decoder reads the denoised video latent directly.
It does \emph{not} need to first decode the latent into RGB and then run a separate geometry model.
Visual RGB generation and 3D reconstruction are instead two synchronized readouts of the same latent video state.

\subsection{Stitched Decoder Finetuning}
\label{sec:method_finetuning}

Here,
we describe how to finetune the stitched decoder to adapt to the new latent input distribution and obtain accurate geometry predictions.
Unlike VIST3A~\citep{go2025vist3a} which only distills RGB features,
we employ a hybrid training strategy that combines teacher-student distillation with ground-truth (GT) supervision.
The teacher-student distillation follows the same principle as VIST3A,
where the teacher is the original frozen DA3 model augmented with the new scene flow head.
This teacher model takes the ground-truth RGB video clip as input and produces all types of 4D geometric outputs as supervision signals.
The student is the stitched decoder, which takes the Wan VAE latent of the same clip as input and produces the same set of geometric outputs.
For the GT supervision, we utilize the 3D annotations provided in the dataset as explicit regression targets.

Following the formulation in \citet{lin2025depth}, our training objective integrates dense geometric matching with spatial and temporal regularization constraints,
\begin{equation}
\label{equ:l_geo}
\mathcal{L}_{\text{geo}}
=
\lambda_D \mathcal{L}_\text{depth}
+ \lambda_R \mathcal{L}_\text{ray}
+ \lambda_C \mathcal{L}_\text{camera}
+ \lambda_M \mathcal{L}_\text{motion}
+ \lambda_T \mathcal{L}_\text{temporal}.
\end{equation}
We also apply distillation to the depth and ray confidence maps using small weighting factors.
The training details are discussed in~\Cref{sec:app_train_details} in detail.
We keep the video diffusion transformer strictly frozen throughout this stage.
This entire training phase exclusively focuses on transforming the DA3 decoder into a robust geometric readout mechanism for the Wan latent space.

\subsection{Joint Video and 4D Geometry Training}
\label{sec:method_joint}

Once the stitched decoder is finetuned, we use it as a differentiable geometry critic for the video diffusion model.
The video model is trained with the standard latent flow-matching objective~\citep{zhang2026corrflow}.
For each clip, we sample a scalar diffusion time $\tau\in[0,1]$ and use the corresponding scheduler noise scale $\sigma_{\tau}\in\mathbb{R}$.
Both values are shared by all latent frames in the clip and broadcast over space and channels.
Omitting the batch dimension, let $\bm{z}_{\tau}^{1:T}\in\mathbb{R}^{T_{\ell}\times h\times w\times C_z}$ denote the noisy Wan latent trajectory at time $\tau$, where $T_{\ell}$ is the latent-frame length, $(h,w)$ is the VAE spatial grid, and $C_z=16$ for Wan.
The conditioning tuple $c=(\bm{e}_{\bm{p}},\bm{z}_{\mathrm{ctx}},\bm{m}_{\mathrm{ctx}})$ contains the text-action embeddings $\bm{e}_{\bm{p}}\in\mathbb{R}^{L_p\times d_c}$ and context-frame latents $\bm{z}_{\mathrm{ctx}}\in\mathbb{R}^{T_c\times h\times w\times C_z}$ with a binary prefix mask $\bm{m}_{\mathrm{ctx}}\in\{0,1\}^{T_{\ell}\times h\times w\times 1}$.
Given $\bm{z}_{\tau}^{1:T}$, $\tau$, and $c$, the diffusion transformer predicts a velocity field $\bm{v}_{\phi}(\bm{z}_{\tau}^{1:T},\tau,c)\in\mathbb{R}^{T_{\ell}\times h\times w\times C_z}$.
We convert the velocity prediction into a clean-latent estimate
\begin{equation}
\hat{\bm{z}}_0^{1:T} = \bm{z}_{\tau}^{1:T} - \sigma_{\tau}\,\bm{v}_{\phi}(\bm{z}_{\tau}^{1:T}, \tau, c),
\end{equation}
where $\hat{\bm{z}}_0^{1:T}$ has the same tensor shape as $\bm{z}_{\tau}^{1:T}$ and is passed directly into the frozen stitched decoder.
The joint objective is
\begin{equation}
\label{equ:joint}
\mathcal{L}_{\text{joint}}
=
\mathcal{L}_{\text{FM}}
+ \lambda_{\text{geo}}\,
\mathcal{L}_{\text{geo}}
\left(
\mathcal{D}_{\text{stitch}}(\hat{\bm{z}}_0)
\right),
\end{equation}
where $\mathcal{L}_{\text{FM}}$ denotes the latent flow-matching loss and $\mathcal{D}_{\text{stitch}}$ represents the frozen stitched decoder.
The geometric penalty $\mathcal{L}_{\text{geo}}$ is strictly identical to the objective formulated in~\Cref{equ:l_geo}.
During this joint optimization phase, we exclusively utilize the teacher-student distillation mechanism for action-conditioned videos.
As introduced previously, an independently loaded raw DA3 model serves as the teacher network.
It processes the ground-truth RGB clip to provide the reference targets for depth, ray maps, and scene flow. 
For image-to-video generation, the first frame is encoded as the conditioning prefix~\citep{team2025longcat}. 
The diffusion loss can mask the conditioned prefix, while the geometry loss can be applied to the generated suffix. 
As a result, Plan4D learns a single latent trajectory that must satisfy both appearance realism and dense 4D consistency. At inference time, the final latent sequence is decoded by the Wan VAE into RGB video and by the stitched DA3 decoder into depth, camera rays, and scene flow.

%% file: sections/4_experiments.tex
\section{Experiments}
\label{sec:experiments}

\subsection{Experimental Settings.}
\label{sec:exp_settings}
\input{tables/scene_flow_eval}

\heading{Implementation}
\shortname is trained in three stages on diverse datasets due to the lack of large-scale data with paired action and dense 4D annotations. 
\textbf{(1)} We train the designed scene-flow head on top of frozen DA3 using PointOdyssey~\citep{zheng2023pointodyssey}, Dynamic Replica~\citep{karaev2023dynamicstereo}, Virtual KITTI 2~\citep{cabon2020vkitti2}, BlinkVision~\citep{li2024blinkvision}, and OmniWorld-Game~\citep{zhou2025omniworld} for 20 epochs on 16 H100 GPUs.
\textbf{(2)} We fit the latent-to-geometry stitching module on a small subset of the above datasets to initialize the interface from Wan latents to DA3 tokens.
Then we finetune the stitched decoder with the scene-flow-head data mixture augmented by ScanNet++~\citep{yeshwanth2023scannet++} and MVS-Synth~\citep{DeepMVS}. This stage uses 64 H20 GPUs and runs one hybrid epoch over the full data mixture, where the same mixed sampling schedule is traversed once with ground-truth geometric supervision and once with teacher-student distillation. When a ground-truth-supervised sample does not provide scene-flow annotations, we set its scene-flow loss to zero.
\textbf{(3)} We finally train the DiT on a subset of Action100M~\citep{chen2026action100m} for one epoch using 64 H20 GPUs. All stages use a per-GPU batch size of 1. Additional dataset details, optimization details, losses, and supervision definitions are provided in \Cref{sec:app_train_details}.

\paragraph{Evaluation.} We evaluate \shortname along three axes that correspond to its intended use as a generative 4D world model.
\textbf{(1)} we evaluate dynamic 4D geometry through scene flow on MOVi-F~\citep{greff2021kubric}, Dynamic Replica, and PointOdyssey using 81-frame clips, and report endpoint error together with $APD_{0.05}$ against reconstruction and motion baselines. 
\textbf{(2)} we test whether the stitched decoder preserves DA3 geometry after replacing the RGB encoder with a Wan-latent input pathway. This evaluation covers reconstruction, camera, and depth benchmarks, including ETH3D~\citep{schops2017multi}, ScanNet++, Dynamic Replica, Dynamic-RE10K~\citep{chen2026wildrayzer}, MOVi-F, and RobbySimVal~\citep{tan2026masked}, with comparisons to RGB-based 3D foundation models and latent-to-geometry baselines. Scene-flow metrics are computed after rescaling predictions to the target point-map scale. Depth uses per-clip scale-shift alignment, while camera and reconstruction follow the official DA3 bench protocol, with first-camera pose alignment for relative-pose metrics and Umeyama scale alignment for reconstruction. 
\textbf{(3)} finally, we evaluate full action-conditioned 4D world generation on RLBench~\citep{james2020rlbench} (210 samples), HOIGen-1M~\citep{liu2025hoigen} (200 samples), Dynamic Replica (480 samples), and PointOdyssey (375 samples). To further assess in-the-wild generalization, we additionally sample 100 five-second clips from Epic-Kitchens-100~\citep{damen2020epic}, which contains everyday human-object interactions in kitchen environments, and evaluate all compared methods under the same joint-generation protocol. These benchmarks test whether a single generated rollout can jointly preserve visual quality~\citep{ling2025vmbench,zhu2026artifact}, action-text alignment~\citep{chen2025finger}, and temporally coherent 4D structure. We report video quality and instruction following with VBench-style video metrics and video-caption similarity~\citep{huang2024vbench}. For depth, reconstruction, and camera estimation, we follow the same protocol as the stitched-decoder evaluation. We compare with recent 4D world-generation and geometry-generation baselines, including VIST3A~\citep{go2025vist3a}, TesserAct~\citep{zhen2025tesseract}, and Gen3R~\citep{huang2026gen3r}, on the modalities each method supports.















\subsection{Main Results}
\label{sec:exp_results}

\input{figs/dynamic_scene_flow.tex}

\paragraph{Scene Flow Estimation.}
\label{sec:exp_scene_flow}

We first test whether a frozen DA3 geometry prior can support dynamic 4D motion prediction beyond static reconstruction.
This is a stringent setting because scene flow requires temporally coherent geometry rather than accurate per-frame depth alone.
\Cref{tab:scene_flow_main} shows that both scene-flow variants remain competitive on all three benchmarks while using the smallest training set among the compared methods.
The RGB-based DA3 variant is particularly strong on the harder real dynamic benchmarks, suggesting that frozen DA3 tokens already support reliable dense 4D motion readout.
The stitched variant performs best on the controlled synthetic benchmark and remains near the top on Dynamic Replica, which matters because it is the decoder interface used during joint video training.
The qualitative results in \Cref{fig:qualitative-scene-flow-dynamic-recon} show cleaner boundaries, more stable articulated motion, and better global 3D structure than the baselines.
Overall, these results validate the first step of \shortname: DA3 can be extended from geometry reconstruction to strong dynamic 4D readout, and this capability largely survives the latent-to-geometry stitching step.

\input{tables/stiched_model_eval}
\input{figs/geometry_decoder.tex}

\paragraph{Stitched Geometry Decoder Preserves DA3 Behavior.}
\label{subsec:appendix_stitched_da3_alignment}

We evaluate whether replacing DA3's RGB input with Wan video latents preserves its depth, camera, and reconstruction behavior on held-out benchmarks.
We compare against the latest state-of-the-art RGB-based DA3 and VGGT, and the aligned VIST3A-Stitched and Gen3R-VAE decoders.
\Cref{tab:stitched-da3-alignment} shows that our decoder remains closer to the RGB-based models than prior aligned decoders across all three geometry families.
It slightly exceeds DA3 on Dynamic-RE10K camera estimation and remains competitive with VGGT on Dynamic Replica.
Depth changes little on MOVi-F and declines moderately on RobbySimVal.
Reconstruction shifts are larger, but our decoder remains closer to DA3 and VGGT in most settings and achieves the best completeness on Dynamic Replica.
\Cref{fig:qualitative-stitched-results} shows that our decoder preserves DA3-like surface layout and global scene organization, while competing aligned models produce less faithful structure.
This preserved geometry allows the frozen decoder to provide stable supervision for the video generator directly from latent rollouts.

\input{tables/4d_generation_subset}
\begin{figure}[tb!]
\centering
\includegraphics[width=\linewidth]{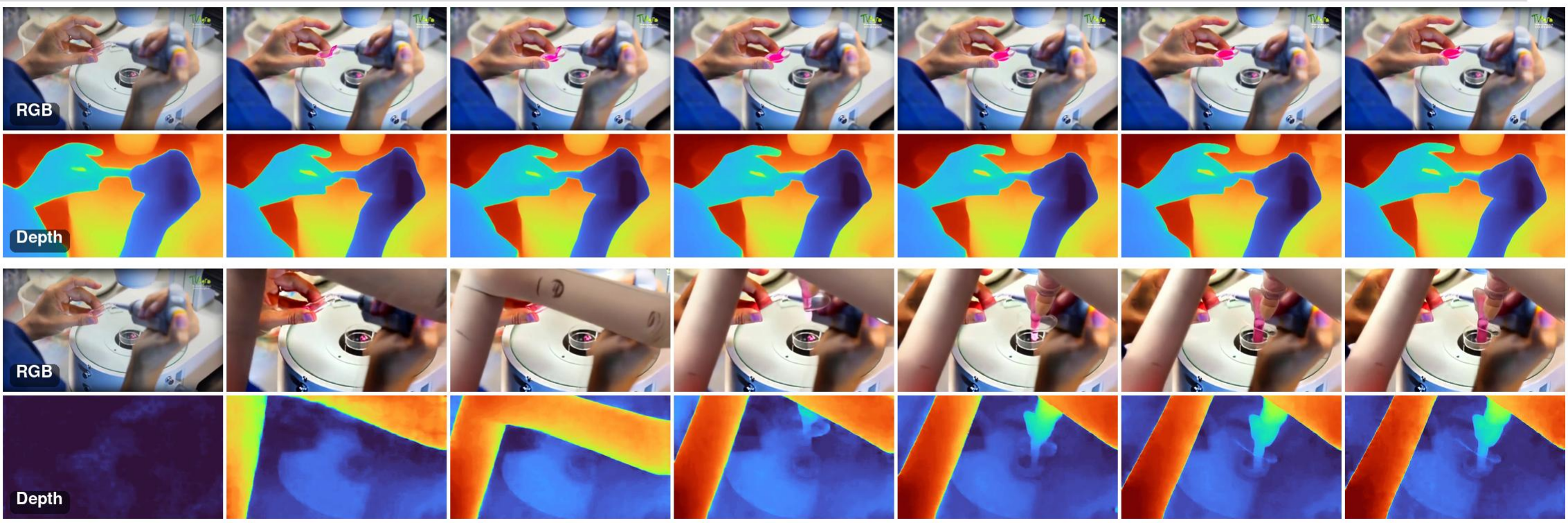}
\caption{\textbf{Qualitative comparison between \shortname and TesserAct~\citep{zhen2025tesseract} for action-conditioned RGB-D world generation.} The input consists of the first RGB frame and the action text: \emph{Hold the small glass dish. Dispense pink liquid using a pipette. Release the liquid smoothly. Fill the dish gradually.}. The first two rows show our results, and the lower rows show TesserAct.}
\label{fig:stage3-rgbd-comparison}
\end{figure}

\paragraph{Joint 4D World Generation}
\label{para:exp_generation}

\input{figs/dynamic_4D_reconstruction.tex}
\Cref{tab:main-results-subset} reports a compact subset of the full joint 4D generation benchmark in \Cref{tab:main_results}.
The subset covers HoiGen, Epic-Kitchens-100, Dynamic Replica, and PointOdyssey across video quality, depth and reconstruction, and camera estimation.
On the 100 five-second Epic-Kitchens-100 clips, \shortname leads on 10 of the 13 metrics in the full comparison.
On video quality, \shortname achieves the best subject consistency on all four datasets and the best caption similarity on Epic-Kitchens-100 and PointOdyssey.
VIST3A has the highest caption similarity on HoiGen and Dynamic Replica.
On Epic-Kitchens-100, \shortname also achieves the best background consistency and image quality, while TesserAct has the best motion smoothness and VIST3A the highest aesthetic quality.
These results show that the shared latent rollout supports strong visual consistency and action-text alignment across diverse scenes.

For depth and reconstruction, \shortname gives the lowest AbsRel on all four datasets in the subset.
It also obtains the best Chamfer-L1 on Epic-Kitchens-100, Dynamic Replica, and PointOdyssey, while Gen3R has the lowest Chamfer-L1 on HoiGen.
On Epic-Kitchens-100, \shortname leads on all three depth metrics.
The HoiGen exception separates local reconstruction quality from the full world-generation objective.
Gen3R can produce a strong point-cloud distance on one hand-object benchmark, but its video quality and camera stability are much weaker in the same setting.

Camera metrics show the largest gap in global 4D consistency.
\shortname obtains the best RRA@5 on all four datasets in the subset and the best RPE-R on HoiGen, Dynamic Replica, and PointOdyssey.
On Epic-Kitchens-100, it leads on both relative-rotation accuracy thresholds.
TesserAct has the lowest RPE-R on Epic-Kitchens-100, while \shortname maintains strong performance across datasets and metric families.
The Epic-Kitchens-100 results complement HoiGen with quantitative evidence across two real-world human-object interaction benchmarks, supporting the applicability of \shortname to everyday interactions beyond the original evaluation setting.
The qualitative comparisons in \Cref{fig:stage3-rgbd-comparison} and \Cref{fig:stage3-gen3r-comparison} show the same pattern visually: the generated RGB frames, RGB-D structure, and global dynamic point clouds remain better aligned under complex action-conditioned rollouts.
These results support the core design choice of supervising a pretrained video generator with a frozen geometric readout rather than learning dense 4D structure from scratch.

\input{tables/ablation_study}

\subsection{Ablation Studies}
\label{sec:exp_abl}

We conduct an ablation study to understand the contribution of each component in our joint-training objective.
The results are summarized in \Cref{tab:ablation},
with results on HoiGen and Dynamic Replica using the same action-conditioned generation protocol.
The RGB-only variant retains strong subject and background consistency, but its depth and camera estimates are less accurate than those of the full model on both datasets. The geometry-only variant improves depth on HoiGen, but sacrifices visual consistency and performs worse across all reported metrics on Dynamic Replica.
The full objective gives the best subject and background consistency and camera estimation on both datasets, and also leads on depth on Dynamic Replica.
These results support combining flow matching with geometric supervision to preserve action-conditioned video quality while improving the depth and camera structure decoded from the same future latent state.
Additional settings and analysis appear in \Cref{subsec:app_joint_objective_ablation}.

%% file: tables/scene_flow_eval.tex
\newcommand{\ph}{\texttt{xx.xx}}   

\begin{table*}[tb!]
\centering
\footnotesize
\setlength{\tabcolsep}{5pt}
\renewcommand{\arraystretch}{1.12}
\caption{\textbf{Scene flow evaluation} on MOVi-F~\citep{greff2021kubric}, Dynamic Replica~\citep{karaev2023dynamicstereo}, and PointOdyssey~\citep{zheng2023pointodyssey} using 81-frame clips. We report training data in millions of frames. EPE and APD$_{0.05}$ are shown without percentage symbols for readability.
The best, second-best, and third-best results within each dataset are highlighted in red, orange, and yellow, respectively.}
\label{tab:scene_flow_main}
\begin{tabular}{@{}lccccccc@{}}
\toprule
\multirow{2}{*}{\textbf{Method}}
  & \multirow{2}{*}{\shortstack{\textbf{Train}\\\textbf{frames / M} $\downarrow$}}
  & \multicolumn{2}{c}{\textbf{MOVi-F}~\citep{greff2021kubric}}
  & \multicolumn{2}{c}{\textbf{Dynamic Replica}~\citep{karaev2023dynamicstereo}}
  & \multicolumn{2}{c}{\textbf{PointOdyssey}~\citep{zheng2023pointodyssey}} \\
\cmidrule(lr){3-4}
\cmidrule(lr){5-6}
\cmidrule(lr){7-8}
  &
  & $\textbf{EPE}\downarrow$ & $\textbf{APD}_{0.05}\uparrow$
  & $\textbf{EPE}\downarrow$ & $\textbf{APD}_{0.05}\uparrow$
  & $\textbf{EPE}\downarrow$ & $\textbf{APD}_{0.05}\uparrow$ \\
\midrule
TraceAnything~\citep{liu2025trace}          & 1.68 & \snd{13.27} & \trd{91.69} & \trd{0.74} & 99.23 & \trd{0.95} & 95.34 \\
Any4D~\citep{karhade2025any4d}              & 1.98 & 27.97 & 47.75 & 3.06 & 85.85 & 24.95 & 37.17 \\
MotionCrafter~\citep{zhu2026motioncrafter}  & \trd{1.67} & 13.48 & 90.86 & 0.82 & \trd{99.25} & 1.14 & 94.50 \\
V-DPM~\citep{sucar2026v}                    & 1.74 & 15.89 & 81.30 & 0.75 & 99.23 & \fst{0.92} & \fst{95.54} \\
Zero-MSF~\citep{liang2025zero} + GT         & \snd{1.36} & \trd{13.33} & \snd{92.47} & \fst{0.45} & \fst{99.28} & 1.04 & \trd{95.41} \\
Ours (DA3)                                  & \fst{0.97} & 13.94 & \snd{92.47} & 0.76 & 99.23 & \snd{0.93} & \snd{95.45} \\
Ours (Stitched)                             & \fst{0.97} & \fst{13.10} & \fst{93.32} & \snd{0.61} & \snd{99.27} & 1.22 & 93.58 \\
\bottomrule
\end{tabular}
\end{table*}

%% file: figs/dynamic_scene_flow.tex
\begin{figure}[tb!]
\centering
\setlength{\tabcolsep}{2pt}
\begin{tabular}{@{}cccc@{}}
\textbf{GT} & \textbf{Ours (DA3)} & \textbf{Any4D}~\citep{karhade2025any4d} & \textbf{MotionCrafter}~\cite{zhu2026motioncrafter} \\
\includegraphics[width=0.24\textwidth]{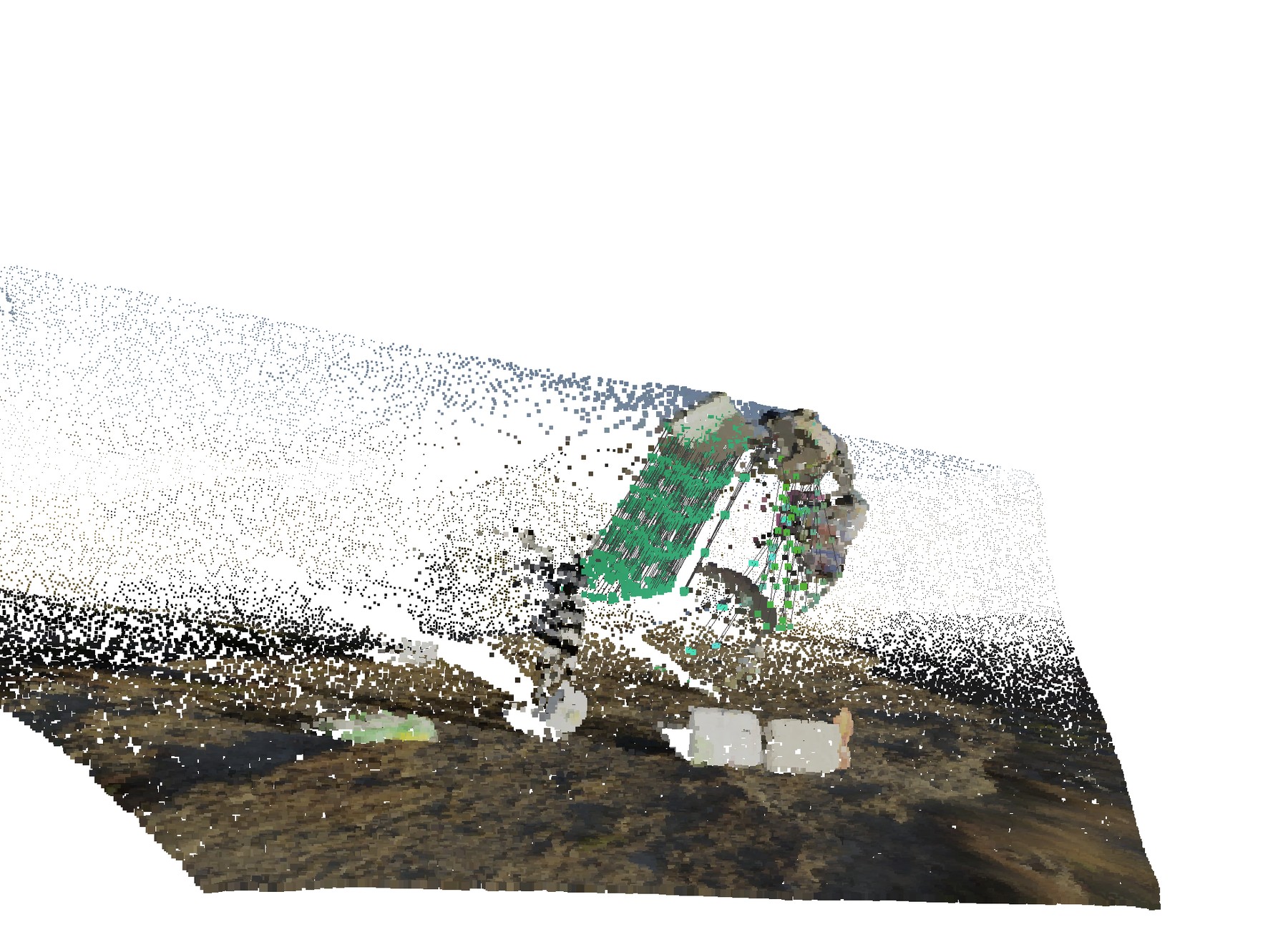} &
\includegraphics[width=0.24\textwidth]{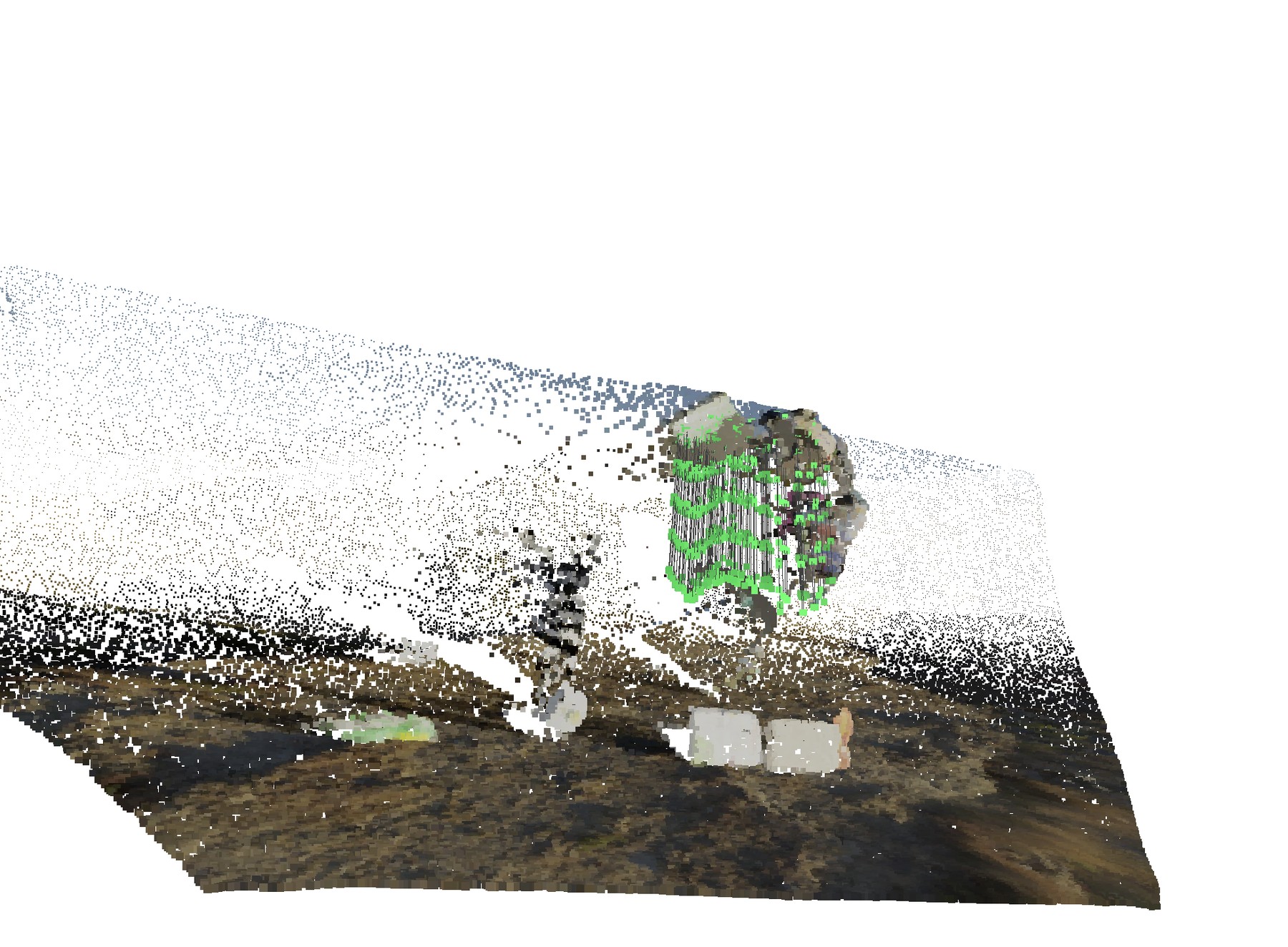} &
\includegraphics[width=0.24\textwidth]{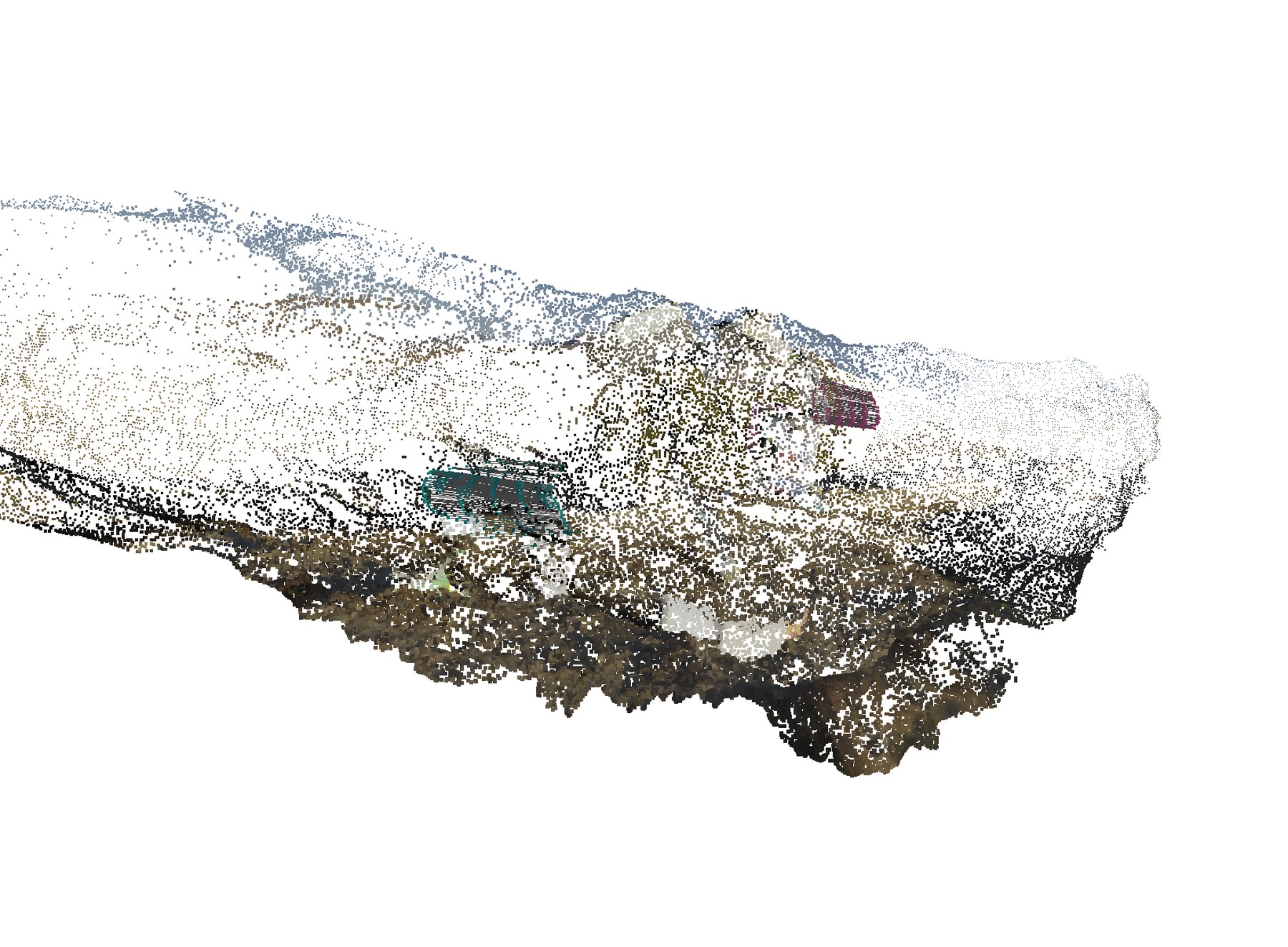} &
\includegraphics[width=0.24\textwidth]{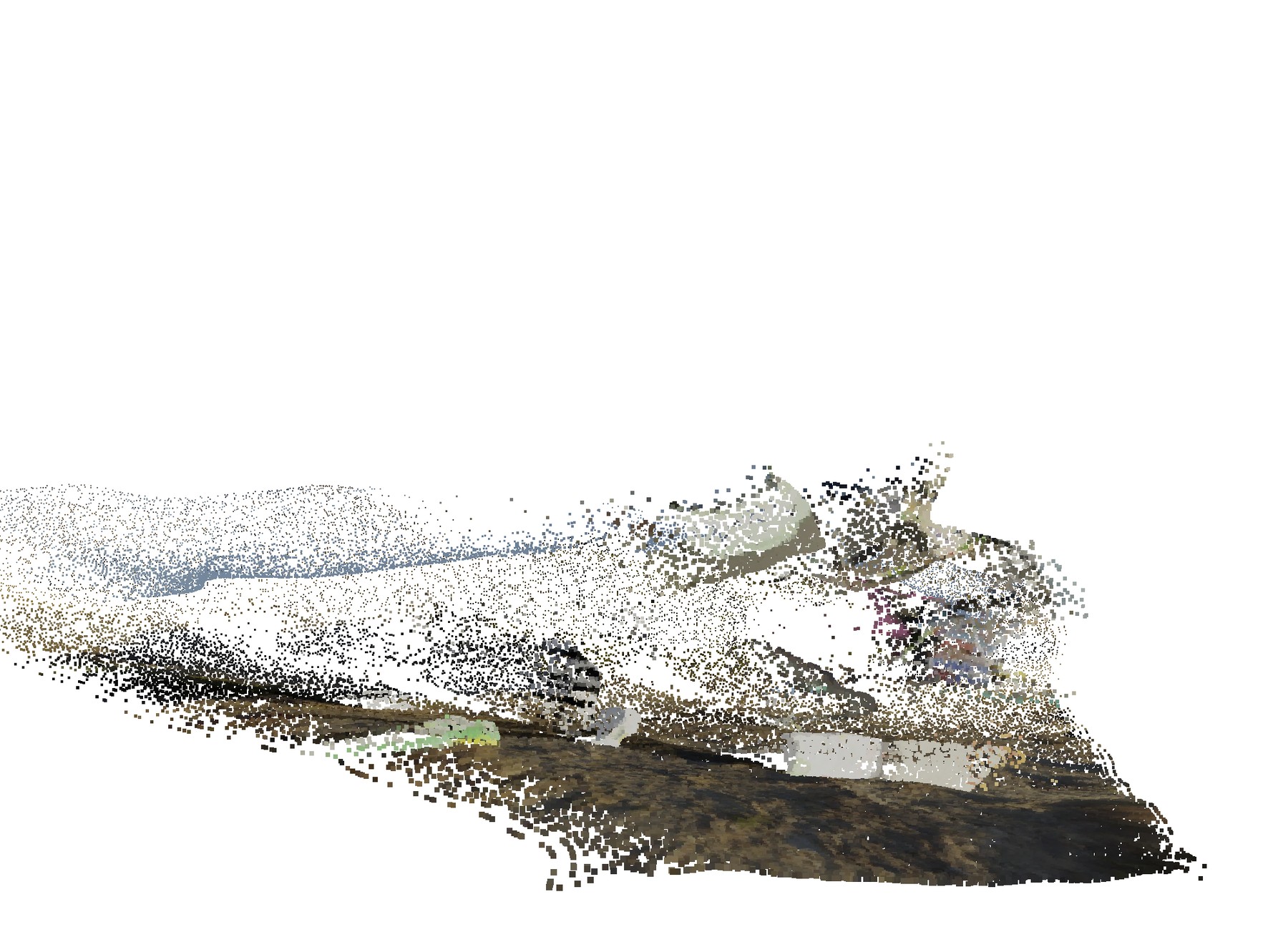} \\
\includegraphics[width=0.24\textwidth]{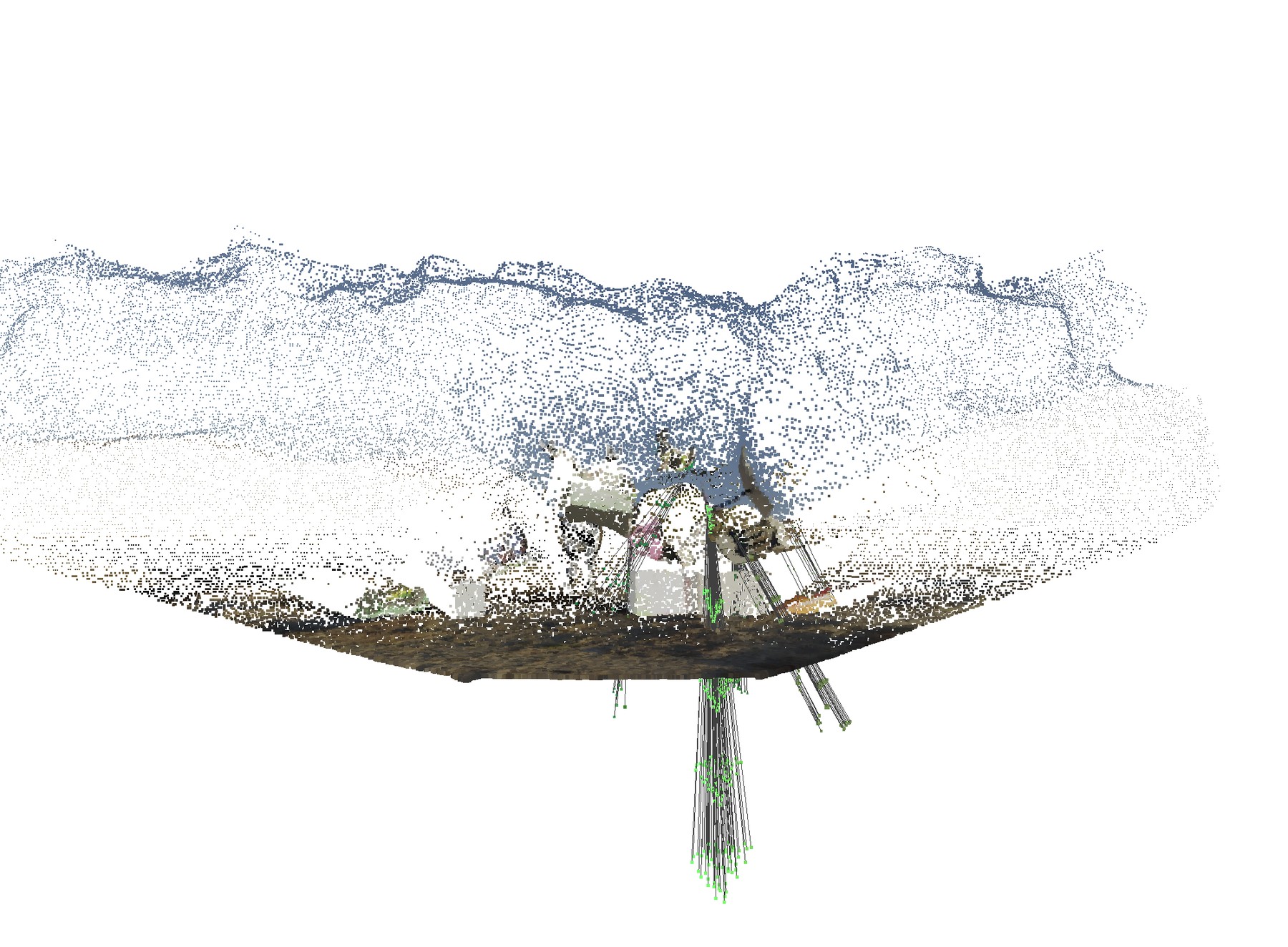} &
\includegraphics[width=0.24\textwidth]{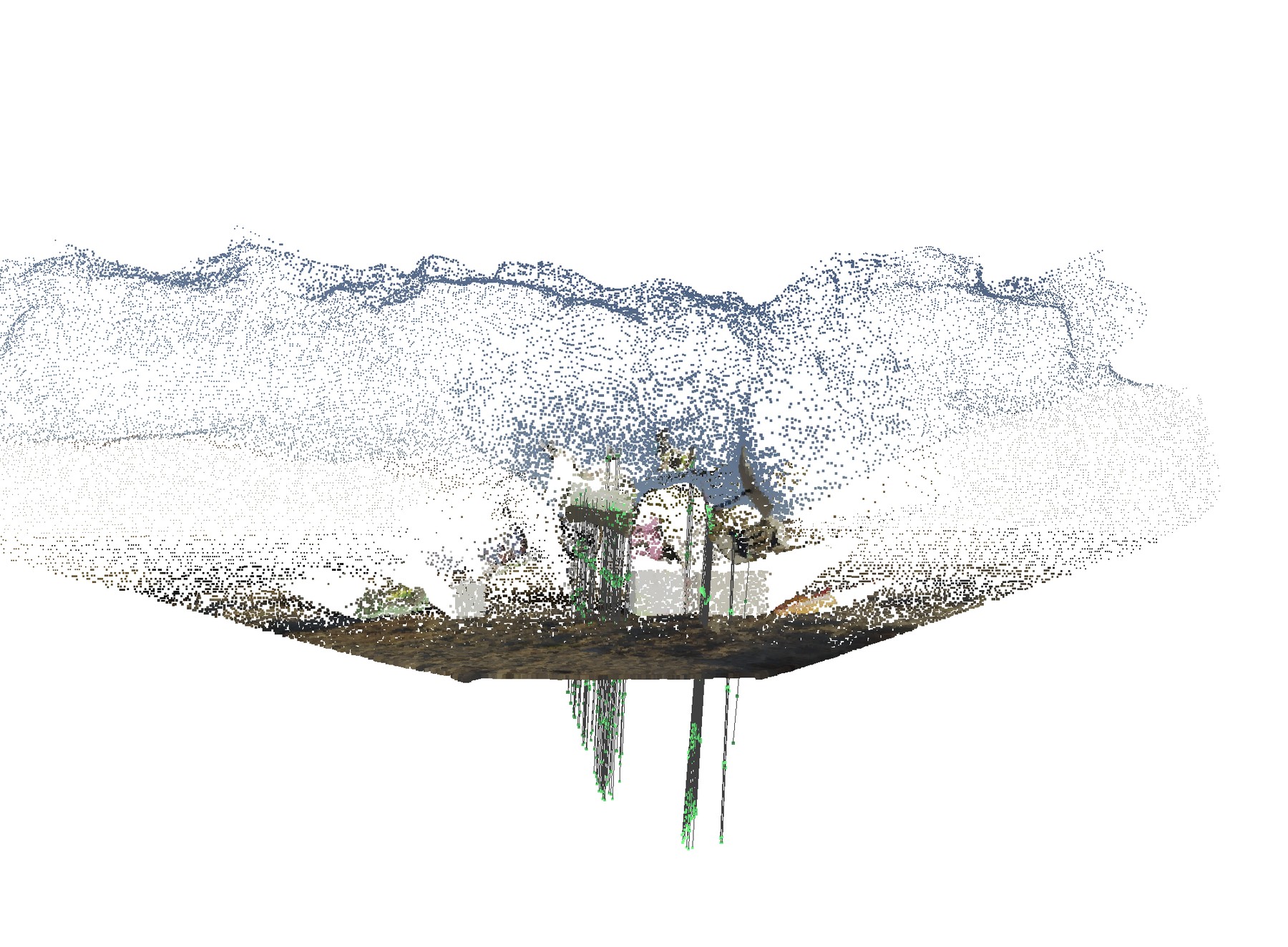} &
\includegraphics[width=0.24\textwidth]{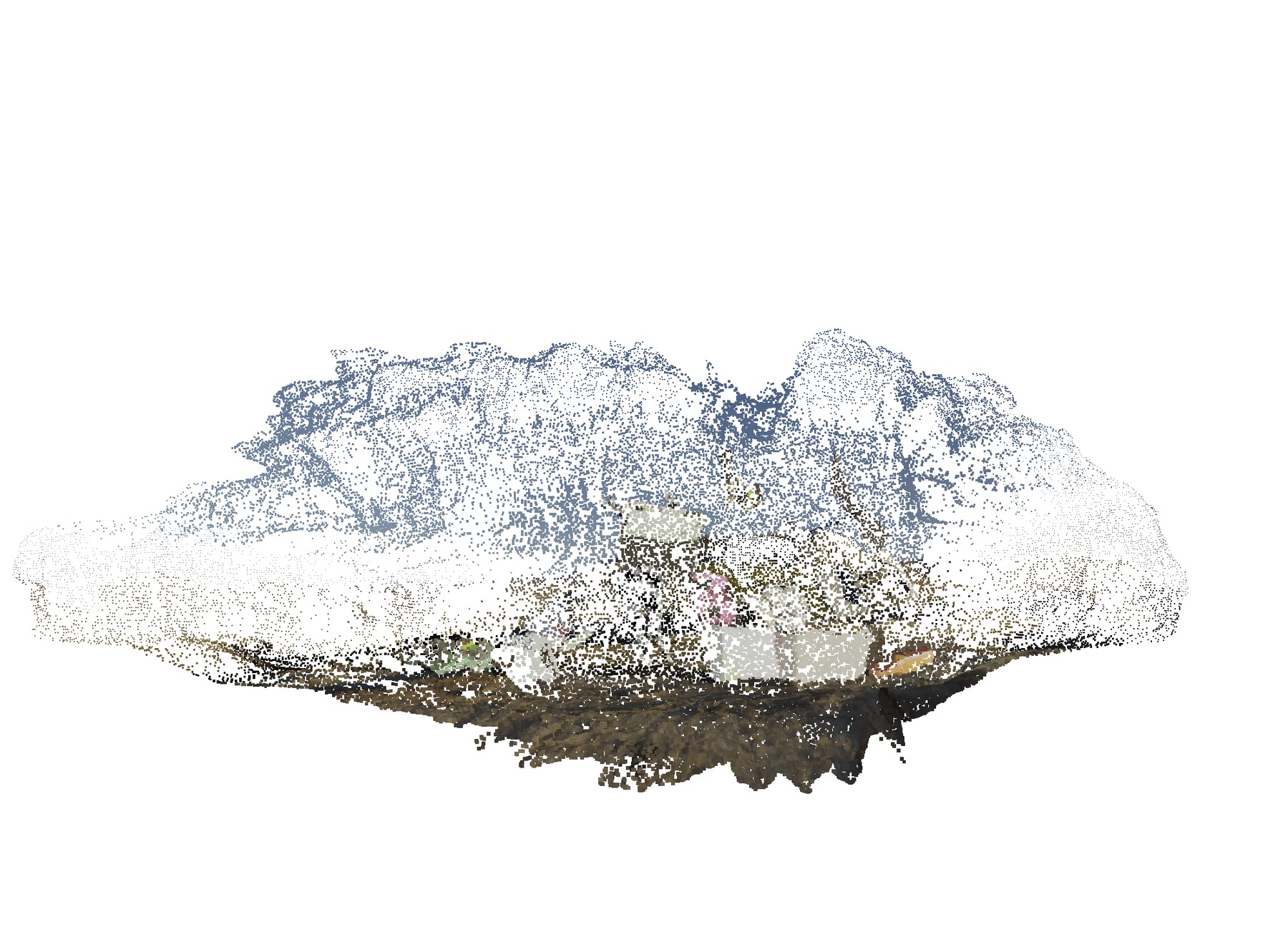} &
\includegraphics[width=0.24\textwidth]{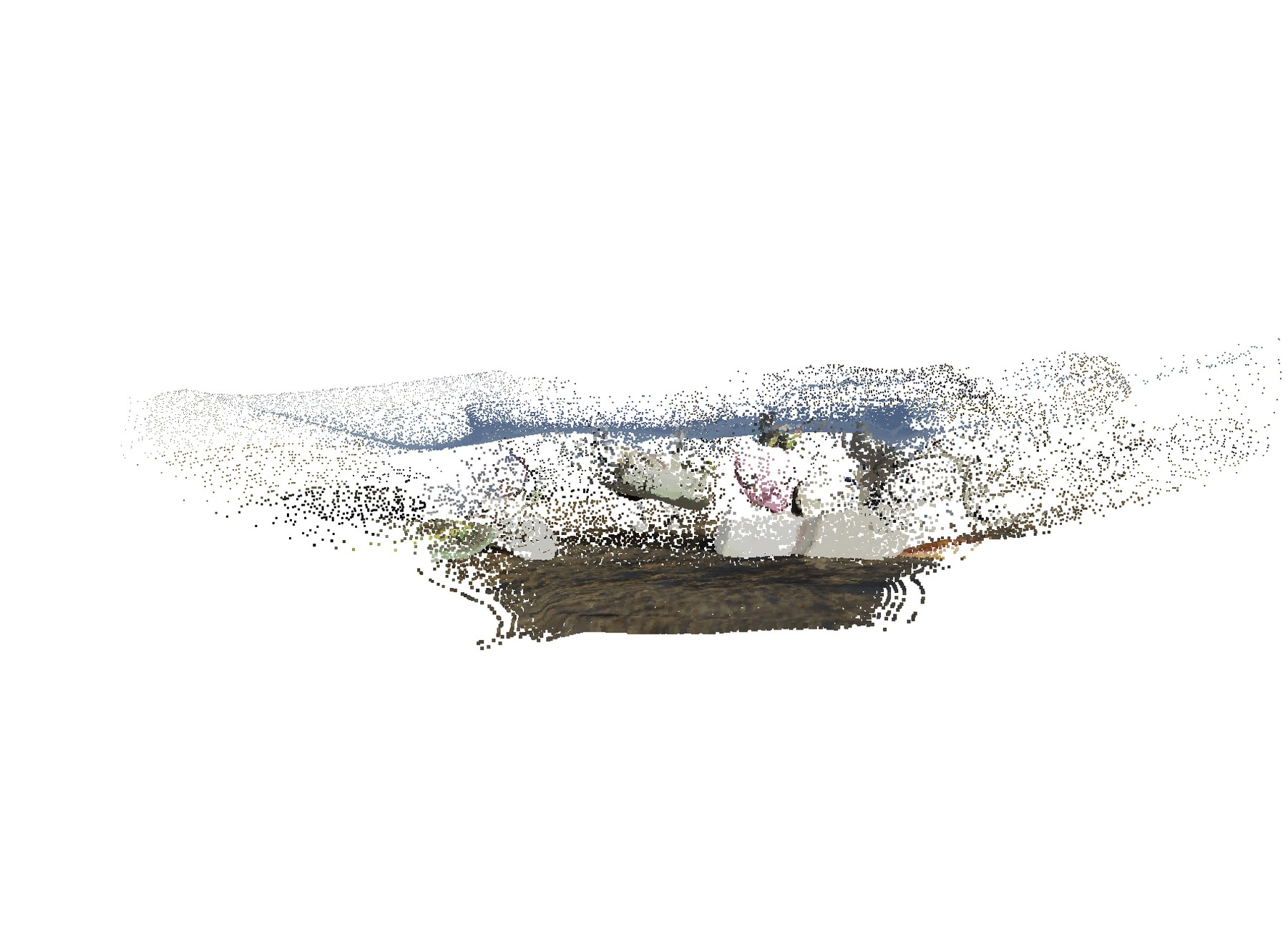}
\end{tabular}
\caption{\textbf{Qualitative results for scene flow and dynamic reconstruction.}
Columns from left to right compare the ground truth, results from our DA3-based prediction, Any4D, and MotionCrafter.
Our result better preserves object-level motion and dynamic 3D structure while maintaining consistency with the reference geometry.}
\label{fig:qualitative-scene-flow-dynamic-recon}
\end{figure}

%% file: tables/stiched_model_eval.tex
\begin{table*}[t]
\centering
\scriptsize
\setlength{\tabcolsep}{3pt}
\caption{
\textbf{Stitched Geometry Decoder Evaluation.}
3D foundation models such as VGGT~\citep{wang2025vggt} and DA3~\citep{lin2025depth} calculate 3D metrics using raw RGB inputs. The remaining three approaches decode geometry from intermediate feature spaces. Gen3R-VAE~\citep{huang2026gen3r} accomplishes this by projecting and reconstructing VGGT tokens.
VIST3A-Stitched~\citep{go2025vist3a} and our method directly process Wan video latents to decode scene geometry.}
\begin{tabular}{@{}lllccccc@{}}
\toprule
\textbf{Dataset} & \textbf{Target} & \textbf{Metric} & \textbf{VGGT}~\citep{wang2025vggt} & \textbf{DA3}~\citep{lin2025depth} & \textbf{VIST3A-Stitched}~\citep{go2025vist3a} & \textbf{Gen3R-VAE}~\citep{huang2026gen3r} & \textbf{Ours (Stitched)} \\
\midrule
ETH3D~\citep{schops2017multi} & Reconstruction & Acc. $\downarrow$ & \snd{0.2802} & \fst{0.2783} & 0.9184 & 1.5954 & \trd{0.2898} \\
ScanNet++~\citep{yeshwanth2023scannet++} & Reconstruction & Comp. $\downarrow$ & \snd{0.1115} & \fst{0.1015} & \trd{0.1425} & 0.2235 & \trd{0.1425} \\
\midrule
Dynamic Replica~\citep{karaev2023dynamicstereo} & Camera & RRA@5$^\circ$ $\uparrow$ & \fst{0.9403} & 0.8971 & \trd{0.9370} & \snd{0.9383} & 0.8987 \\
Dynamic Replica~\citep{karaev2023dynamicstereo} & Reconstruction & Acc. $\downarrow$ & \snd{0.0733} & \fst{0.0700} & 0.3936 & 0.3402 & \trd{0.0967} \\
Dynamic Replica~\citep{karaev2023dynamicstereo} & Reconstruction & Comp. $\downarrow$ & \snd{0.0679} & \trd{0.0834} & 0.4154 & 0.4308 & \fst{0.0658} \\
Dynamic Replica~\citep{karaev2023dynamicstereo} & Reconstruction & F-score $\uparrow$ & \snd{0.6298} & \fst{0.6843} & 0.0740 & 0.0687 & \trd{0.6125} \\
\midrule
Dynamic-RE10K~\citep{chen2026wildrayzer} & Camera & AUC$_{30}$ $\uparrow$ & 0.5149 & \snd{0.5333} & \trd{0.5186} & 0.4834 & \fst{0.5350} \\
Dynamic-RE10K~\citep{chen2026wildrayzer} & Camera & AUC$_{15}$ $\uparrow$ & 0.3854 & \snd{0.4105} & \trd{0.3956} & 0.3268 & \fst{0.4124} \\
\midrule
MOVi-F~\citep{greff2022kubric} & Depth & $\delta_1$ $\uparrow$ & \fst{0.9830} & 0.8932 & 0.8455 & \snd{0.9720} & \trd{0.8943} \\
MOVi-F~\citep{greff2022kubric} & Depth & AbsRel $\downarrow$ & \fst{0.0374} & \trd{0.0877} & 0.1257 & \snd{0.0573} & 0.0931 \\
RobbySimVal~\citep{tan2026masked} & Depth & $\delta_1$ $\uparrow$ & \snd{0.9070} & \fst{0.9273} & 0.8592 & 0.9042 & \trd{0.9055} \\
RobbySimVal~\citep{tan2026masked} & Depth & AbsRel $\downarrow$ & \snd{0.1032} & \fst{0.0808} & 0.1465 & 0.1121 & \trd{0.1085} \\
\bottomrule
\end{tabular}
\label{tab:stitched-da3-alignment}
\end{table*}

%% file: figs/geometry_decoder.tex
\begin{figure}[tb!]
\centering
\begingroup
\setlength{\tabcolsep}{3pt}
\renewcommand{\arraystretch}{1.0}
\begin{tabular}{@{}lcccc@{}}
 & \textbf{DA3}~\citep{lin2025depth} & VIST3A-Stitched~\citep{go2025vist3a} & Gen3R-VAE~\citep{huang2026gen3r} & \textbf{Ours} \\
 &
\includegraphics[width=0.24\textwidth]{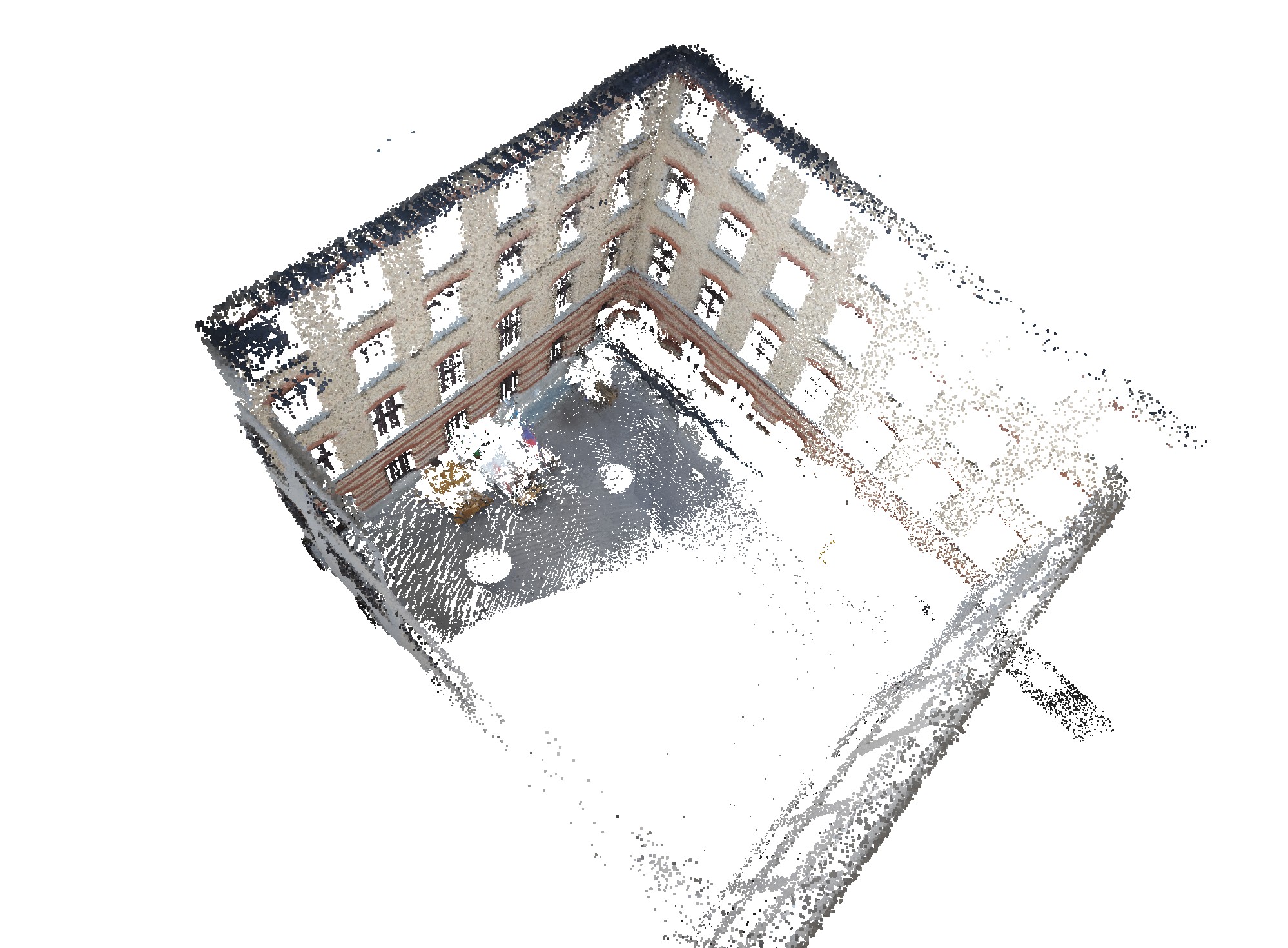} &
\includegraphics[width=0.24\textwidth]{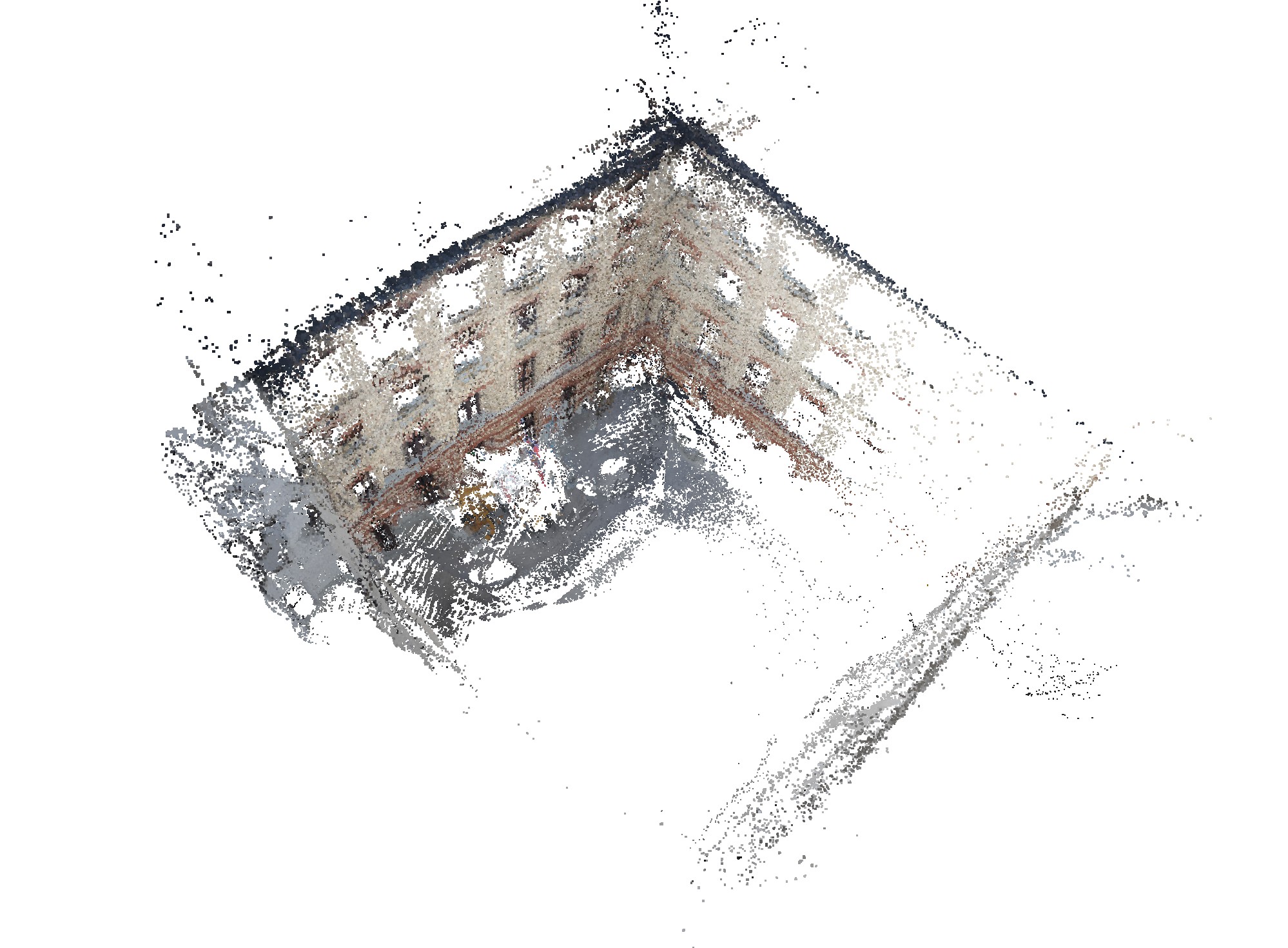} &
\includegraphics[width=0.24\textwidth]{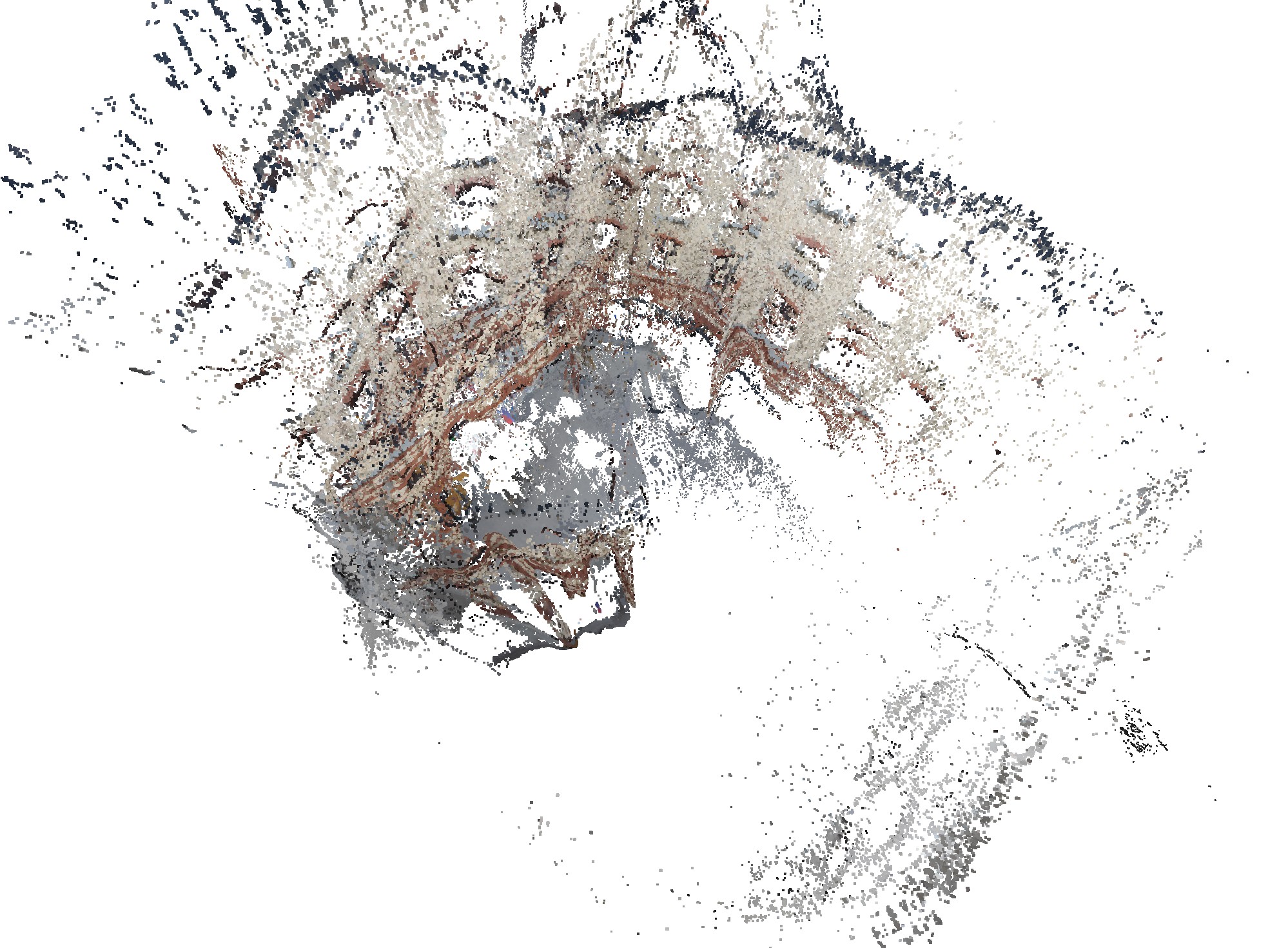} &
\includegraphics[width=0.24\textwidth]{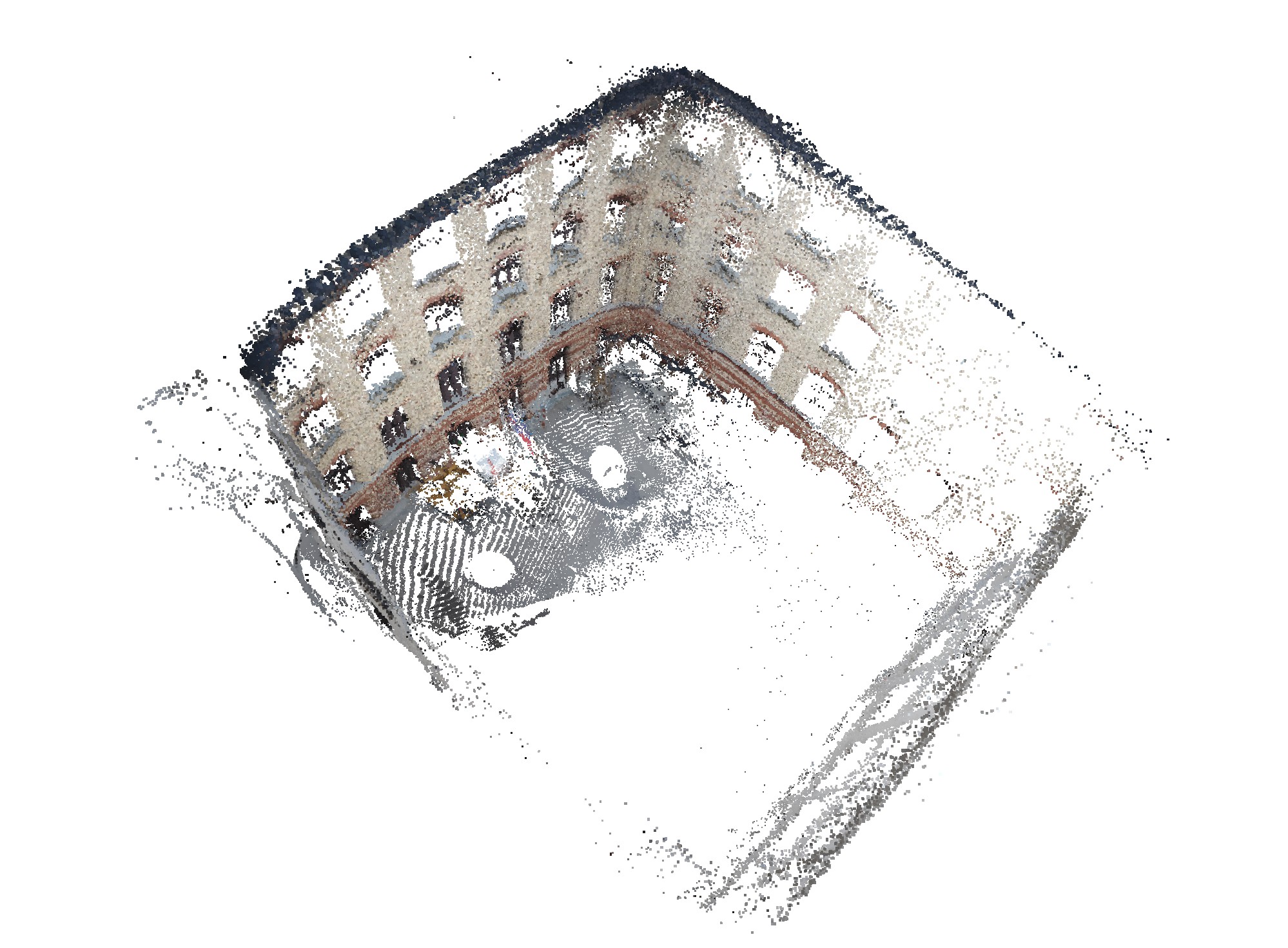} \\
 &
\includegraphics[width=0.24\textwidth]{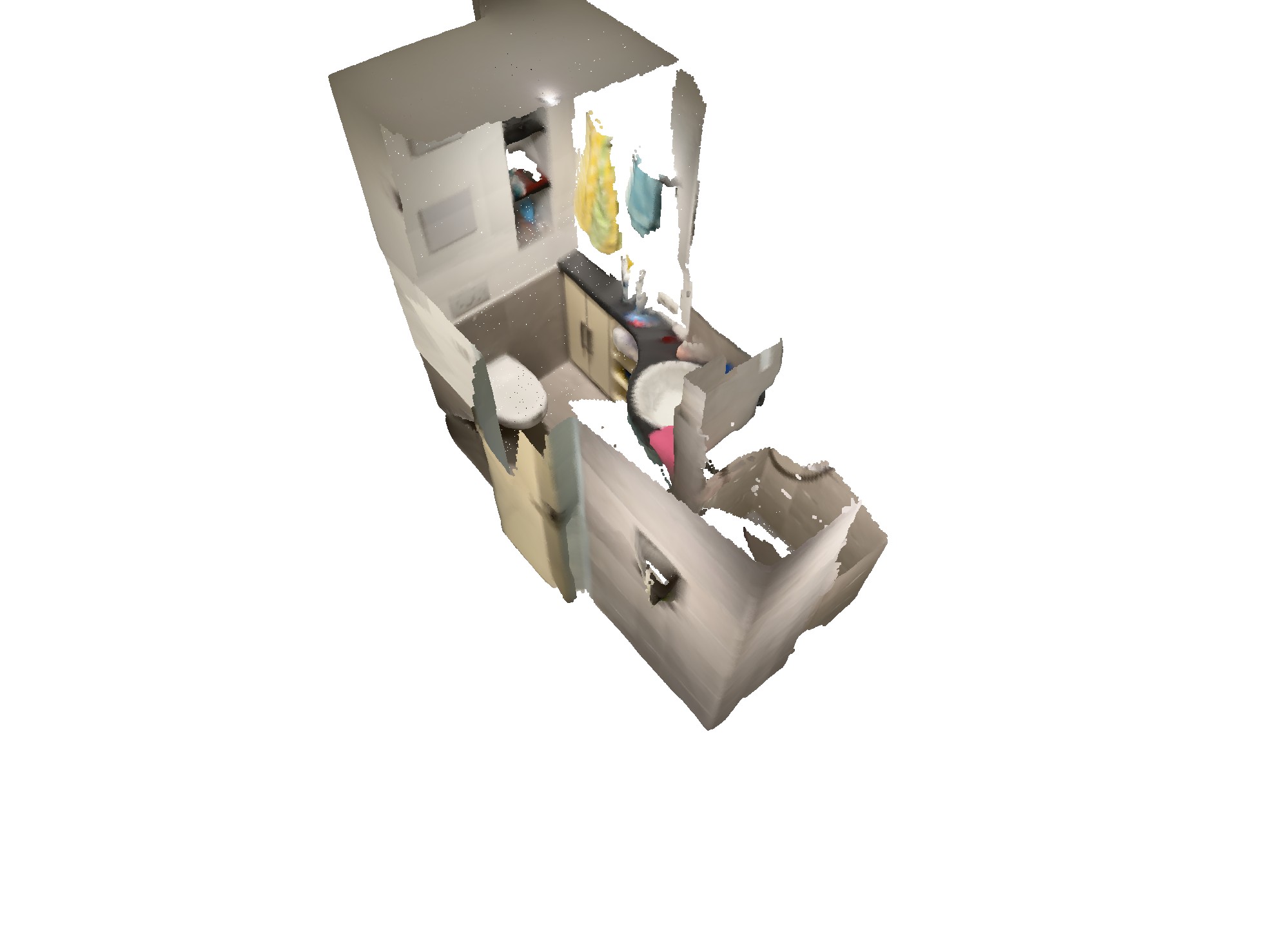} &
\includegraphics[width=0.24\textwidth]{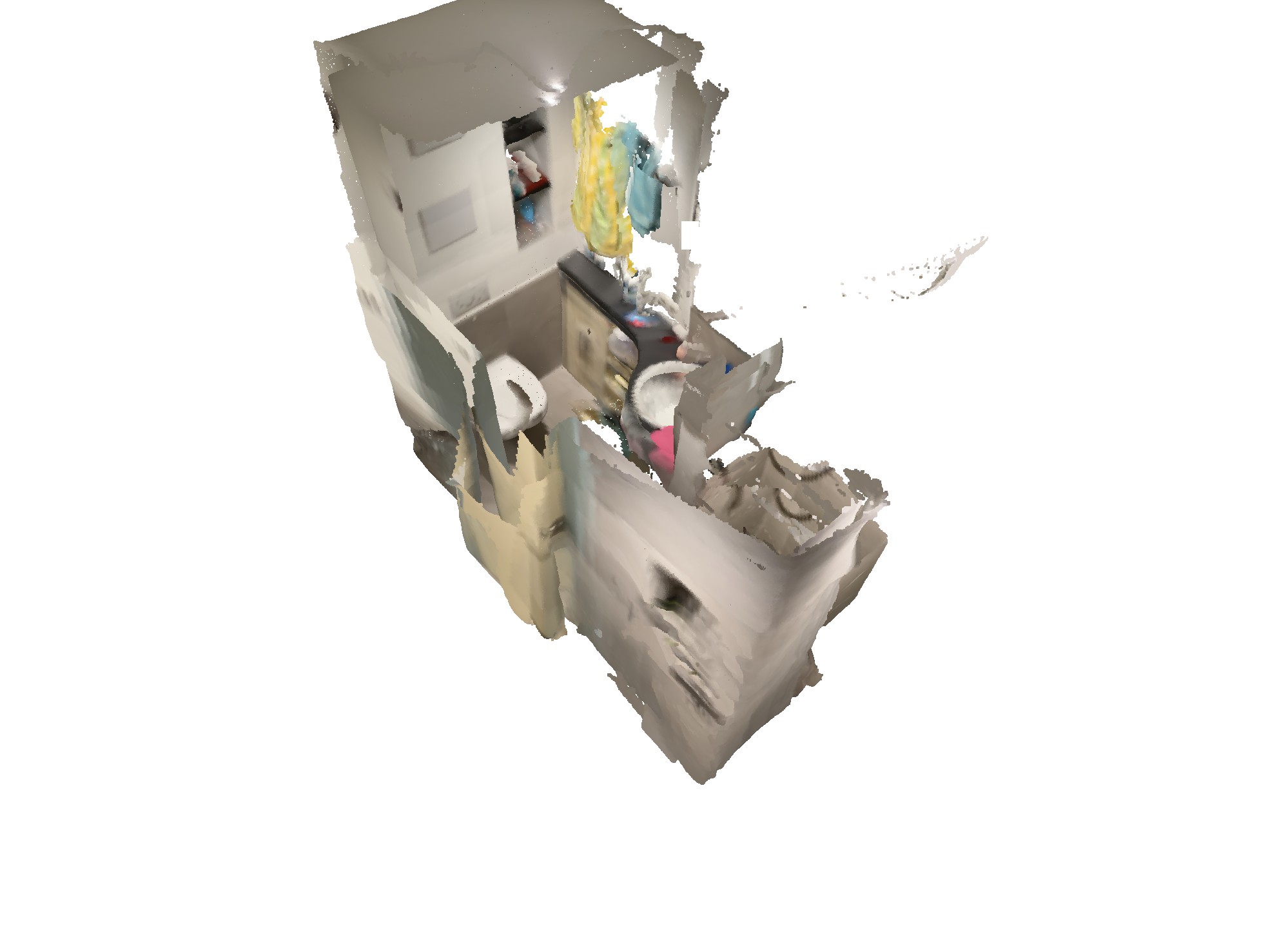} &
\includegraphics[width=0.24\textwidth]{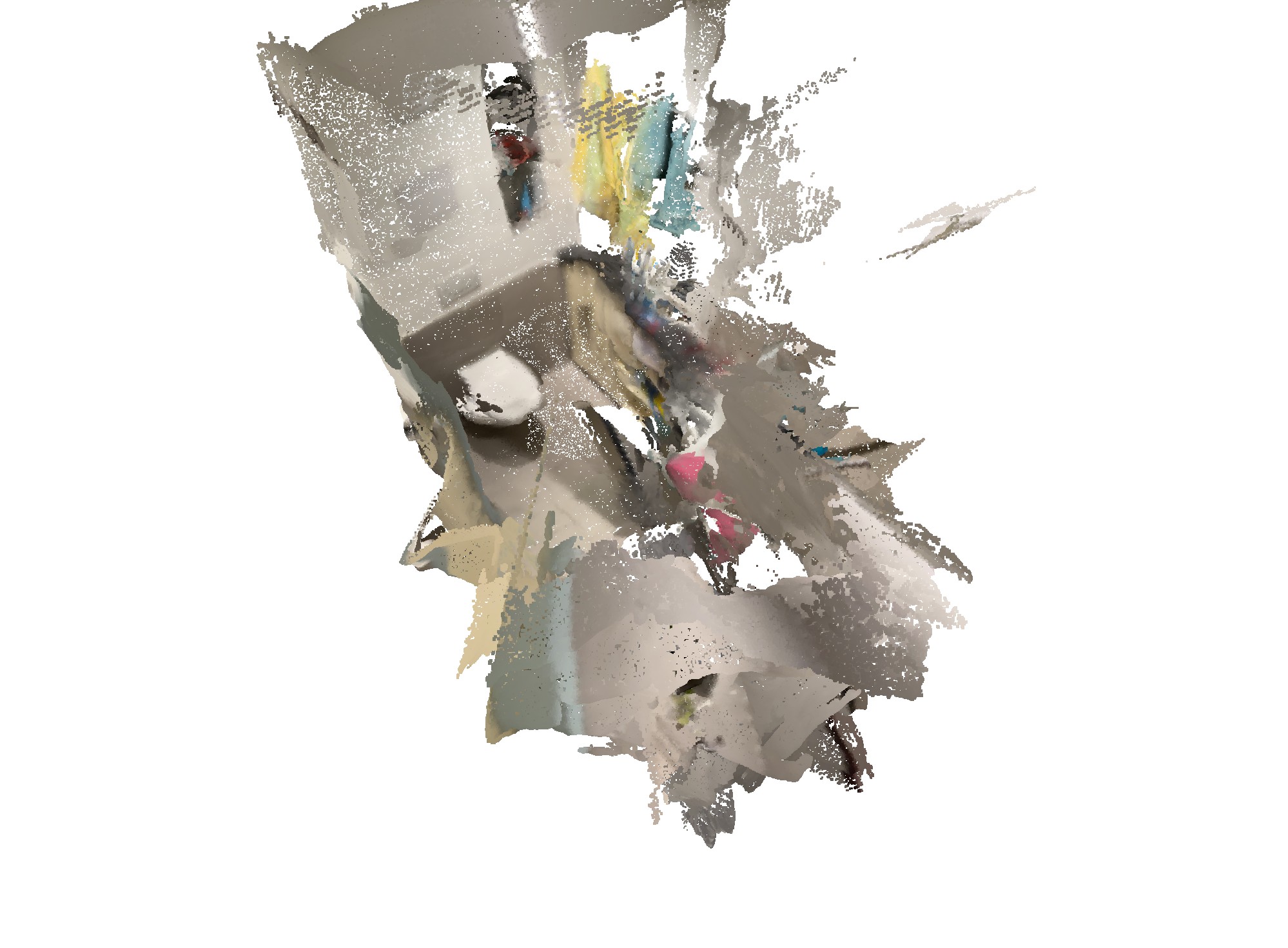} &
\includegraphics[width=0.24\textwidth]{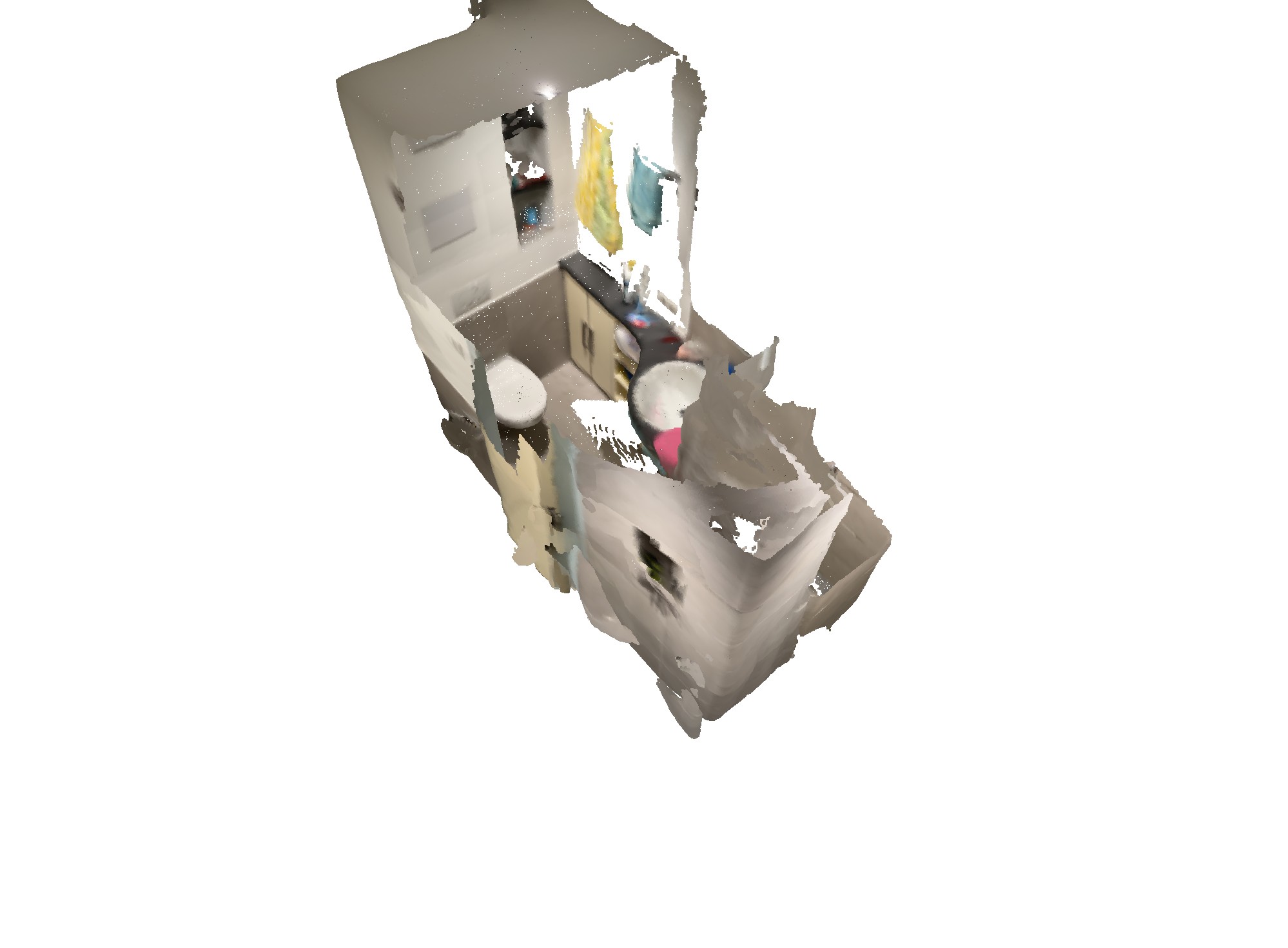} \\
\end{tabular}
\endgroup
\caption{\textbf{Qualitative comparison on ETH3D (top) and ScanNet++ (bottom).}
Our stitched decoder preserves DA3-like surface layout and scene structure more faithfully than the other aligned decoders.}
\label{fig:qualitative-stitched-results}
\end{figure}

%% file: tables/4d_generation_subset.tex
\begin{table*}[tb!]
\centering
\footnotesize
\setlength{\tabcolsep}{5pt}
\renewcommand{\arraystretch}{1.0}
\caption{\textbf{Joint 4D generation results.}
Limited by the space, we only report a subset here; the full results are reported in \Cref{tab:main_results}.
The subset covers video quality, depth and reconstruction, and camera estimation on four benchmarks.
}
\label{tab:main-results-subset}
\begin{tabular}{@{}ll cc cc cc@{}}
\toprule
\multirow{2}{*}{\textbf{Dataset}} & \multirow{2}{*}{\textbf{Model}}
& \multicolumn{2}{c}{\textbf{Video Quality}}
& \multicolumn{2}{c}{\textbf{Depth \& Reconstruction}}
& \multicolumn{2}{c}{\textbf{Camera}} \\
\cmidrule(lr){3-4} \cmidrule(lr){5-6} \cmidrule(lr){7-8}
& & \textbf{SC} $\uparrow$ & \textbf{Cap.Sim} $\uparrow$ & \textbf{AbsRel} $\downarrow$ & \textbf{Chamfer-L1} $\downarrow$ & \textbf{RPE-R} $\downarrow$ & \textbf{RRA@5} $\uparrow$ \\
\midrule
\multirow{4}{*}{HoiGen}
& \textbf{Ours}      & \fst{0.9035} & \snd{0.6854} & \fst{0.1125} & \snd{0.3325} & \fst{0.1240} & \fst{0.9809} \\
& TesserAct & \snd{0.8661} & 0.6401 & 1.850 & 0.3534 & \snd{0.1429} & \snd{0.9256} \\
& Gen3R     & 0.7765 & 0.6803 & \snd{0.2881} & \fst{0.2286} & 5.060 & 0.2689 \\
& VIST3A   & 0.6721       & \fst{0.7851}       & 0.6301       & 0.7633       & 13.09       & 0.0185 \\[1pt]
\midrule
\multirow{4}{*}{\shortstack[l]{Epic-Kitchens-\\100}}
& \textbf{Ours}      & \fst{0.9506} & \fst{0.4010} & \fst{0.0766} & \fst{0.2321} & \snd{0.1568} & \fst{1.0000} \\
& TesserAct & \snd{0.8923} & 0.3343 & 0.4201       & 1.3321 & \fst{0.0946} & \snd{0.9736} \\
& Gen3R     & 0.7813 & 0.3305       & 0.3641 & \snd{0.3390} & 2.3992 & 0.3510 \\
& VIST3A   & 0.5499       & \snd{0.3710} & \snd{0.1326} & 6.4722      & 34.7905      & 0.0184 \\[1pt]
\midrule
\multirow{4}{*}{\shortstack[l]{Dynamic\\Replica}}
& \textbf{Ours}      & \fst{0.9356} & \snd{0.5639} & \fst{0.0696} & \fst{0.2579} & \fst{0.1550} & \fst{1.000} \\
& TesserAct & \snd{0.8490} & 0.4419 & 1.289 & \snd{0.4067} & \snd{0.7609} & \snd{0.6772} \\
& Gen3R     & 0.6833 & 0.4620       & \snd{0.2018} & 1.034 & 22.47       & 0.1393 \\
& VIST3A   & 0.6013       & \fst{0.5744} & 0.4084       & 1.474      & 16.31       & 0.1230 \\[1pt]
\midrule
\multirow{4}{*}{\shortstack[l]{Point\\Odyssey}}
& \textbf{Ours}      & \fst{0.9233} & \fst{0.5579} & \fst{0.0822} & \fst{0.3013} & \fst{0.1985} & \fst{0.9864} \\
& TesserAct & \snd{0.9019} & 0.5094 & 0.6347 & \snd{0.3904} & \snd{0.8804} & \snd{0.6446} \\
& Gen3R     & 0.7486 & 0.4930 & \snd{0.4227} & 0.6703 & 9.409       & 0.2188 \\
& VIST3A   & 0.6270       & \snd{0.5465}       & 0.4371       & 1.347      & 6.953       & 0.2962 \\
\bottomrule
\end{tabular}
\end{table*}

%% file: figs/dynamic_4D_reconstruction.tex
\begin{wrapfigure}{R}{0.55\textwidth}
    \vspace{-1em}
    \centering
    \includegraphics[width=\linewidth]{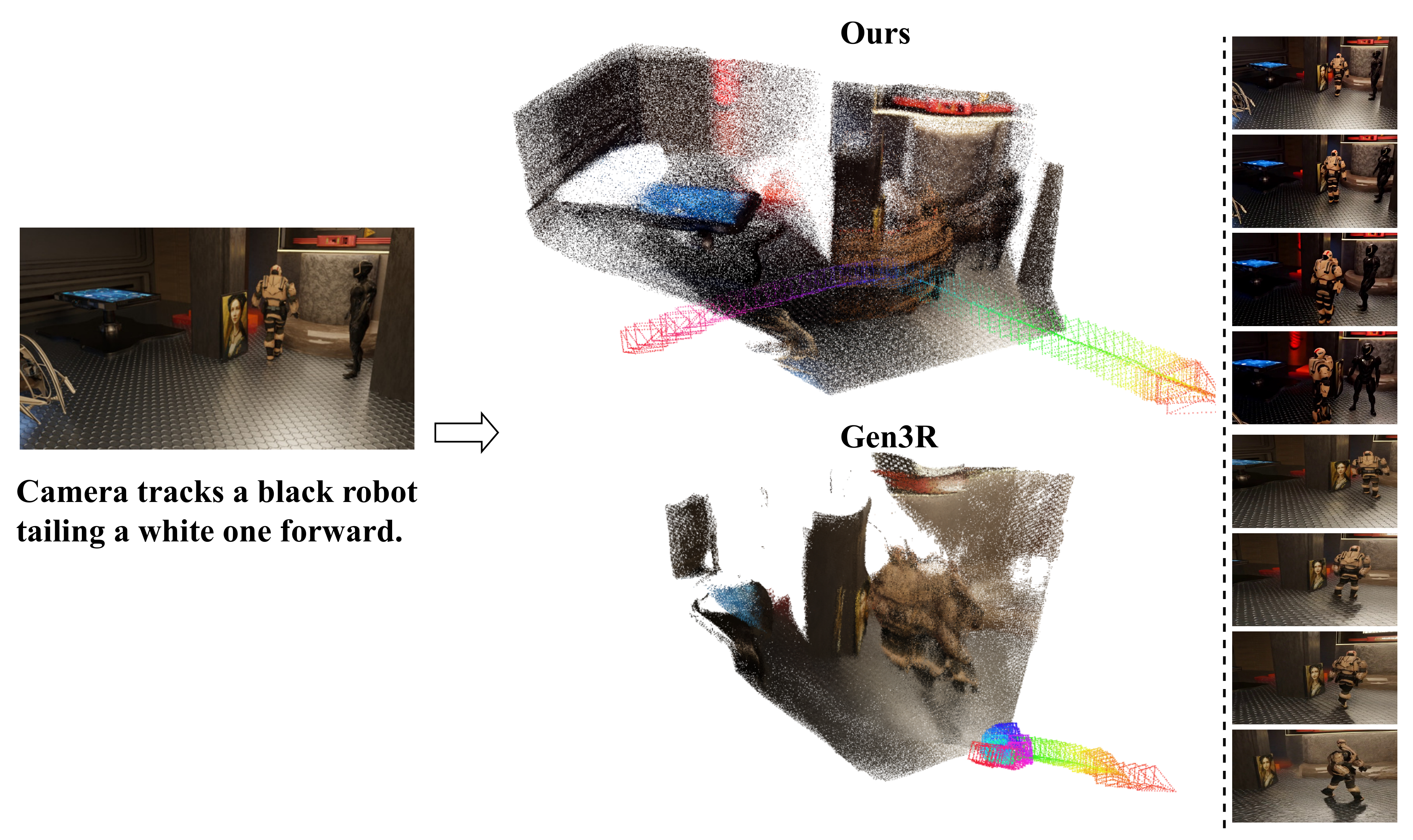}
    \caption{\textbf{Qualitative comparison between \shortname and Gen3R~\citep{huang2026gen3r} for dynamic 4D world generation.}
    The middle panels show the global dynamic point-cloud reconstruction, and the right panels show the corresponding RGB frames.}
    \label{fig:stage3-gen3r-comparison}
    \vspace{-1.0em}
\end{wrapfigure}

%% file: tables/ablation_study.tex
\begin{table}[t]
\centering
\footnotesize
\setlength{\tabcolsep}{4pt}
\renewcommand{\arraystretch}{1.1}
\caption{\textbf{Ablation of the joint-training objective.}
We ablate RGB-only, geometry-only, and full objectives on HoiGen and Dynamic Replica under the same action-conditioned 4D generation protocol.
SC and BC denote VBench subject and background consistency.
Arrows indicate the preferred direction, and the best and second-best results within each dataset are highlighted.}
\label{tab:ablation}
\begin{tabular*}{\linewidth}{@{\extracolsep{\fill}}ll cc cc c@{}}
\toprule
\multirow{2}{*}{\textbf{Dataset}} & \multirow{2}{*}{\textbf{Model}}
& \multicolumn{2}{c}{\textbf{Video Quality}}
& \multicolumn{2}{c}{\textbf{Depth}}
& \textbf{Camera} \\
\cmidrule(lr){3-4} \cmidrule(lr){5-6} \cmidrule(lr){7-7}
& & \textbf{SC} $\uparrow$ & \textbf{BC} $\uparrow$ & \textbf{AbsRel} $\downarrow$ & $\boldsymbol{\delta_1}$ $\uparrow$ & \textbf{RPE-R} $\downarrow$ \\
\midrule
\multirow{3}{*}{HoiGen}
& Only-RGB-Loss & \snd{0.8889} & \snd{0.9272} & 0.1559      & 0.7769      & 0.4010 \\
& Only-Geo-Loss & 0.6198       & 0.8184       & \fst{0.0745} & \fst{0.9391} & \snd{0.3232} \\
& Ours (Full)   & \fst{0.9035} & \fst{0.9334} & \snd{0.1125} & \snd{0.8540} & \fst{0.1240} \\[1pt]
\midrule
\multirow{3}{*}{\shortstack[l]{Dynamic\\Replica}}
& Only-RGB-Loss & \snd{0.8998} & \snd{0.9321} & \snd{0.0874} & \snd{0.9169} & \snd{0.2615} \\
& Only-Geo-Loss & 0.6009       & 0.8459       & 0.1733      & 0.7077      & 0.7698 \\
& Ours (Full)   & \fst{0.9356} & \fst{0.9496} & \fst{0.0696} & \fst{0.9306} & \fst{0.1550} \\
\bottomrule
\end{tabular*}
\end{table}

%% file: sections/5_conclusion.tex
\section{Conclusion}
\label{sec:conclusion}

We introduce the problem of \emph{generative plannable 4D worlds}, where a model must generate future scenes with rich 4D attributes,
including RGB appearance, depth, camera poses, point clouds, and scene flow,
from observed context and action descriptions.
To this end, we propose \shortname, a unified framework that decouples the acquisition of strong geometric priors from the generation of future dynamics, then reconnects them through a stitched latent world model.
By augmenting frozen DA3 with a token-only scene-flow head and learning a latent-to-geometry stitching interface from video VAE latents to geometry tokens, \shortname turns a pretrained video generator into an actionable 4D world model whose RGB, depth, camera rays, point clouds, and scene flow are read out from the same evolving latent state in a single feed-forward pass.
Experiments show that this design preserves strong geometric behavior while improving dense 4D world generation across scene flow, reconstruction, and action-conditioned future prediction, without requiring manual 4D annotations for action videos.
We hope this perspective helps move future world modeling beyond video-only prediction toward unified, measurable, and plannable 4D world models.

\section*{Acknowledgement}
This research was done while Hangwei Zhang was an intern at AMAP, Alibaba Group.
This research is supported by the NTU SUG-NAP and National Research Foundation, Singapore, under its NRF Fellowship Award NRF-NRFF17-2025-0009, and the National Research Foundation, Singapore under its AI Singapore Programme (AISG Award No: AISG3-PhD-2026-08-079).

%% file: sections/appendix.tex
\newpage
\appendix
\makeatletter
\captionsetup[table]{position=top,aboveskip=\@neuripsabovecaptionskip,belowskip=\@neuripsbelowcaptionskip}
\makeatother
\renewcommand{\topfraction}{0.95}
\renewcommand{\bottomfraction}{0.9}
\renewcommand{\textfraction}{0.05}
\renewcommand{\floatpagefraction}{0.9}
\setlength{\floatsep}{12pt plus 2pt minus 2pt}
\setlength{\textfloatsep}{14pt plus 2pt minus 2pt}
\setcounter{topnumber}{4}
\setcounter{bottomnumber}{4}
\setcounter{totalnumber}{6}

\section{Frozen DA3 Tokens Already Encode Cross-Frame Motion Cues}
\label{sec:appendix_frozen_da3_analysis}
The lightweight DPT scene flow head above is only useful if the frozen DA3 encoder tokens already preserve motion-relevant structure. The central question of this section is therefore simple: before any scene-flow supervision reaches the decoder, do frozen DA3 tokens already know which location in frame $t+1$ corresponds to a location in frame $t$? If the answer is yes, then a shallow head can decode dense motion by reading out a representation that already contains cross-frame geometry and correspondence cues. If the answer is no, then good scene-flow predictions would have to come from the capacity of the head alone.

\begin{figure*}[!tbp]
\centering
\begin{minipage}[t]{0.49\textwidth}
\centering
\includegraphics[width=\linewidth]{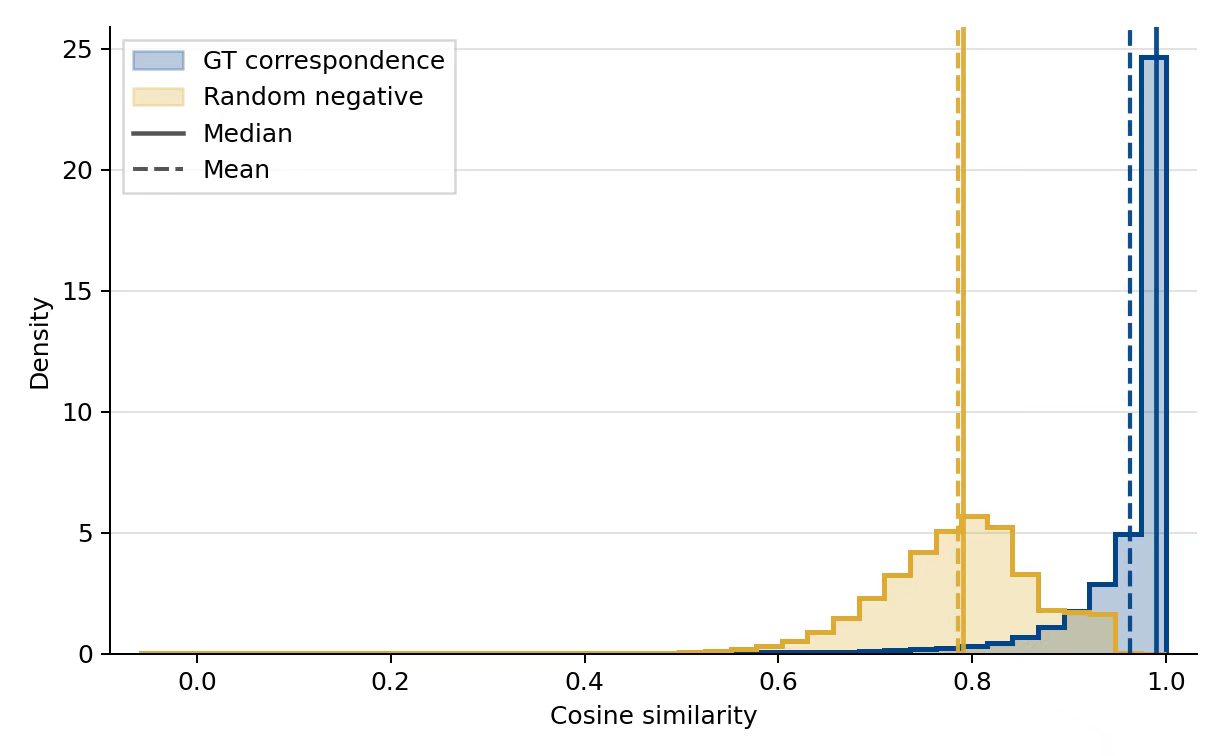}
\end{minipage}\hfill
\begin{minipage}[t]{0.49\textwidth}
\centering
\includegraphics[width=\linewidth]{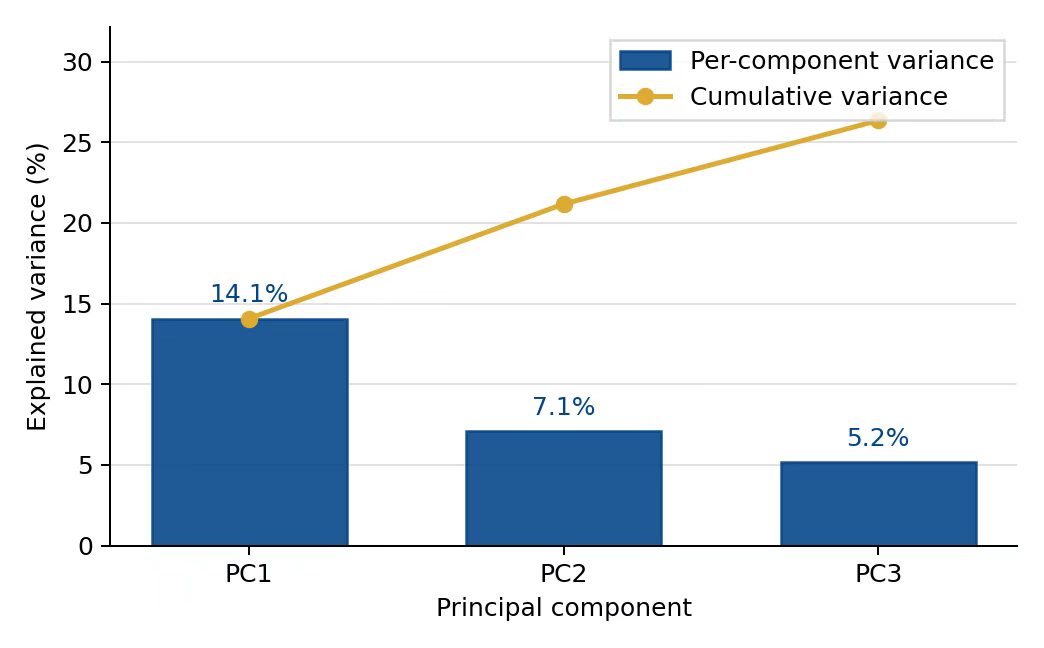}
\end{minipage}
\caption{Aggregated token-space evidence on the Point Odyssey test split. Left: cosine similarity distributions computed from frozen DA3 tokens. Right: explained variance of the top principal components of token differences $\Delta z_t = z_{t+1} - z_t$, aggregated over all test clips.}
\label{fig:da3-frozen-token-motion}
\end{figure*}

To make this question concrete, let $z_t(p)$ denote the frozen DA3 token at spatial cell $p$ in frame $t$. Point Odyssey~\citep{zheng2023pointodyssey} provides ground-truth tracks, which let us identify the corresponding target cell $p^{+}$ in frame $t+1$ for each valid source cell $p$. We then probe the token space without using the scene flow head. Our analysis uses the Point Odyssey test split with 128 clips from 5 scenes. All values in Table~\ref{tab:da3-frozen-token-motion} are weighted over clips by the number of valid token pairs or valid flow tokens. Figure~\ref{fig:da3-frozen-token-motion} aggregates all 6,132,636 positive pairs, 6,132,636 negative pairs, and 2,160,723 token-difference vectors.

We use four complementary probes, each answering a different question about the frozen representation. The first probe is \textbf{correspondence similarity} \citep{tang2023emergent}. It asks whether a token is more similar to its true match than to an unrelated token from the next frame:
\begin{equation}
s_{\mathrm{pos}} = \cos\!\bigl(z_t(p), z_{t+1}(p^{+})\bigr),
\qquad
s_{\mathrm{neg}} = \cos\!\bigl(z_t(p), z_{t+1}(q^{-})\bigr),
\end{equation}
where $q^{-}$ is a random negative sampled from frame $t+1$. This is the cleanest test of whether correspondence information is present~\citep{jabri2020space}. If frozen DA3 tokens contain motion-relevant structure, the $s_{\mathrm{pos}}$ distribution should shift to the right of the $s_{\mathrm{neg}}$ distribution. The margin $\Delta s = s_{\mathrm{pos}} - s_{\mathrm{neg}}$ and the probability $\Pr(s_{\mathrm{pos}} > s_{\mathrm{neg}})$ summarize how reliably the two distributions separate.

The second probe is \textbf{token retrieval} \citep{oquab2023dinov2}. Instead of comparing one positive and one negative, we search over all target tokens in frame $t+1$:
\begin{equation}
\hat{p} = \arg\max_{q \in \mathcal{P}_{t+1}} \cos\!\bigl(z_t(p), z_{t+1}(q)\bigr).
\end{equation}
This is a stronger operational test. It asks whether nearest-neighbor search in frozen token space can recover the correct destination cell. We report \emph{Top-1 exact}, which measures whether the retrieved cell is exactly the ground-truth cell, and \emph{PCK@1} and \emph{PCK@3}, which measure whether retrieval falls within 1 or 3 token cells of the ground-truth location~\citep{yang2012articulated}. High retrieval accuracy means that the token space is not only separable on average, but also organized enough to support direct correspondence lookup.

The third probe is a \textbf{token-flow readout}. After retrieval returns $\hat{p}$, we interpret it as a token-level motion estimate $\hat{d} = \hat{p} - p$ and compare it with the ground-truth token displacement. This probe is more demanding than similarity or retrieval because it asks a single nearest-neighbor lookup to act as a motion predictor. The resulting endpoint error and direction cosine should be interpreted as evidence for how directly motion can be read out from the frozen tokens without a learned decoder~\citep{xu2022gmflow}.

The fourth probe is \textbf{PCA on token differences}. For each matched pair, we form
\begin{equation}
\Delta z_t(p) = z_{t+1}(p^{+}) - z_t(p).
\end{equation}
We then perform PCA on these token differences~\citep{raghu2017svcca}. This analysis does \emph{not} test whether motion is present at all. The earlier probes already answer that question more directly. Instead, PCA asks a more specific question: is the motion signal concentrated in a small linear subspace, or is it distributed across many token dimensions? If the first few principal components explain most of the variance, then motion could be close to linearly disentangled~\citep{alain2016understanding}. If the variance remains spread out, then motion is still present, but encoded in a more distributed form. This distinction matters for interpreting the role of the DPT head. A low-dimensional linear code would be easy to read out with almost any linear probe, while a distributed code benefits from structured multiscale decoding.

\begin{table}[t]
\centering
\small
\setlength{\tabcolsep}{6pt}
\caption{Dataset-level evidence from frozen DA3 tokens on the Point Odyssey test split. Each block should be read differently. Similarity measures token-level separability between true matches and negatives. Retrieval measures whether nearest-neighbor search recovers the correct target cell. Token flow measures how well that retrieval can be turned into a direct motion estimate. PCA measures whether motion is concentrated in a small linear subspace.}
\label{tab:da3-frozen-token-motion}
\begin{tabular}{lll}
\hline
Category & Metric & Value \\
\hline
Similarity & Mean GT similarity $s_{\mathrm{pos}}$ & 0.962 \\
Similarity & Mean negative similarity $s_{\mathrm{neg}}$ & 0.784 \\
Similarity & Mean gap $\Delta s$ & 0.178 \\
Similarity & $\Pr(s_{\mathrm{pos}} > s_{\mathrm{neg}})$ & 96.98\% \\
\hline
Retrieval & Top-1 exact & 69.65\% \\
Retrieval & PCK@1 token & 86.01\% \\
Retrieval & PCK@3 token & 95.66\% \\
Retrieval & Mean token error (tokens) & 0.57 \\
\hline
Token flow & EPE (tokens) & 0.62 \\
Token flow & Direction cosine & 0.331 \\
Token flow & PCK@1 token & 83.35\% \\
\hline
PCA & PC1 explained variance & 14.1\% \\
PCA & PC2 explained variance & 7.1\% \\
PCA & PC3 explained variance & 5.2\% \\
PCA & Top-3 cumulative variance & 26.4\% \\
\hline
\end{tabular}
\end{table}

Before discussing the numbers, it is useful to clarify how they should be interpreted. Similarity and retrieval are the strongest probes in this section because they make the fewest modeling assumptions. They only ask whether true cross-frame matches are already nearby in frozen token space. Token flow is a harder probe because it asks nearest-neighbor retrieval to behave like a motion estimator. PCA is the most indirect probe because a failure to find a low-dimensional linear motion axis does not mean that motion information is absent. It only means that the information is not cleanly isolated in the top principal components.

Figure~\ref{fig:da3-frozen-token-motion} and Table~\ref{tab:da3-frozen-token-motion} provide direct evidence that frozen DA3 tokens preserve cross-frame correspondence structure. The similarity distributions are cleanly separated, with ground-truth matches consistently closer in token space than random negatives. Nearest-neighbor retrieval leads to the same conclusion under a stronger all-candidates test. These two probes show that, even without any learned motion decoder, the frozen representation already places true future matches near the source token and usually retrieves a target very close to the correct location. 

The token-flow probe refines this picture. A simple nearest-neighbor displacement readout already recovers meaningful motion, which indicates that the frozen tokens encode where points are likely to move. At the same time, this probe is weaker than similarity and retrieval, which suggests that the motion signal is not exposed as a trivial one-step velocity code. This is precisely where a structured decoder becomes useful. Our DPT head can pool multiscale context and impose spatial coherence to transform this latent correspondence signal into dense scene flow.

The PCA results should be read as auxiliary evidence. The dominant components explain only a limited portion of the token-difference variance, and their alignment with flow magnitude remains weak. This does not argue against motion information in the frozen tokens. Instead, PCA indicates that motion is encoded in a distributed high-dimensional form rather than concentrated in a single linear direction.

These results support a precise interpretation of our method. Frozen DA3 encoder tokens already contain scene-flow-relevant information, especially in the form of cross-frame correspondence structure. The strongest evidence comes from the probes that require almost no decoder assumptions, namely similarity separation and nearest-neighbor retrieval. The weaker token-flow and PCA probes refine this picture. They suggest that the motion signal is real, but distributed and not trivially linearized. Our DPT head succeeds because it decodes and regularizes this latent signal into dense metric predictions. In other words, the scene flow branch acts as an informed readout of the pretrained geometry representation, rather than a separate module that has to create motion information from scratch.

\section{Training Details}
\label{sec:app_train_details}

\subsection{Dataset and Optimization Summary}
\label{subsec:app_dataset_summary}

All dataset statistics in this paper correspond to the 81-frame manifests. Every manifest row therefore satisfies the temporal constraint $T=4k+1$ with $T=81$. \Cref{tab:app_training_datasets} summarizes the stage-wise training data. Stage 1 trains only the DA3 scene-flow readout. Stage 2 finetunes the stitched decoder with hybrid ground-truth and teacher-student supervision. Stage 3 optimizes the action-conditioned video model, where Action100M contributes text-conditioned video clips without paired 4D labels.
\begin{table*}[t]
\centering
\small
\renewcommand{\arraystretch}{1.15}
\setlength{\tabcolsep}{10pt}
\caption{Stage-wise training datasets for the 81-frame \shortname{} pipeline. Counts are the effective rows used in each stage. Stage~1 keeps only clips with \texttt{scene\_flow\_available=true}, while Stage~2 uses the same GT-capable manifests to build a GT branch and a distillation branch, so one hybrid epoch traverses the full mixture once under GT supervision and once under teacher-student supervision.}
\label{tab:app_training_datasets}
\begin{tabular}{@{}clrl@{}}
\toprule
\textbf{Stage} & \textbf{Dataset} & \textbf{Clips} & \textbf{Primary Supervision} \\
\midrule
1 & PointOdyssey train      &   2{,}619 & GT scene flow \\
1 & Dynamic Replica train   &   3{,}864 & GT scene flow \\
1 & BlinkVision train       &       223 & GT scene flow \\
1 & Virtual KITTI 2 train   &       580 & GT scene flow \\
1 & OmniWorld train         &   4{,}670 & GT scene flow \\
\midrule
\multirow{7}{*}{2}
  & PointOdyssey            &   3{,}040 & GT geometry + flow \\
  & BlinkVision             &       241 & GT geometry + flow \\
  & Dynamic Replica         &   3{,}864 & GT geometry + flow \\
  & Virtual KITTI 2         &       580 & GT geometry + flow \\
  & OmniWorld-Game          &   5{,}796 & GT geometry + flow \\
  & MVS-Synth               &       240 & GT geometry only \\
  & ScanNet++               & 108{,}464 & GT geometry only \\
\midrule
3 & Action100M              & 249{,}422 & Distill only \\
\bottomrule
\end{tabular}
\end{table*}
Stage 1 mixes 11,956 training clips from five dynamic datasets. The sampler uses dataset-wise mixture weights of 1.0 together with a size-aware mixing exponent $\alpha=0.5$, which interpolates between uniform-over-datasets and proportional-to-size sampling. The stage-1 manifest directory also contains held-out or diagnostic subsets such as PointOdyssey test, PointOdyssey sample, and Dynamic Replica valid clips for validation and smoke tests.

Stage 2 contains 122,225 GT-capable training clips. Among them, 13,521 clips expose scene-flow labels, while the remaining 108,704 clips provide only depth, ray, camera, and point-cloud supervision. In hybrid decoder training, we build separate GT and distill dataloaders from the same manifest list and randomly shuffle the union of GT and distill steps once per epoch. This realizes one GT pass and one distillation pass over the full mixture in a single hybrid epoch.

The stage-3 Action100M subset is constructed in two steps. We first start from the filtered 81-frame TI2V manifest \texttt{action100m\_train\_ti2v\_81f.filtered.jsonl}, whose rows already satisfy \texttt{task\_type=ti2v}, \texttt{cond\_frames=1}, and \texttt{num\_frames=81}. We then score the corresponding clip-level entries from \texttt{action100m\_clip\_index\_81f.filtered.jsonl} and keep only clips that pass a high-precision action-quality filter. The clip-level decision uses fields such as \texttt{prompt\_action}, \texttt{level}, and \texttt{node\_duration\_sec}. We drop clips with node duration shorter than 1.6 seconds, hierarchy depth above 10, missing action text, action text identical to the generic description, action text shorter than four words, prompt patterns dominated by on-screen text or audio-style narration, uncertain descriptions, or generic action stubs. The retained manifest \texttt{action100m\_train\_ti2v\_81f.action\_hq.keep.jsonl} contains 249,422 clips, which is 28.6\% of the 871,785 filtered TI2V rows. Because Action100M does not provide paired 4D geometry, these clips are used only in the distillation branch during stage 3.

\paragraph{Per-dataset roles.}
PointOdyssey provides articulated outdoor motion with sparse track supervision that we rasterize into dense forward scene flow, which makes it particularly useful for supervising correspondence-aware motion. Dynamic Replica contributes denser dynamic stereo geometry with strong camera and scene-flow signals. BlinkVision, Virtual KITTI 2, and OmniWorld broaden the synthetic motion distribution across indoor scenes, driving scenes, and open-world gameplay trajectories. ScanNet++ and MVS-Synth do not provide scene flow in our 81-frame setup, but they substantially enrich multiview depth, camera, and point-cloud supervision during stitched-decoder finetuning. Action100M serves a different purpose from the geometry datasets. It injects large-scale action-conditioned video semantics into the joint model while relying on the frozen DA3 teacher for geometric targets.

\paragraph{Evaluation datasets.}
We report three groups of held-out evaluations. For dynamic 4D motion, we evaluate on MOVi-F, Dynamic Replica, and PointOdyssey using 81-frame clips. MOVi-F tests controlled synthetic object motion, Dynamic Replica tests realistic dynamic stereo scenes, and PointOdyssey emphasizes articulated motion with long temporal support. For stitched-decoder geometry preservation, we evaluate on ETH3D, ScanNet++, Dynamic Replica, Dynamic-RE10K, MOVi-F, and RobbySimVal. These datasets cover static reconstruction, indoor multiview geometry, dynamic camera estimation, and robot-centric depth prediction. For action-conditioned world generation, we evaluate on RLBench with 210 samples, HOIGen-1M with 200 samples, Dynamic Replica with 480 samples, and PointOdyssey with 375 samples, which together span manipulation, hand-object interaction, and general dynamic-scene futures. We additionally evaluate 100 five-second Epic-Kitchens-100 clips to assess in-the-wild generalization to everyday human-object interactions in kitchen environments. This evaluation complements HoiGen with a second real-world human-object interaction benchmark.

\paragraph{Stage-3 comparison scope.}
We exclude several recent world-generation systems from the stage-3 comparison because they assume additional control inputs beyond the setting targeted in this paper. PointWorld requires depth input \citep{huang2026pointworld}. FantasyWorld, HY-World 2.0, Lyra 2.0, Geometry Forcing, and TeleWorld rely on trajectory-level control signals \citep{dai2025fantasyworld,hy2026hy,shen2026lyra,wu2025geometry,chen2025teleworld}. Kinema4D further requires robot-action kinematic representations \citep{xu2026kinema4d}. These methods therefore do not match the task of predicting future RGB video together with dense 4D geometry from only an observed scene context and a text-specified action, so we do not include them in the stage-3 comparison.

\paragraph{Evaluation protocol and metric definitions.}
We follow the same three-way evaluation split as in the main paper and expand the scoring protocol here. \textbf{Scene flow.} We score only valid moving pixels indicated by the deformation mask. Predictions are first rescaled to the target point-map scale of each clip. When a model predicts relative flow, we restore the target scale with the clip-level scene-flow scale before metric computation. If a benchmark provides native-resolution labels, we resize predictions to that evaluation resolution before scoring. We report
\begin{equation}
\mathrm{EPE}
=
\frac{1}{|\mathcal{V}_{\mathrm{sf}}|}
\sum_{(t,p)\in\mathcal{V}_{\mathrm{sf}}}
\left\|
\hat{\mathbf{m}}_{t,p}-\mathbf{m}_{t,p}
\right\|_2,
\qquad
APD_{0.05}
=
\frac{1}{|\mathcal{V}_{\mathrm{sf}}|}
\sum_{(t,p)\in\mathcal{V}_{\mathrm{sf}}}
\mathbb{1}
\!\left[
\left\|
\hat{\mathbf{m}}_{t,p}-\mathbf{m}_{t,p}
\right\|_2
<
0.05
\right],
\end{equation}
where $\mathcal{V}_{\mathrm{sf}}$ is the set of valid motion pixels, $\hat{\mathbf{m}}_{t,p}$ is the predicted 3D displacement from frame $t$ to frame $t+1$, and $\mathbf{m}_{t,p}$ is the ground-truth displacement. Lower EPE is better, while higher $APD_{0.05}$ is better.

\textbf{Depth.} Depth metrics use per-clip scale-shift alignment. For each clip, we solve one scale $a$ and one shift $b$ jointly over all valid depth pixels and evaluate $a\hat{d}+b$ against ground-truth depth:
\begin{equation}
(a^\star,b^\star)
=
\arg\min_{a,b}
\sum_{(t,p)\in\mathcal{V}_{d}}
\left(
a\hat{d}_{t,p}+b-d_{t,p}
\right)^2.
\end{equation}
We report
\begin{equation}
\mathrm{AbsRel}
=
\frac{1}{|\mathcal{V}_{d}|}
\sum_{(t,p)\in\mathcal{V}_{d}}
\frac{
\left|
a^\star\hat{d}_{t,p}+b^\star-d_{t,p}
\right|
}{
d_{t,p}
}
\end{equation}
\begin{equation}
\delta_1
=
\frac{1}{|\mathcal{V}_{d}|}
\sum_{(t,p)\in\mathcal{V}_{d}}
\mathbb{1}
\!\left[
\max\!\left(
\frac{a^\star\hat{d}_{t,p}+b^\star}{d_{t,p}},
\frac{d_{t,p}}{a^\star\hat{d}_{t,p}+b^\star}
\right)
<
1.25
\right].
\end{equation}
Lower AbsRel is better, while higher $\delta_1$ is better.

\textbf{Camera.} Camera metrics follow the official DA3 bench. Predicted and ground-truth camera trajectories are first converted to the first-camera gauge. We then evaluate all frame pairs $(i,j)$ through their relative transforms. Let $e^{r}_{ij}$ denote the relative rotation error in degrees and let $e^{t}_{ij}$ denote the angle between the predicted and ground-truth relative translation directions. On Dynamic Replica we report
\begin{equation}
RRA@5^\circ
=
\frac{1}{|\mathcal{P}|}
\sum_{(i,j)\in\mathcal{P}}
\mathbb{1}\!\left[e^{r}_{ij}<5^\circ\right],
\end{equation}
which measures relative rotation accuracy over all frame pairs. On Dynamic-RE10K we report $AUC_{15}$ and $AUC_{30}$. These metrics are computed from the per-pair error
\begin{equation}
m_{ij}=\max\!\left(e^{r}_{ij}, e^{t}_{ij}\right),
\end{equation}
and summarize the cumulative accuracy curve of $m_{ij}$ up to the corresponding angular threshold. Higher camera metrics are better.

\textbf{3D reconstruction.} Reconstruction also follows the official DA3 bench. We estimate a global Umeyama alignment between predicted and ground-truth camera trajectories and use the recovered scale to place predicted depth in metric units before TSDF fusion. When predicted poses are used for reconstruction, the same fit also aligns the fused camera trajectory. The fused prediction is then compared with the benchmark ground-truth geometry using the dataset-specific threshold and downsampling settings from the DA3 bench. ETH3D and ScanNet++ evaluate against points sampled from the ground-truth mesh, and ScanNet++ additionally crops the predicted point cloud to the ground-truth bounding box with a small margin before scoring. For predicted points $P$ and ground-truth points $G$, we report
\begin{equation}
\mathrm{Acc.}
=
\frac{1}{|P|}
\sum_{x\in P}
\min_{y\in G}
\|x-y\|_2,
\qquad
\mathrm{Comp.}
=
\frac{1}{|G|}
\sum_{y\in G}
\min_{x\in P}
\|x-y\|_2.
\end{equation}
Precision and recall are the fractions of predicted and ground-truth points whose nearest-neighbor distance is below the dataset threshold $\epsilon$, and the F-score is their harmonic mean:
\begin{equation}
\mathrm{Prec.}
=
\frac{1}{|P|}
\sum_{x\in P}
\mathbb{1}
\!\left[
\min_{y\in G}\|x-y\|_2<\epsilon
\right],
\end{equation}
\begin{equation}
\mathrm{Rec.}
=
\frac{1}{|G|}
\sum_{y\in G}
\mathbb{1}
\!\left[
\min_{x\in P}\|x-y\|_2<\epsilon
\right],
\end{equation}
\begin{equation}
\mathrm{F\text{-}score}
=
\frac{2\,\mathrm{Prec.}\,\mathrm{Rec.}}
{\mathrm{Prec.}+\mathrm{Rec.}}.
\end{equation}
Lower Acc. and Comp. are better, while higher F-score is better.

\textbf{Joint world generation.}
For RGB video quality, higher video-caption similarity indicates stronger alignment between the generated video and the conditioning action text. RLBench and HOIGen-1M use their native action annotations, while Dynamic Replica and PointOdyssey do not provide action-level text in our stage-3 evaluation split, so we generate per-clip action captions with Qwen-VL-30B-Thinking and use them as the conditioning text for all compared methods. When a baseline exposes a compatible geometry output, we score its generated 4D structure with the same geometry protocol described above.

\paragraph{Epic-Kitchens-100 evaluation setting.}
We sample 100 clips, each five seconds long, and follow the same action-conditioned joint 4D generation protocol as the original comparison. Each method receives the initial RGB frame and an action description. We compare \shortname, TesserAct, Gen3R, and VIST3A using the same metric definitions and alignment procedures. The full evaluation reports six video metrics, namely subject consistency (SC), background consistency (BC), aesthetic quality (AQ), image quality (IQ), motion smoothness (MS), and video-caption similarity (Cap.Sim). Geometry is evaluated with depth AbsRel, $\delta_1$, $\delta_2$, and Chamfer-L1, together with camera RPE-R, RRA@10, and RRA@5. Higher scores are better except for AbsRel, Chamfer-L1, and RPE-R. The complete results appear in \Cref{tab:main_results}, and \Cref{tab:main-results-subset} retains six representative metrics for comparison with the other benchmarks.

\paragraph{Stage-wise optimization details.}
For stage 1, the 81-frame manifest configuration uses AdamW with learning rate $10^{-5}$, weight decay $10^{-2}$, $\beta=(0.9,0.999)$, 100 warmup steps, gradient accumulation 2, and bf16 mixed precision. Inputs are resized to $504\times504$ with upper-bound resizing, the clip overlap is 1 frame, and the scene-flow loss adds a zero-last-frame penalty with weight 0.1. The main paper reports a 20-epoch run on 16 H100 GPUs with per-GPU batch size 1.

For stage 2, the stitched-decoder hybrid launchers use AdamW with learning rate $10^{-4}$, weight decay $10^{-4}$, bf16 mixed precision, $504\times504$ inputs, and per-GPU batch size 1. The representative 81-frame launchers use a 50-step warmup and train the stitching layer together with LoRA adapters on the stitched DA3 blocks by default, while still supporting full finetuning of the stitched parts. The main paper reports one hybrid epoch on 64 H20 GPUs.

For stage 3, the stitched joint LoRA launcher uses AdamW with learning rate $10^{-4}$, cosine decay, 1000 warmup steps, LoRA rank 32 on the DiT, gradient checkpointing, 81 frames at $532\times532$ resolution, condition dropout 0.1, and geometry-loss warmup from 0.001 to 1.0 over the first 500 steps. We apply \texttt{mask\_condition\_loss} on conditioned prefixes and \texttt{geo\_supervise\_generated\_only} on the generated suffix. The main paper reports one epoch on 64 H20 GPUs with per-GPU batch size 1.

\subsection{Scene Flow Head Training}
\label{subsec:scene_head_training}
We train the scene-flow branch as a readout on top of a fixed DA3 representation. The DA3 backbone, depth head, ray head, and camera modules are loaded from the pretrained checkpoint and kept frozen. We inject the DPT-style scene-flow head and its coordinate adapter from the scene-flow preset, load the original DA3 weights with non-strict matching, and optimize only the parameters of the new head. During training, each clip is passed through the frozen DA3 encoder to extract the four multiscale token maps. The head decodes these tokens into a dense three-channel motion field, and the adapter rotates the local camera-frame prediction into the first-frame camera basis.

The training target is a forward displacement field from frame $t$ to frame $t+1$. All datasets expose the same fields: RGB frames, camera intrinsics, world-to-camera extrinsics, a scene-flow tensor, a per-clip scale, and a binary deformation mask. We use relative scene flow by default. Metric displacements are expressed in the first-frame camera basis and normalized by a robust clip-level scale, which makes the head less sensitive to the absolute size of a scene.

Let $\hat{\mathbf{m}}_{t,p}$ be the predicted motion at pixel $p$ in frame $t$, let $\mathbf{m}_{t,p}$ be the target, and let $a_{t,p}\in\{0,1\}$ be the deformation mask. The scene-flow loss~\citep{li2021neural, teed2021raft} is
\begin{equation}
\mathcal{L}_{\mathrm{sf}}
=
\frac{
\sum_{t,p} a_{t,p}
\left\|
\hat{\mathbf{m}}_{t,p} - \mathbf{m}_{t,p}
\right\|_1
}{
\max\left(1,\sum_{t,p} a_{t,p}\right)
}
+
\lambda_{\mathrm{last}}
\frac{1}{HW}
\sum_p
\left\|
\hat{\mathbf{m}}_{T,p}
\right\|_2 .
\end{equation}
The first term is evaluated only at valid motion pixels. It never falls back to full-image supervision, which is important for datasets where motion labels come from sparse tracks. The second term softly enforces the last frame to be zero because there is no target frame $T+1$. We use masked L1 loss in the main configuration. 

\subsection{Latent-to-Geometry Model Stitching Training}
\label{subsec:transfer_finding}
The stitching stage determines where the video latent space should enter DA3 and initializes the projection into that token space. For each training clip, we encode the RGB video with the frozen Wan VAE and extract candidate DA3 token maps from encoder blocks. The Wan latent has 16 channels and is temporally compressed. Before fitting the stitch, we upsample it back to the video frame rate with
\(
T = 4(T_{\mathrm{vae}}-1)+1
\)
and resize it to the spatial resolution required by the DA3 patch grid.

For each candidate block $k$, we fit a 3D convolutional stitch $\mathcal{S}_{\theta}^{(k)}$ from the upsampled Wan latent to the frozen DA3 token grid. The fit is a streaming ridge regression problem shown in~\Cref{equ:linear_fitting}.

The streaming solver accumulates normal-equation statistics over cached tensor batches, which avoids storing a large design matrix. We rank candidate blocks by the reconstruction mean squared error and choose the block with the lowest error. In our DA3 setting, the stitch maps the 16-channel Wan latent to the 1536-channel DA3 token width through a compact spatiotemporal convolution. This procedure gives both the injection layer and the initial weights for the later trainable stitched decoder.

\subsection{Stitched Decoder Finetuning}
\label{subsec:stiching_training}
The linear stitch is only an initialization. We next train a stitched decoder that accepts Wan latents but preserves the geometry behavior of DA3. The Wan VAE is frozen. The pretrained DA3 teacher is frozen. The trainable student contains the stitching module and stitched DA3 parts. The video diffusion transformer is not trained in this stage.

We use two complementary sources of supervision. In the distillation branch, the teacher consumes the ground-truth RGB clip, while the student consumes the Wan latent of the same clip. The teacher and student outputs are aligned in time and space before computing losses. The distillation objective is
\begin{equation}
\begin{aligned}
\mathcal{L}_{\mathrm{distill}}
=&
\lambda_D
\left\|
\hat{\mathbf{D}}-\mathbf{D}^{\mathrm{T}}
\right\|_1
+
\lambda_{\nabla D}
\mathcal{L}_{\nabla D}
\left(
\hat{\mathbf{D}},\mathbf{D}^{\mathrm{T}}
\right)
+
\lambda_R
\left\|
\hat{\mathbf{R}}-\mathbf{R}^{\mathrm{T}}
\right\|_1
\\
&
+
\lambda_{cD}
\left\|
\hat{\mathbf{C}}^D-\mathbf{C}^{D,\mathrm{T}}
\right\|_1
+
\lambda_{cR}
\left\|
\hat{\mathbf{C}}^R-\mathbf{C}^{R,\mathrm{T}}
\right\|_1
+
\lambda_C
\mathcal{L}_{C}
\\
&
+
\lambda_M
\left\|
\hat{\mathbf{M}}-\mathbf{M}^{\mathrm{T}}
\right\|_1
+
\lambda_T
\mathcal{L}_{T}.
\end{aligned}
\end{equation}
Here $\mathbf{D}$ denotes depth, $\mathbf{R}$ denotes the six-channel ray map, $\mathbf{C}^D$ and $\mathbf{C}^R$ denote confidence maps, and $\mathbf{M}$ denotes scene flow. The camera term $\mathcal{L}_C$ compares decoded intrinsics and relative camera motion. The temporal term $\mathcal{L}_T$ matches depth differences between adjacent frames, which discourages frame-wise geometry flicker.

In the ground-truth branch, the student is supervised against dataset depth, ray, camera, valid-mask, and optional scene-flow annotations. The objective is
\begin{equation}
\begin{aligned}
\mathcal{L}_{\mathrm{GT}}
=&
\lambda_D
\mathcal{L}_{D}
+
\lambda_{\nabla D}
\mathcal{L}_{\nabla D}
+
\lambda_R
\mathcal{L}_{R}
+
\lambda_P
\mathcal{L}_{P}
\\
&
+
\lambda_C
\mathcal{L}_{C}
+
\lambda_M
\mathcal{L}_{M}
+
\lambda_T
\mathcal{L}_{T}.
\end{aligned}
\end{equation}
The depth loss follows DA3's confidence-aware form. For sample $b$, let $\gamma_b$ be the common scene scale computed from the valid target point map, and let $\Omega_b$ be the valid pixel set over all frames. The depth prediction is first aligned to the target camera scale, producing $\tilde{D}$. We then use
\begin{equation}
\mathcal{L}_{D}
=
\frac{1}{B}
\sum_b
\frac{1}{\gamma_b}
\frac{1}{|\Omega_b|}
\sum_{(t,p)\in\Omega_b}
\left[
\hat{c}^{D}_{b,t,p}
\left|
\tilde{D}_{b,t,p}-D^\star_{b,t,p}
\right|
-
\beta_D
\log \hat{c}^{D}_{b,t,p}
\right]_+ ,
\end{equation}
where $\hat{c}^{D}\geq 1$ is the predicted depth confidence and $[\cdot]_+$ clamps the per-pixel contribution to be non-negative. The confidence term allows the model to express uncertainty, while the logarithmic regularizer prevents the trivial low-confidence solution.

The depth-gradient loss supervises local geometry changes rather than absolute depth alone,
\begin{equation}
\mathcal{L}_{\nabla D}
=
\frac{1}{B}
\sum_b
\frac{1}{\gamma_b}
\frac{1}{|\Omega_b^{x}|+|\Omega_b^{y}|}
\left(
\sum_{(t,p)\in\Omega_b^{x}}
\left|
\nabla_x\tilde{D}_{b,t,p}
-
\nabla_xD^\star_{b,t,p}
\right|
+
\sum_{(t,p)\in\Omega_b^{y}}
\left|
\nabla_y\tilde{D}_{b,t,p}
-
\nabla_yD^\star_{b,t,p}
\right|
\right),
\end{equation}
where $\Omega_b^{x}$ and $\Omega_b^{y}$ contain neighboring valid pixels along the horizontal and vertical directions. This term preserves depth discontinuities and discourages over-smoothed surfaces.

Each ray map is represented as a direction-origin pair
\(
\mathbf{R}_{t,p} =
\left(
\mathbf{d}_{t,p},\mathbf{o}_{t,p}
\right)
\),
where $\mathbf{d}_{t,p}\in\mathbb{R}^3$ is a unit ray direction and $\mathbf{o}_{t,p}\in\mathbb{R}^3$ is the camera-origin field. Before supervision, the predicted ray map is aligned to the official DA3 camera-pose space. The ray loss is
\begin{equation}
\mathcal{L}_{R}
=
\frac{1}{B}
\sum_b
\left[
\frac{1}{|\Omega_b|}
\sum_{(t,p)\in\Omega_b}
\left\|
\tilde{\mathbf{d}}_{b,t,p}
-
\mathbf{d}^{\star}_{b,t,p}
\right\|_1
+
\lambda_o
\frac{1}{|\Omega_b|}
\sum_{(t,p)\in\Omega_b}
\left\|
\tilde{\mathbf{o}}_{b,t,p}
-
\mathbf{o}^{\star}_{b,t,p}
\right\|_1
\right]
\end{equation}
The direction term controls viewing geometry, while the origin term keeps the predicted camera-origin field at the correct scale.

The point-cloud loss turns depth and ray geometry into explicit 3D points. In ray-map form, each valid pixel produces
\begin{equation}
\mathbf{P}_{b,t,p}
=
\mathbf{o}_{b,t,p}
+
D_{b,t,p}\mathbf{d}_{b,t,p}.
\end{equation}
Let $\hat{\mathcal{P}}_b$ and $\mathcal{P}^{\star}_b$ be the resulting valid point sets sampled from all frames. We use a symmetric dense point-set distance,
\begin{equation}
\mathcal{L}_{P}
=
\frac{1}{B}
\sum_b
\frac{1}{2\gamma_b}
\left[
\frac{1}{|\hat{\mathcal{P}}_b|}
\sum_{\hat{\mathbf{p}}\in\hat{\mathcal{P}}_b}
\min_{\mathbf{q}\in\mathcal{P}^{\star}_b}
\left\|
\hat{\mathbf{p}}-\mathbf{q}
\right\|_2
+
\frac{1}{|\mathcal{P}^{\star}_b|}
\sum_{\mathbf{q}\in\mathcal{P}^{\star}_b}
\min_{\hat{\mathbf{p}}\in\hat{\mathcal{P}}_b}
\left\|
\hat{\mathbf{p}}-\mathbf{q}
\right\|_2
\right].
\end{equation}
This term is a training-time surrogate for reconstruction quality. It penalizes global 3D misalignment that can be hidden if depth and ray maps are considered independently.

The camera loss is computed after converting both predicted and target poses to a first-camera-relative gauge. Let $\bar{\mathbf{E}}_{b,t}$ be the aligned world-to-camera extrinsic, let $\bar{\mathbf{K}}_{b,t}$ be the image-size-normalized intrinsic matrix, and let $\mathbf{u}_{b,t}$ be the unit translation direction. We use
\begin{equation}
\mathcal{L}_{C}
=
\frac{1}{B}
\sum_b
\frac{1}{|\mathcal{T}_b|}
\sum_{t\in\mathcal{T}_b}
\left(
\left\|
\hat{\bar{\mathbf{E}}}^{\,R}_{b,t}
-
\bar{\mathbf{E}}^{\star,R}_{b,t}
\right\|_1
+
1
-
\left\langle
\hat{\mathbf{u}}_{b,t},
\mathbf{u}^{\star}_{b,t}
\right\rangle
+
\left\|
\hat{\bar{\mathbf{K}}}_{b,t}
-
\bar{\mathbf{K}}^{\star}_{b,t}
\right\|_1
\right),
\end{equation}
where $\mathcal{T}_b$ denotes frames with valid camera estimates and $\bar{\mathbf{E}}^{R}$ denotes the rotation block. 

The temporal depth term matches depth changes across adjacent valid frames~\citep{chen2025video},
\begin{equation}
\mathcal{L}_{T}
=
\frac{1}{B}
\sum_b
\frac{1}{\gamma_b}
\frac{1}{|\Psi_b|}
\sum_{(t,p)\in\Psi_b}
\left|
\left(
\tilde{D}_{b,t+1,p}
-
\tilde{D}_{b,t,p}
\right)
-
\left(
D^\star_{b,t+1,p}
-
D^\star_{b,t,p}
\right)
\right|,
\end{equation}
where $\Psi_b$ contains adjacent frame-pixel pairs that are valid in both frames and pass the depth-change filter. This term prevents the stitched decoder from producing temporally plausible RGB-aligned depth in each frame but inconsistent depth evolution across the clip.

The motion term $\mathcal{L}_{M}$ uses the same masked scene-flow loss as \Cref{subsec:scene_head_training}. For GT datasets without scene-flow annotations, including ScanNet++ and MVS-Synth in our 81-frame stage-2 manifests, we set $\mathcal{L}_{M}=0$ and keep the remaining geometry losses active.

\subsection{Joint Video and 4D Geometry Training}
\label{subsec:joint_training}
After the stitched decoder is finetuned, we freeze it and use it as a differentiable geometry critic for LongCat-Video. The only trainable video parameters in our main joint run are LoRA parameters on the DiT. The training batch first follows standard latent flow matching. Given a clean latent $\mathbf{z}_0$, noise $\boldsymbol{\epsilon}$, and sampled noise level $\tau$, the scheduler constructs a noisy latent $\mathbf{z}_{\tau}$ and a target velocity $\mathbf{v}^{\star}_{\tau}$. The diffusion loss is a masked weighted mean squared error,
\begin{equation}
\mathcal{L}_{\mathrm{FM}}
=
w(\tau)
\frac{
\sum_j b_j
\left\|
\mathbf{v}_{\phi,j}
-
\mathbf{v}^{\star}_{\tau,j}
\right\|_2^2
}{
\max\left(10^{-8},\sum_j b_j\right)
},
\end{equation}
where $b_j$ is the latent loss mask. For image-to-video training, the first frame is used as the conditioning prefix. For continuation training, the observed prefix is encoded as LongCat condition latents. When condition masking is enabled, the prefix receives zero diffusion loss.

The predicted velocity also gives a denoised latent estimate,
\begin{equation}
\hat{\mathbf{z}}_0
=
\mathbf{z}_{\tau}
-
\sigma_{\tau}
\mathbf{v}_{\phi}
\left(
\mathbf{z}_{\tau},\tau,c
\right).
\end{equation}
We pass $\hat{\mathbf{z}}_0$ directly into the frozen stitched decoder and compute a geometry loss on the decoded 4D outputs. The joint objective is shown in~\Cref{equ:joint}.

For action-conditioned joint training, the teacher-student distillation branch is always active and supplies the only supervision available for Action100M clips. A raw DA3 teacher is loaded independently from the stitched decoder and is kept frozen. The raw teacher processes the ground-truth RGB video and supplies depth, ray, confidence, camera, and scene-flow targets. The frozen stitched decoder processes the model-predicted latent $\hat{\mathbf{z}}_0$. The same distillation loss used in decoder finetuning is then applied to the two geometry outputs.

In the manifest-based joint setup, GT-capable geometry datasets can be used with \texttt{distill}, \texttt{gt}, or \texttt{hybrid} supervision modes. Action100M rows are wrapped as distill-only manifest entries with prompt metadata and no GT geometry. For the subset used in the main paper, every Action100M row is an 81-frame TI2V clip with \texttt{cond\_frames=1}. This design lets the video model learn long-horizon action semantics from large-scale web video while still receiving metric 4D regularization from the stitched geometry branch.


\subsection{Model Information}
\label{subsec:model_information}
Our system combines a large pretrained video generator with a frozen geometric readout. The dominant parameter budget comes from the LongCat-Video diffusion transformer, while the Wan VAE provides the latent video interface used by both the RGB and geometry branches. Based on the local pretrained weights used in our pipeline, the LongCat-Video DiT contains 13{,}581.07M parameters, and the Wan VAE contains 126.89M parameters. These two components are both frozen during stitched-decoder finetuning, and only the DiT is updated during the final joint training stage, where our main setting optimizes LoRA parameters on top of the pretrained backbone.

The geometry branch is implemented as a stitched DA3 decoder that reads the same Wan latent trajectory and recovers depth, camera rays, camera parameters, and scene flow. This branch reuses the pretrained DA3 backbone and dense geometry heads together with our added scene-flow head and the latent-to-token stitching module. We do not count the raw DA3 teacher separately because it is only used to provide supervision during training and is not part of the inference-time Plan4D pipeline. In this sense, the deployed system consists of three principal model components: a 13.58B video diffusion transformer, a 126.89M video VAE, and a stitched geometric decoder built on top of DA3. The overall parameter budget is therefore dominated by the video backbone rather than the geometric readout.


\section{Other Ablation Studies}
\subsection{Ablation of the Joint Training Objective}
\label{subsec:app_joint_objective_ablation}

We ablate the final joint-training objective and evaluate each variant on HoiGen and Dynamic Replica under the action-conditioned world-generation protocol in \Cref{sec:exp_settings}. Each test example provides an initial RGB frame and an action description. HoiGen uses its native action annotations, while Dynamic Replica uses the per-clip action captions from the main evaluation. Each variant produces RGB video and dense geometry from its predicted latent trajectory using the frozen stitched decoder described in \Cref{subsec:joint_training}.

The variants isolate the terms of \Cref{equ:joint}. \textbf{Only-RGB-Loss} optimizes the latent flow-matching loss $\mathcal{L}_{\mathrm{FM}}$ without the geometry term. \textbf{Only-Geo-Loss} optimizes the weighted geometric loss $\lambda_{\mathrm{geo}}\mathcal{L}_{\mathrm{geo}}$ without the flow-matching term. \textbf{Ours (Full)} combines both losses to train the video generator while keeping the stitched geometry decoder frozen. When geometric supervision is used, its targets come from the frozen DA3 teacher described in \Cref{subsec:joint_training}.

For both datasets, we report subject consistency, background consistency, depth AbsRel, depth $\delta_1$, and camera rotation RPE$_R$ in \Cref{tab:ablation}. Higher consistency scores and $\delta_1$ are better, while lower AbsRel and RPE$_R$ are better. Depth uses per-clip scale-shift alignment, and camera evaluation uses the same first-camera alignment and relative-rotation protocol as the main joint-generation comparison.

On Dynamic Replica, the full objective achieves the best result on all five metrics. Compared with RGB-only training, subject and background consistency increase from 0.8998 and 0.9321 to 0.9356 and 0.9496. Depth AbsRel decreases from 0.0874 to 0.0696, $\delta_1$ increases from 0.9169 to 0.9306, and camera RPE$_R$ decreases from 0.2615 to 0.1550. Geometry-only training performs worse on all reported metrics than both RGB-only training and the full objective, indicating that geometric supervision alone does not preserve visual consistency or ensure accurate geometry on this benchmark.

The HoiGen results reveal a different depth tradeoff. Geometry-only training attains the best depth scores, but weakens subject and background consistency and has a higher camera rotation error than the full objective. Across both datasets, the full model gives the best subject and background consistency and camera estimates. These results support the complementary roles of the two losses: flow matching preserves action-conditioned visual quality, while geometric supervision regularizes the shared future latent state toward consistent depth and camera structure.

\subsection{Ablation for Stitching Adapter Architecture}
\label{subsec:app_stitching_adapter_architecture}

The stitching adapter has a narrow but important role in \shortname. It maps the upsampled Wan VAE latent into the DA3 token space at a candidate encoder block, after which the remaining DA3 encoder and dense heads perform the geometric decoding. We therefore ask whether this interface needs a high-capacity nonlinear module, or whether the compact linear 3D convolution used in the main method is sufficient.

We compare three adapter architectures under the same cached OmniWorld-Game~\citep{zhou2025omniworld} feature sweep. Let
$\mathbf{Z}\in\mathbb{R}^{B\times 16\times T\times H\times W}$ denote the temporally upsampled Wan latent, and let
$\mathbf{Y}\in\mathbb{R}^{B\times 1536\times T\times H'\times W'}$ denote the target DA3 token features. The adapter learns a mapping
\begin{equation}
    \mathcal{A}_{\theta}:\mathbf{Z}\mapsto \hat{\mathbf{Y}},
    \qquad
    \hat{\mathbf{Y}}\in\mathbb{R}^{B\times 1536\times T\times H'\times W'} .
\end{equation}
We study three designs with increasing modeling capacity. The linear adapter provides a purely local projection, the residual bottleneck augments this projection with a nonlinear correction, and the axial-attention adapter further introduces long-range mixing along the temporal and spatial axes.

\paragraph{Linear convolutional adapter.}
The first variant uses a single 3D convolution to map the 16-channel Wan latent directly into the 1536-channel DA3 token space:
\begin{equation}
    \hat{\mathbf{Y}}
    =
    \mathcal{A}_{\theta}^{\mathrm{lin}}(\mathbf{Z})
    =
    \operatorname{Conv3D}_{5\times 3\times 3,\;1\times 2\times 2,\;2\times 1\times 1}^{16\rightarrow 1536}
    (\mathbf{Z}) .
\end{equation}
Equivalently, after unfolding the local spatiotemporal neighborhoods, this model is a linear regression problem:
\begin{equation}
    \operatorname{vec}(\hat{\mathbf{Y}})
    =
    \mathbf{\Phi}(\mathbf{Z})\mathbf{w},
\end{equation}
where $\mathbf{\Phi}(\mathbf{Z})$ is the design matrix induced by the convolutional receptive field and $\mathbf{w}$ contains the convolutional weights. We can fit this adapter by streaming ridge regression as in~\Cref{equ:linear_fitting}.
This gives a strong linear baseline with minimal optimization overhead, but its receptive field and representational capacity are limited.

\paragraph{Residual bottleneck adapter.}
The second variant keeps the same linear convolution as a skip path and adds a nonlinear bottleneck branch. The skip branch is
\begin{equation}
    \mathbf{S}
    =
    \operatorname{Conv3D}_{5\times 3\times 3,\;1\times 2\times 2,\;2\times 1\times 1}^{16\rightarrow 1536}
    (\mathbf{Z}) .
\end{equation}
The residual branch first expands the channel dimension with a pointwise projection, applies normalization and GELU, performs local spatiotemporal mixing, and then projects the features back to the DA3 token dimension:
\begin{equation}
    \mathbf{U}
    =
    \operatorname{GELU}
    \left(
    \operatorname{Norm}
    \left(
    \operatorname{Conv3D}_{1\times 1\times 1}^{16\rightarrow d}
    (\mathbf{Z})
    \right)
    \right),
\end{equation}
\begin{equation}
    \mathbf{V}
    =
    \operatorname{GELU}
    \left(
    \operatorname{Norm}
    \left(
    \operatorname{Conv3D}_{5\times 3\times 3,\;1\times 2\times 2,\;2\times 1\times 1}^{d\rightarrow d}
    (\mathbf{U})
    \right)
    \right),
\end{equation}
\begin{equation}
    \mathbf{R}
    =
    \operatorname{Conv3D}_{1\times 1\times 1}^{d\rightarrow 1536}
    (\mathbf{V}) .
\end{equation}
The final prediction is the sum of the linear fit and the nonlinear residual correction:
\begin{equation}
    \hat{\mathbf{Y}}
    =
    \mathcal{A}_{\theta}^{\mathrm{res}}(\mathbf{Z})
    =
    \mathbf{S}+\mathbf{R}.
\end{equation}
This design preserves the stability of the linear adapter while allowing the bottleneck branch to model nonlinear deviations from the ridge-regression solution.

\paragraph{Axial-attention adapter.}
The third variant uses a convolutional stem followed by axial self-attention~\citep{vaswani2017attention}. The stem first lifts the Wan latent into a hidden feature space:
\begin{equation}
    \mathbf{H}_0
    =
    \operatorname{GELU}
    \left(
    \operatorname{Norm}
    \left(
    \operatorname{Conv3D}_{5\times 3\times 3,\;1\times 2\times 2,\;2\times 1\times 1}^{16\rightarrow d}
    (\mathbf{Z})
    \right)
    \right).
\end{equation}
We then apply multi-head self-attention independently along the temporal, height, and width axes. For an axis $a\in\{T,H,W\}$, axial attention is written as
\begin{equation}
    \operatorname{Attn}_{a}(\mathbf{H})
    =
    \operatorname{Reshape}_{a}^{-1}
    \left[
    \operatorname{MHA}
    \left(
    \operatorname{LN}
    \left(
    \operatorname{Reshape}_{a}(\mathbf{H})
    \right)
    \right)
    \right],
\end{equation}
where $\operatorname{Reshape}_{a}$ groups tokens into one-dimensional sequences along axis $a$ while treating the remaining axes as batch dimensions. The hidden state is updated by residual axial attention:
\begin{equation}
    \mathbf{H}_{T}
    =
    \mathbf{H}_0
    +
    \operatorname{Attn}_{T}(\mathbf{H}_0),
\end{equation}
\begin{equation}
    \mathbf{H}_{TH}
    =
    \mathbf{H}_{T}
    +
    \operatorname{Attn}_{H}(\mathbf{H}_{T}),
\end{equation}
\begin{equation}
    \mathbf{H}_{THW}
    =
    \mathbf{H}_{TH}
    +
    \operatorname{Attn}_{W}(\mathbf{H}_{TH}).
\end{equation}
Finally, a pointwise projection maps the attended representation to the DA3 token dimension:
\begin{equation}
    \hat{\mathbf{Y}}
    =
    \mathcal{A}_{\theta}^{\mathrm{axial}}(\mathbf{Z})
    =
    \operatorname{Conv3D}_{1\times 1\times 1}^{d\rightarrow 1536}
    (\mathbf{H}_{THW}) .
\end{equation}
This architecture gives the adapter access to long-range temporal and spatial context while avoiding the quadratic cost of full 3D attention. It is the most expressive variant among the three, but it also has the highest computational and memory cost.

\begin{figure}[!tbp]
\centering
\includegraphics[width=0.78\linewidth]{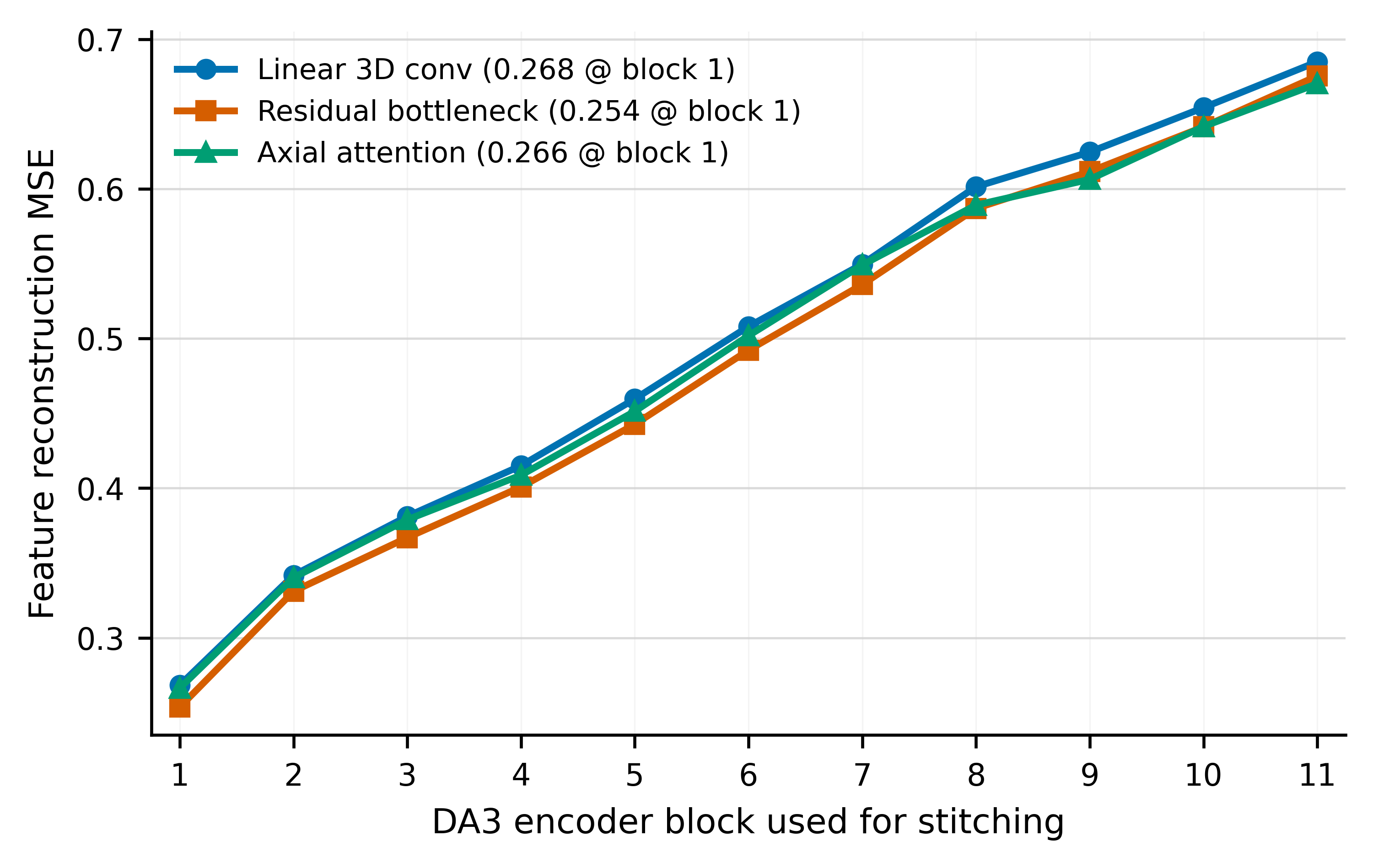}
\caption{Stitching adapter architecture ablation over DA3 encoder blocks 1--11. Lower MSE means that the adapter reconstructs the frozen DA3 feature grid more accurately from the Wan VAE latent. All three variants prefer the earliest candidate block, and the linear 3D convolution remains close to the heavier residual and attention-based adapters across the sweep.}
\label{fig:stitching_adapter_architecture_ablation}
\end{figure}

\Cref{fig:stitching_adapter_architecture_ablation} shows the reconstruction MSE for blocks 1--11. All three curves have the same dominant trend. The earliest DA3 block is easiest to reach from the Wan latent, while deeper blocks become progressively harder to reconstruct. This aligns with the insight from \citet{go2025vist3a} that early 3D foundation model encoder features serve as a foundational geometric interface. This observation directly motivates our architectural design in \Cref{sec:methodology}. By introducing the stitch at an early stage, we capture these representations before the network accumulates highly specialized geometric processing.

The architectural comparison is more important than the absolute block ranking. At the best block, the linear 3D convolution obtains an MSE of 0.268. The residual bottleneck obtains 0.254, and axial attention obtains 0.266. Averaged over blocks 1--11, the MSEs are 0.499, 0.485, and 0.491 for the linear convolution, residual bottleneck, and axial attention variants. The nonlinear residual branch gives the lowest standalone fitting error, but the gain is small relative to the added normalization, activation, and extra convolutional layers. Axial attention is even closer to the linear convolution despite its substantially higher memory and compute cost.

These results support the simpler adapter used in the main system. The stitching layer is not intended to replace the remaining DA3 decoder. Its job is to place the video latent on a compatible DA3 token manifold and provide a stable initialization for the stitched decoder finetuning stage. A linear 3D convolution already captures most of this alignment. More expressive residual or attention-based adapters provide only modest reconstruction gains, so they do not justify their additional cost for this interface.

\subsection{Ablation for Stitched Decoder Finetuning Strategy}

The stitched decoder is trained for a specific role. It must accept Wan VAE latents as input while retaining the geometric behavior of the original DA3 model. This creates a tension between adaptation and preservation. Teacher-student distillation keeps the student close to the pretrained DA3 function, while ground-truth geometric supervision anchors the predictions to metric depth, camera, point-cloud, and motion annotations when such labels are available. We ablate this finetuning strategy to identify which ingredients are needed for a stable latent-to-geometry decoder.

We follow the same benchmarks and metrics as the main stitched-model evaluation in Table~\ref{tab:stitched-da3-alignment}. The variants isolate the main design choices in the finetuning stage. \textbf{Distillation only} trains the student only from the frozen DA3 teacher outputs. The teacher receives the ground-truth RGB clip, and the student receives the Wan latent of the same clip. \textbf{Distillation with feature loss} introduces an auxiliary feature-matching term to regularize the stitched DA3 representation. Specifically, it applies an $L_1$ penalty between the output features of the stitching layer and the corresponding block features from the teacher model. This constraint forces the projected latent tokens to remain aligned with the teacher's feature manifold prior to dense decoding. \textbf{GT supervision only} removes the teacher and trains only from available dataset annotations. \textbf{No scene-flow head} keeps the depth, ray, camera, and reconstruction objectives but removes the motion branch during stitched-decoder finetuning. \textbf{LoRA adaptation} replaces the full update of the stitched decoder with a parameter-efficient alternative~\citep{hu2022lora} on the adapted DA3 computation path. Specifically, we apply a low-rank update with both the rank and scaling factor $\alpha$ set to 32 and maintain a learning rate of $1 \times 10^{-4}$.

\begin{table}[t]
\centering
\small
\setlength{\tabcolsep}{6pt}
\renewcommand{\arraystretch}{1.15}
\caption{Ablation of stitched-decoder finetuning strategies under the same evaluation protocol as Table~\ref{tab:stitched-da3-alignment}. The table reports representative reconstruction and camera metrics that diagnose metric 3D preservation and dynamic-scene camera stability. In each metric column, the best value is highlighted in red and bold, the second-best value in orange, and the third-best value in yellow.}
\label{tab:model-stitching-variant-ablation}
\begin{tabular}{llcc}
\toprule
Category & Variant & ETH3D Acc. $\downarrow$ & Dynamic-RE10K AUC$_{30}$ $\uparrow$ \\
\midrule
Supervision & Distillation only & 1.4276 & \trd{0.5259} \\
Supervision & Distillation with feature loss & \snd{0.5209} & \snd{0.5330} \\
Supervision & GT supervision only & 1.7899 & 0.3141 \\
Architecture & No scene-flow head & \trd{0.8459} & 0.5139 \\
Adaptation & LoRA adaptation & 2.6344 & 0.5144 \\
Full model & Ours (full) & \fst{0.2898} & \fst{0.5350} \\
\bottomrule
\end{tabular}
\end{table}

The results show that output distillation alone is not enough. It preserves coarse camera behavior relatively well, but its reconstruction quality degrades noticeably. This suggests that matching final teacher predictions can transfer some global geometric structure, yet it does not fully constrain the accurate geometry after the RGB encoder has been replaced by a latent input pathway. Compared to our full model trained with mixed supervision, combining output-space distillation and ground-truth data, incorporating an additional feature loss actually degrades performance. This phenomenon aligns with the insights of \citet{huang2026self}. It suggests that rigid representation alignment cannot independently recover all the geometric behaviors required by the stitched decoder. Furthermore, enforcing such strict alignment may impair the flexibility of the learned representations since the underlying model architecture has changed.

The GT-only variant fails for a distinct reason. While direct supervision provides explicit targets, it eliminates the pretrained DA3 teacher signal that contains broad reconstruction priors. Furthermore, some low quality ground-truth data can be sparse or inaccurate locally, hindering the transfer learning capabilities of the model. This observation aligns with the insights of \citet{lin2025depth}. Consequently, the network becomes highly sensitive to dataset coverage, annotation quality, and task imbalances. This vulnerability is particularly detrimental to camera estimation, where the teacher provides a strong geometric prior that is exceedingly difficult to learn from supervised labels alone. These results clarify the proper role of ground-truth supervision within our system. It serves as a valuable corrective signal but must complement distillation rather than replace it.

The architectural and adaptation ablations further clarify what the finetuning stage must optimize. Removing the scene-flow head weakens both reported axes, even though the compact table does not directly score scene flow. This indicates that motion supervision contributes useful temporal pressure to the shared decoder representation and benefits the geometry heads beyond the motion output itself. LoRA adaptation is also insufficient for this transfer. It keeps some camera behavior, but reconstruction degrades severely, which suggests that the distribution shift from RGB tokens to Wan latents is too large for a narrow low-rank update.

The full model is strongest on both reported axes. Its advantage is most visible in reconstruction, where combining distillation and ground-truth supervision best preserves 3D structure under the latent input path. The camera gains are more modest because several variants already retain coarse camera cues, but the full objective still gives the most stable result. This ablation supports the hybrid finetuning strategy used in \shortname. Distillation preserves the broad DA3 prior, ground-truth geometry corrects metric errors left by teacher-only learning, and scene-flow supervision helps maintain temporal structure in the shared decoder features.

\section{Limitations}
\label{sec:appendix_limitations}

Our method serves as a geometry-aware intermediate world model for predicting future visual and 4D geometric states, rather than an end-to-end embodied agent. Because it lacks closed-loop physical deployment and relies purely on offline benchmarks, the impact of prediction error accumulation on downstream decision-making remains untested. Future work must therefore integrate this predictive representation with explicit action policies to evaluate task success under real-world execution constraints.

\section{Broader Impact}
\label{sec:appendix_broader_impact}

While our geometry-aware world model advances embodied planning and simulation through structured 4D future predictions, it introduces potential risks, including misuse in surveillance and overreliance in safety-critical applications susceptible to prediction errors and data biases. To ensure responsible deployment, these predictive outputs must be treated as probabilistic forecasts requiring uncertainty-aware downstream validation, augmented with strict task-specific safeguards prior to physical execution, and rigorously evaluated against distribution shifts and embodied failure modes. These measures support more responsible use of geometry-aware world models.

\section{Licenses and Terms of Use}
\label{sec:appendix_licenses}
All external code, models, and datasets utilized in our experiments are properly cited and strictly governed by their respective licenses and usage restrictions.

\section{LLM Usage Statement}
\label{sec:appendix_llm_usage}

We used a general-purpose large language model as a writing assistant for limited language polishing during manuscript preparation, including improvements to phrasing, readability, and English fluency. We also used Qwen-VL-30B-Thinking to generate per-clip action captions for the Dynamic Replica and PointOdyssey evaluation splits. These captions were used as the conditioning text for all compared methods. The captioning model was used for evaluation annotation preparation and was not a component of the proposed \shortname{} model.

\section{Experimental Compute Resources}
Sec. 4 and Appendix~\ref{sec:app_train_details} report GPU types 
(H100 80GB for Stage 1, H20 96GB for Stages 2 and 3), GPU counts 
(16 for Stage 1, 64 for Stages 2 and 3), per-GPU batch size (1), 
and training epochs for all three stages. Stage 1 takes 
approximately 60 hours on 16 H100 GPUs ($\sim$960 H100 GPU 
hours); Stage 2 takes approximately 90 hours on 64 H20 GPUs 
($\sim$5{,}760 H20 GPU hours); Stage 3 takes approximately 160 
hours on 64 H20 GPUs ($\sim$10{,}240 H20 GPU hours), totaling 
approximately 16{,}960 GPU hours across all three stages. 
Inference runs on 8 H20 GPUs and takes approximately 15 minutes 
per clip. The full research project required additional compute 
beyond the reported training runs. Ablation studies on stitching 
adapter architectures and stitched-decoder finetuning strategies 
follow the same stage-wise GPU configuration and duration as the 
corresponding main experiments, contributing an estimated 
$\sim$34{,}000 additional GPU hours in total across all ablation 
variants.

\section{More Experimental Results}

\input{tables/4d_generation}

\paragraph{Full quantitative comparison.}
\Cref{tab:main_results} reports the complete joint 4D generation evaluation across HoiGen, Epic-Kitchens-100, Dynamic Replica, PointOdyssey, and RLBench. Epic-Kitchens-100 extends this evaluation to everyday kitchen interactions under the same protocol, where \shortname achieves the best values on 10 of the 13 metrics. We group the metrics into video quality, depth and reconstruction, and camera estimation. The discussion below focuses only on metrics that are reported for the corresponding methods.

Across video-quality metrics, \shortname achieves the best subject consistency on all five benchmarks and the best motion smoothness on HoiGen, Dynamic Replica, and RLBench. It leads in caption similarity on Epic-Kitchens-100 and PointOdyssey, while VIST3A leads on HoiGen, Dynamic Replica, and RLBench. On Epic-Kitchens-100, its SC and BC are 0.9506 and 0.9555, compared with 0.8923 and 0.9223 for TesserAct, and its Cap.Sim is 0.4010, compared with 0.3710 for the next-best method, VIST3A. TesserAct remains competitive on individual appearance metrics, such as image quality on HoiGen and RLBench, aesthetic quality on Dynamic Replica, and motion smoothness on Epic-Kitchens-100. VIST3A also leads in aesthetic quality on Epic-Kitchens-100. These results indicate strong visual consistency across benchmarks, while individual appearance and caption-alignment metrics favor different methods.

The depth and reconstruction columns show the value of reading geometry from the same latent future. \shortname obtains the best depth metrics on all five datasets, with substantially lower AbsRel and higher threshold accuracy than TesserAct and Gen3R. On Epic-Kitchens-100, it obtains an AbsRel of 0.0766 with $\delta_1=0.9265$ and $\delta_2=0.9846$. It also gives the best Chamfer-L1 reconstruction on Epic-Kitchens-100, Dynamic Replica, PointOdyssey, and RLBench among methods with reported reconstruction outputs. On Epic-Kitchens-100, its Chamfer-L1 of 0.2321 is lower than Gen3R's 0.3390, TesserAct's 1.3321, and VIST3A's 6.4722. Gen3R is strongest on HoiGen Chamfer-L1, but its depth and camera metrics are much worse on the same benchmark. This indicates that a lower point-cloud distance in one setting does not necessarily imply a temporally coherent RGB-D world model. By contrast, \shortname keeps depth and reconstruction strong across scene types, from hand-object interaction to dynamic stereo and robot manipulation.

Camera estimation further separates the methods that preserve global 4D structure from those that only recover local geometry. \shortname achieves the best reported RPE-R on HoiGen, Dynamic Replica, and PointOdyssey, and matches or approaches perfect relative-rotation accuracy on Dynamic Replica and PointOdyssey. On Epic-Kitchens-100, it achieves RRA@10 and RRA@5 of 1.0000, although TesserAct has a lower RPE-R of 0.0946 compared with our 0.1568. TesserAct also leads on the RLBench camera metrics. Gen3R consistently has much larger rotation errors, which reflects the difficulty of maintaining stable global camera motion when geometry is generated without the same latent-level coupling to the video model. The Epic-Kitchens-100 results provide quantitative evidence of visual, geometric, and camera consistency on in-the-wild interactions and, together with HoiGen, support the generality of \shortname across two real-world human-object interaction benchmarks. Overall, the full table supports the same conclusion as the main paper: \shortname balances action-conditioned video quality, dense geometry, and camera stability in one forward generation process. These quantitative scores complement the qualitative snapshots in the paper, while full qualitative videos remain useful for inspecting the temporal behavior of generated 4D scenes.

\subsection{Video-Only World Models with Post Hoc Geometry Reconstruction}
\label{subsec:app_video_only_baselines}

This comparison tests whether a strong video-only world model followed by an independent geometry reconstructor can match joint video and 4D generation. We evaluate two additional baselines on HoiGen, including a direct combination of the two pretrained backbones used by \shortname{}.

\paragraph{Evaluation setting.}
We use the same 200 HoiGen test examples and action-conditioned generation protocol as the main comparison in \Cref{sec:exp_settings,tab:main_results}. Each method receives the initial RGB frame and an action description. For geometric evaluation, we feed the ground-truth video into DA3 and use its reconstructed depth and camera poses as the common reference. The geometric scores therefore measure agreement with DA3-derived pseudo ground truth. We report subject consistency and background consistency for video quality, depth AbsRel and $\delta_1$ for depth accuracy, and camera relative rotation error (RPE-R). Depth uses the same per-clip scale-shift alignment as the main evaluation, and camera evaluation follows the same pose-alignment protocol described in \Cref{sec:app_train_details}. Higher consistency scores and $\delta_1$ are better, while lower AbsRel and RPE-R are better.

\paragraph{Baseline construction.}
\textbf{Cosmos-3-Nano + VGGT-Omega} uses Cosmos-3-Nano~\citep{agarwal2026cosmos} to generate RGB video and applies VGGT-Omega~\citep{wang2026vggt} to reconstruct geometry from the generated frames. This baseline tests post hoc geometry recovery from a strong video-only world model. \textbf{LongCat-Video + DA3} first generates an RGB video with LongCat-Video and then applies DA3 directly to the generated frames. It combines our two pretrained backbones without joint training and tests whether their independent use is sufficient for consistent 4D generation. In both baselines, geometric reconstruction takes place after RGB video generation.

\begin{table}[t]
\centering
\small
\setlength{\tabcolsep}{5pt}
\renewcommand{\arraystretch}{1.15}
\caption{Comparison with video-only generation followed by post hoc geometry reconstruction on HoiGen under the main action-conditioned evaluation protocol. Geometric references are reconstructed by applying DA3 to the ground-truth videos. Arrows indicate the preferred direction, and the best value in each column is in \textbf{bold}.}
\label{tab:app_video_only_baselines}
\resizebox{\linewidth}{!}{%
\begin{tabular}{lrrrrr}
\toprule
Method & Subj.\ Cons.\ $\uparrow$ & BG Cons.\ $\uparrow$ & Depth AbsRel $\downarrow$ & Depth $\delta_1$ $\uparrow$ & Camera RPE-R $\downarrow$ \\
\midrule
Cosmos-3-Nano + VGGT-Omega & 0.8760 & 0.9213 & 0.4547 & 0.5850 & 0.2933 \\
LongCat-Video + DA3 & 0.8864 & 0.9235 & 0.2815 & 0.6875 & 0.3450 \\
\shortname{} (Ours) & \textbf{0.9035} & \textbf{0.9334} & \textbf{0.1125} & \textbf{0.8540} & \textbf{0.1240} \\
\bottomrule
\end{tabular}%
}
\end{table}

\paragraph{Results and analysis.}
\Cref{tab:app_video_only_baselines} shows that \shortname{} achieves the best result on all five reported metrics. Compared with LongCat-Video + DA3, subject and background consistency increase from 0.8864 and 0.9235 to 0.9035 and 0.9334. Depth AbsRel decreases from 0.2815 to 0.1125, depth $\delta_1$ increases from 0.6875 to 0.8540, and camera RPE-R decreases from 0.3450 to 0.1240. The comparison with Cosmos-3-Nano + VGGT-Omega shows the same advantage across video, depth, and camera metrics.

These results illustrate a limitation of post hoc geometry recovery. A video generator trained only with appearance objectives receives no explicit supervision on the geometry, camera motion, or scene dynamics of its generated future. A downstream reconstructor must infer these quantities from the generated images, so visual inconsistencies can propagate into the recovered geometry. In \shortname{}, appearance, depth, camera geometry, and scene flow supervise a shared latent representation during joint training. The resulting latent future must support both RGB generation and geometric readout, encouraging visual and 4D consistency within the same rollout. Our stitching framework makes this supervision practical by connecting large-scale video data to a pretrained geometry model, while paired dynamic 3D datasets provide additional supervision where available.

\subsection{More Qualitative Results}

We provide additional qualitative examples to further evaluate the behavior of \shortname across diverse action scenarios. Each example is conditioned on a single initial frame and a natural-language description, from which the model predicts an evolving 4D future. We visualize the generated RGB frames, depth maps, and global point clouds from the same prediction, allowing appearance, geometry, and temporal consistency to be inspected jointly. The examples cover object manipulation, human motion, physical interaction, and multi-entity indoor dynamics, showing that \shortname{} can produce action-aligned visual changes while preserving coherent 3D structure over time.

\begin{figure}[!tbp]
  \centering
  \includegraphics[width=\linewidth]{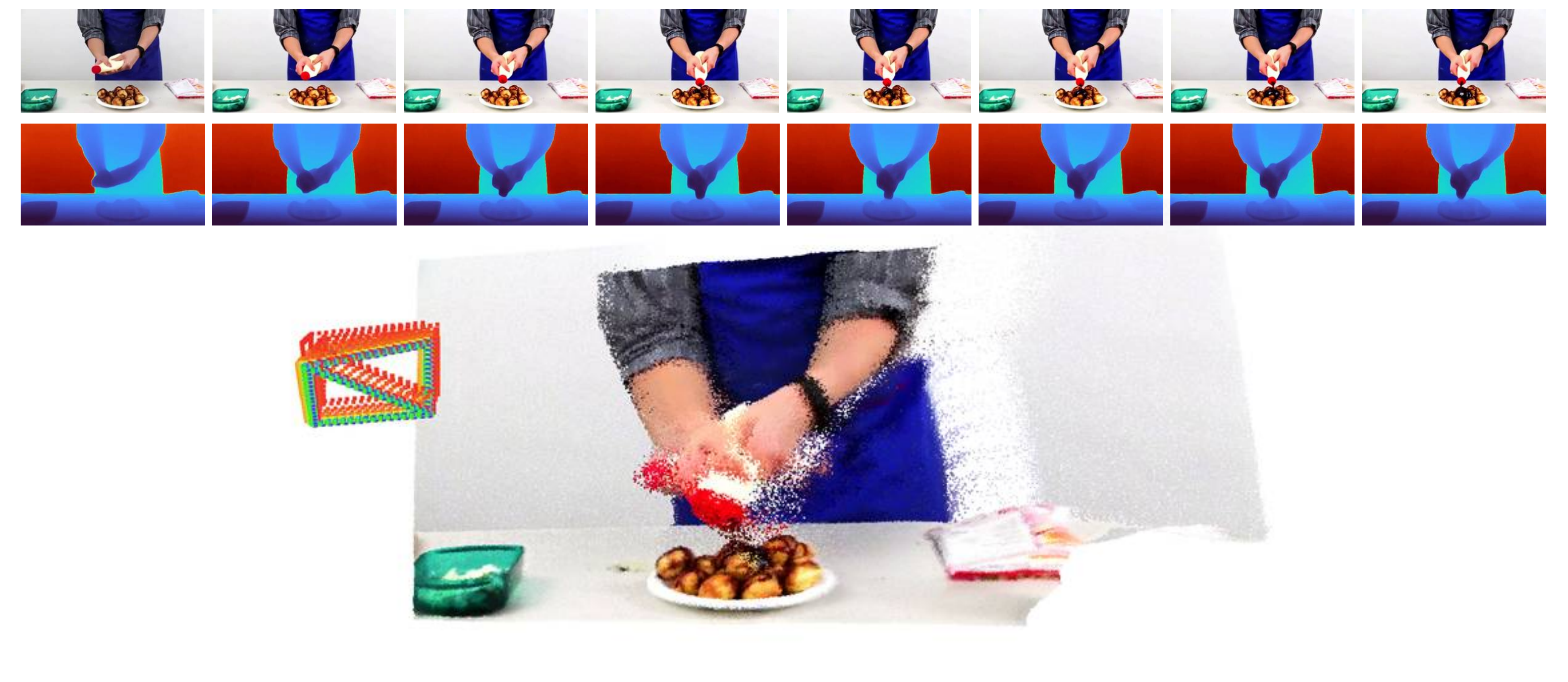}
  \caption{\textbf{Qualitative results of \shortname{}.} We provide the model with an initial frame and a descriptive text prompt: \textit{``Removes the red cap from the squeeze bottle. Holds the bottle with both hands. Squeezes the bottle. Drizzles the sauce over the takoyaki in a controlled pattern. Finishes adding the sauce.''} Based on these inputs, \shortname{} generates a dense 4D prediction. The figure visualizes the synthesized RGB frames (top row), depth maps (middle row), and 4D global point clouds (bottom row).}
  \label{fig:qualitative_results}
\end{figure}

\begin{figure}[!tbp]
  \centering
  \includegraphics[width=\linewidth]{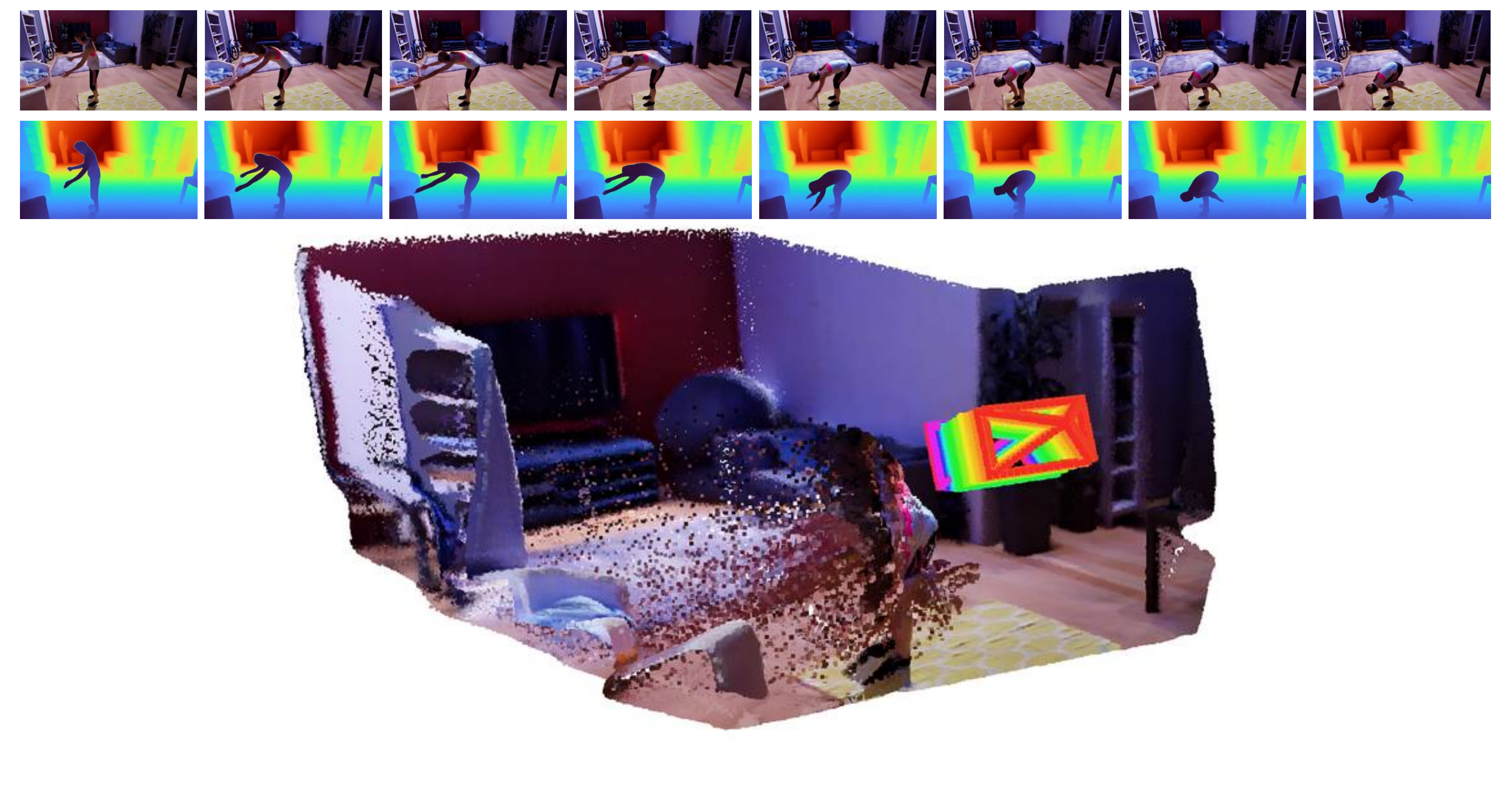}
  \caption{\textbf{Qualitative results of \shortname{}.} We provide the model with an initial frame and a descriptive text prompt: \textit{``Person bends forward, arms extended, performing a slow motion movement on the yellow rug.''} Based on these inputs, \shortname{} generates a dense 4D prediction. The figure visualizes the synthesized RGB frames (top row), depth maps (middle row), and 4D global point clouds (bottom row).}
  \label{fig:qualitative_results2}
\end{figure}

\begin{figure}[!tbp]
  \centering
  \includegraphics[width=\linewidth]{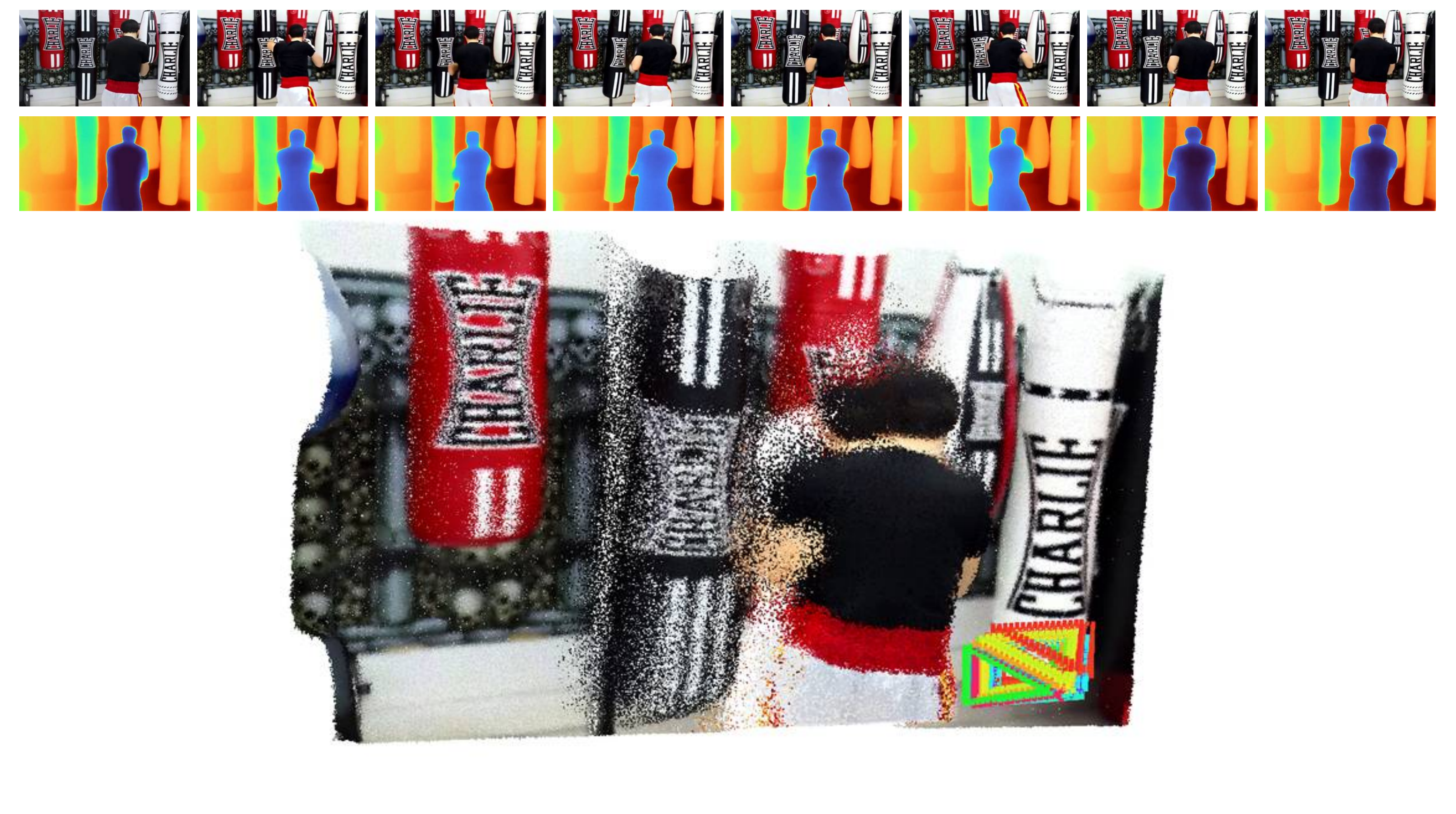}
  \caption{\textbf{Qualitative results of \shortname{}.} We provide the model with an initial frame and a descriptive text prompt: \textit{``Delivers a series of punches to the heavy bag. The bag swings backward / recoils away upon impact.''} Based on these inputs, \shortname{} generates a dense 4D prediction. The figure visualizes the synthesized RGB frames (top row), depth maps (middle row), and 4D global point clouds (bottom row).}
  \label{fig:qualitative_results3}
\end{figure}

\begin{figure}[!tbp]
  \centering
  \includegraphics[width=\linewidth]{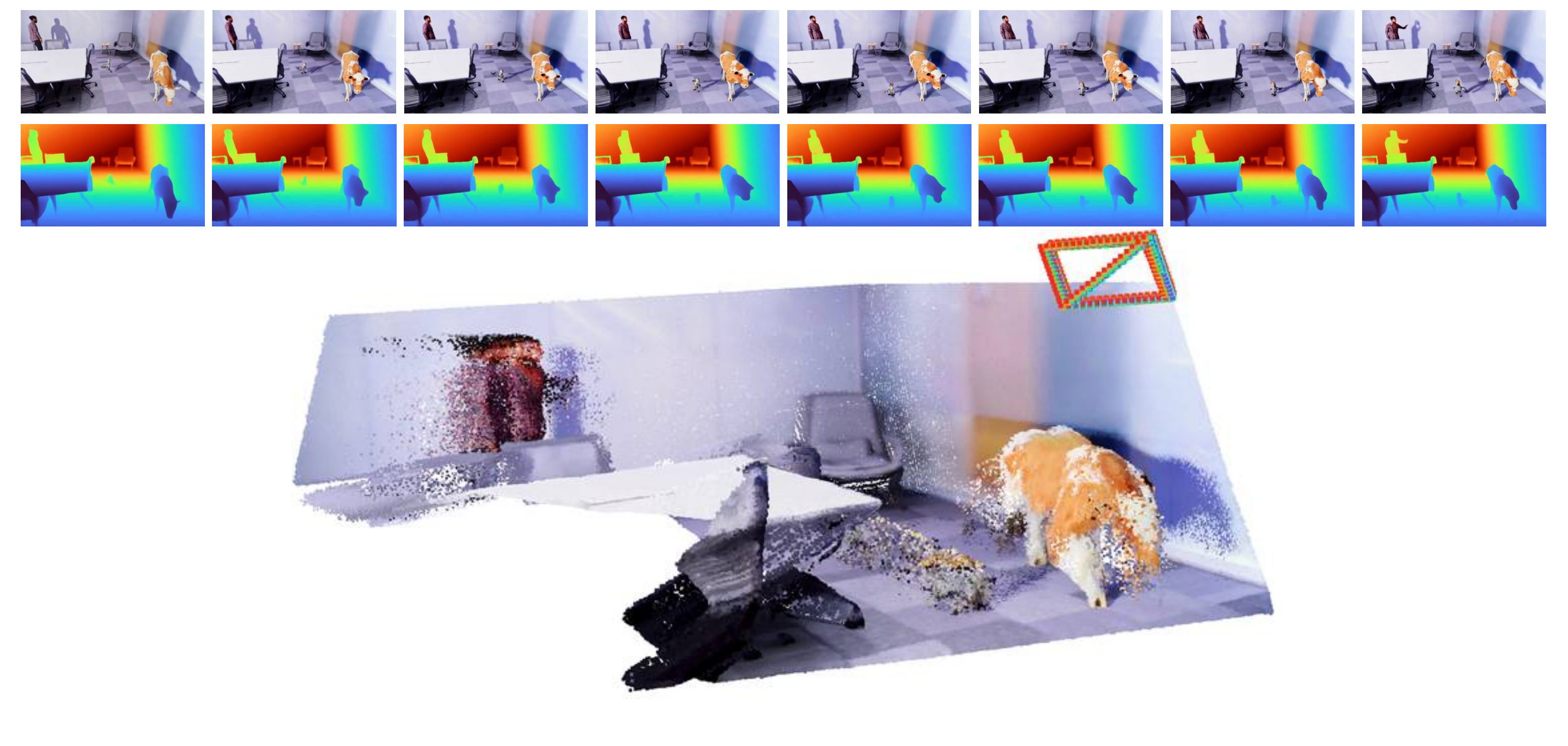}
  \caption{\textbf{Qualitative results of \shortname{}.} We provide the model with an initial frame and a descriptive text prompt: \textit{``Man lifts leg, monkey stands and walks, cow shifts position in room.''} Based on these inputs, \shortname{} generates a dense 4D prediction. The figure visualizes the synthesized RGB frames (top row), depth maps (middle row), and 4D global point clouds (bottom row).}
  \label{fig:qualitative_results4}
\end{figure}

\subsection{Diagnostic Study on Image--Text Misalignment}
\label{subsec:app_image_text_misalignment}

Action-conditioned 4D generation assumes that the input frame and the text prompt describe a compatible scene. In practice, the two conditions may conflict, and it is important to understand which condition \shortname{} follows in that case. No established benchmark currently evaluates image--text conflicts for action-conditioned 4D generation, so we conduct a targeted diagnostic study.

\paragraph{Setting.}
We select 34 held-out image--text pairs and keep each original input frame fixed. For each frame, we construct two mismatched prompts, which yields 68 rollouts in total. The two prompt types probe different degrees of conflict.
\begin{itemize}
    \item \textbf{Type A, semantically disjoint.} The prompt describes an event that cannot occur in the observed scene. For example, an egocentric kitchen frame showing a hand on dough is paired with \textit{``A group of penguins dives from an iceberg into the cold ocean.''}
    \item \textbf{Type B, scene editing.} The prompt requests an implausible transformation of the observed scene, such as \textit{``Transform the dough into a pile of transparent glass marbles and make the dark counter become rippling beach sand.''}
\end{itemize}

\paragraph{Categorization protocol.}
We manually assign each rollout to one of three outcomes. \textbf{C (cut)} denotes a rollout that satisfies the prompt through a scene cut to the described content. \textbf{M (morph)} denotes a rollout that satisfies the prompt through an instantaneous transformation of objects or surfaces in the observed scene. \textbf{P (plausible)} denotes a rollout that preserves the original scene and continues a plausible action instead of realizing the conflicting prompt.

\begin{table}[t]
\centering
\small
\setlength{\tabcolsep}{6pt}
\renewcommand{\arraystretch}{1.15}
\caption{Diagnostic study on image--text misalignment. Each of the 34 held-out input frames is paired with one semantically disjoint prompt (Type A) and one scene-editing prompt (Type B), giving 68 rollouts. Each entry counts the rollouts manually assigned to an outcome. C denotes satisfying the prompt through a scene cut, M denotes satisfying it through an instantaneous morph, and P denotes preserving the original scene and continuing a plausible action. Each row sums to $n=34$.}
\label{tab:app_image_text_misalignment}
\begin{tabular}{lccc}
\toprule
Mismatch type & C: cut & M: morph & P: plausible \\
\midrule
Type A: semantically disjoint ($n=34$) & 34 & 0 & 0 \\
Type B: scene editing ($n=34$) & 0 & 30 & 4 \\
\bottomrule
\end{tabular}
\end{table}

\paragraph{Results.}
\Cref{tab:app_image_text_misalignment} summarizes the outcomes. For Type A, all 34 rollouts begin by preserving the input scene and then transition to the scene described by the prompt, in a manner reminiscent of a cinematic montage cut. For Type B, 30 of the 34 rollouts introduce the requested objects through abrupt transformations, while the remaining 4 preserve the original scene and continue a plausible action.

\begin{figure}[!tbp]
    \centering
    \includegraphics[width=\linewidth]{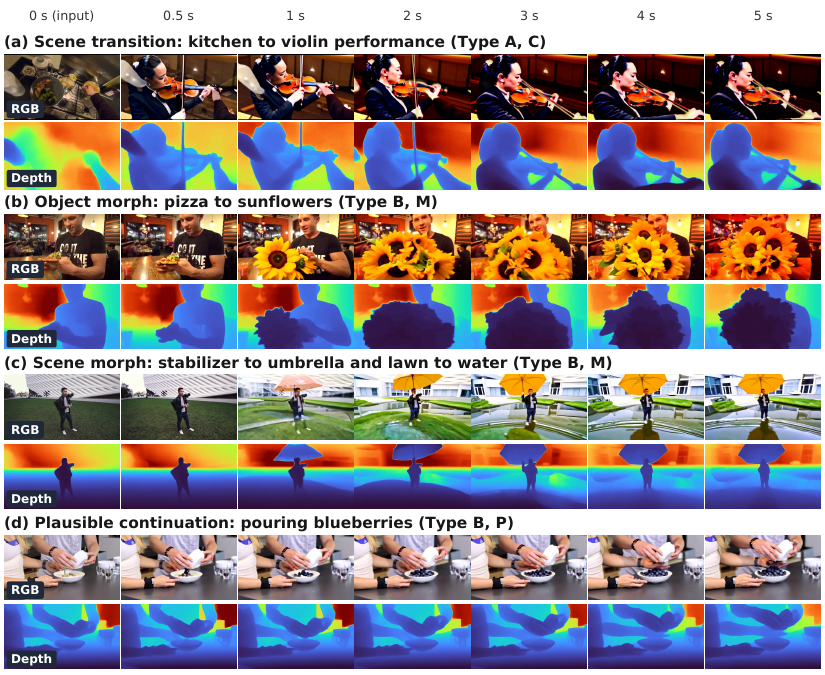}
    \caption{\textbf{Qualitative outcomes under image--text misalignment.} Each case shows RGB above the corresponding predicted depth at seven shared time points. The first RGB column is the conditioning image. Case (a) illustrates a scene transition (C) under Type A. Cases (b) and (c) illustrate object and scene transformations (M) under Type B. Case (d) preserves the original scene and continues a plausible action (P) under Type B. The complete input prompts are reported in the text.}
    \label{fig:app_image_text_misalignment_qualitative}
\end{figure}

\paragraph{Qualitative examples and input prompts.}
\Cref{fig:app_image_text_misalignment_qualitative} illustrates the three outcomes with four selected rollouts. We show the initial image and sample the generated RGB and depth at $0.5$, $1$, $2$, $3$, $4$, and $5$ seconds. The input text prompts are:
\begin{itemize}
    \item \textbf{(a) Scene transition.} \textit{``A violinist performs alone under a spotlight in an empty concert hall.''}
    \item \textbf{(b) Object morph.} \textit{``Keep the same man, pose, and restaurant background, but transform the pizza in his hands into a large bouquet of yellow sunflowers and make his black shirt turn bright red.''}
    \item \textbf{(c) Scene morph.} \textit{``Keep the same man, pose, lawn, and modern building, but transform the camera stabilizer into an open yellow umbrella and turn the grass beneath him into shallow reflective water.''}
    \item \textbf{(d) Plausible continuation.} \textit{``Keep the same people, bowl, and tabletop, but make every blueberry disappear and fill the white bowl with shiny gold coins as the container is tilted above it.''}
\end{itemize}

In (a), the egocentric kitchen view gives way to a close-up of a violinist. In (b), a sunflower bouquet replaces the pizza while the person and restaurant remain recognizable. In (c), an umbrella appears in place of the stabilizer and the lawn develops a reflective water surface. In (d), the model continues pouring blueberries into the bowl without introducing the requested gold coins. The paired depth maps follow the main foreground boundaries in the displayed frames, including the violinist, expanding bouquet, umbrella, and hands around the bowl. This correspondence can coexist with an implausible scene change or incomplete prompt fulfillment, as also illustrated by the unchanged black shirt in (b). The snapshots provide qualitative evidence of agreement between RGB and predicted depth, while physical continuity and complete adherence to the conflicting prompt remain separate properties.

\paragraph{Analysis.}
The diagnostic results in \Cref{tab:app_image_text_misalignment,fig:app_image_text_misalignment_qualitative} show that \shortname{} generally prioritizes the text condition when the image and text conflict. When the prompt cannot be naturally realized in the observed scene, the model resolves the mismatch through a scene transition or an instantaneous transformation. This behavior mainly originates from the text-conditioned video prior. The stitched geometry decoder still produces synchronized depth, camera rays, and scene flow for the generated trajectory, although physical continuity may be broken at the transition. Because \shortname{} explicitly predicts geometry such as scene flow and camera rays, these abrupt transitions may also be detectable as geometric discontinuities. Such discontinuities could provide a lightweight infeasibility signal for downstream planning.

%% file: tables/4d_generation.tex
\begin{table*}[t]
\centering
\caption{Quantitative comparison across five benchmarks. Epic-Kitchens-100 uses 100 five-second clips of in-the-wild kitchen interactions evaluated under the same protocol as the other benchmarks. Arrows indicate the preferred direction. Best results for each dataset and metric are in \textbf{bold}.}
\label{tab:main_results}
\resizebox{\textwidth}{!}{%
\begin{tabular}{ll|ccccc c|ccc c|ccc}
\toprule
& & \multicolumn{6}{c|}{\textbf{Video Quality}} & \multicolumn{4}{c|}{\textbf{Depth \& Reconstruction}} & \multicolumn{3}{c}{\textbf{Camera}} \\
\cmidrule(lr){3-8} \cmidrule(lr){9-12} \cmidrule(lr){13-15}
Dataset & Model
& SC $\uparrow$ & BC $\uparrow$ & AQ $\uparrow$ & IQ $\uparrow$ & MS $\uparrow$ & Cap.Sim $\uparrow$
& AbsRel $\downarrow$ & $\delta_1$ $\uparrow$ & $\delta_2$ $\uparrow$ & Chamfer-L1 $\downarrow$
& RPE-R $\downarrow$ & RRA@10 $\uparrow$ & RRA@5 $\uparrow$ \\
\midrule
\multirow{4}{*}{HoiGen}
& \textbf{Ours}   & \textbf{.9035} & \textbf{.9334} & \textbf{.4802} & .6102          & \textbf{.9832} & .6854 & \textbf{.1125} & \textbf{.854}  & \textbf{.9596} & .3325          & \textbf{.1240} & \textbf{.9903} & \textbf{.9809} \\
& TesserAct       & .8661          & .9187          & .4480          & \textbf{.6164} & .9802          & .6401          & 1.850          & .1839          & .3402          & .3534          & .1429          & .9725          & .9256          \\
& Gen3R           & .7765          & .8774          & .4391          & .4826          & .9532          & .6803          & .2881          & .5685          & .7789          & \textbf{.2286} & 5.060          & .4891          & .2689          \\
& VIST3A         & .6721          & .8346          & .4730          & .5600          & .9383          & \textbf{.7851}           & .6301             & .2570             & .3761             & .7633             & 13.09             & .1298             & .0185             \\
\midrule
\multirow{4}{*}{\shortstack[l]{Epic-Kitchens-\\100}}
& \textbf{Ours}   & \textbf{.9506} & \textbf{.9555} & .4207          & \textbf{.6435} & .9867          & \textbf{.4010} & \textbf{.0766} & \textbf{.9265} & \textbf{.9846} & \textbf{.2321} & .1568          & \textbf{1.0000} & \textbf{1.0000} \\
& TesserAct       & .8923          & .9223          & .4061          & .6072          & \textbf{.9874} & .3343          & .4201          & .7519          & .9207          & 1.3321         & \textbf{.0946} & .9983           & .9736           \\
& Gen3R           & .7813          & .8691          & .3618          & .4706          & .9793          & .3305          & .3641          & .4580          & .6980          & .3390          & 2.3992         & .6104           & .3510           \\
& VIST3A         & .5499          & .8055          & \textbf{.4556} & .4994          & .9768          & .3710          & .1326          & .8204          & .9708          & 6.4722         & 34.7905        & .0275           & .0184           \\
\midrule
\multirow{4}{*}{\shortstack[l]{Dynamic\\Replica}}
& \textbf{Ours}   & \textbf{.9356} & \textbf{.9496} & .5078          & \textbf{.6735} & \textbf{.9927} & .5639 & \textbf{.0696} & \textbf{.9306} & \textbf{.9829} & \textbf{.2579} & \textbf{.1550} & \textbf{1.000} & \textbf{1.000} \\
& TesserAct       & .8490          & .9242          & \textbf{.5115} & .6646          & .9864          & .4419          & 1.289          & .3523          & .5674          & .4067             & .7609             & .9501             & .6772             \\
& Gen3R           & .6833          & .8512          & .4053          & .4906          & .9700          & .4620          & .2018          & .6240          & .9013          & 1.034          & 22.47          & .2998          & .1393          \\
& VIST3A         & .6013          & .8163          & .4867          & .5467          & .9604          & \textbf{.5744 }            & .4084             & .4502             & .5430             & 1.474             & 16.31             & .3098             & .1230             \\
\midrule
\multirow{4}{*}{\shortstack[l]{Point\\Odyssey}}
& \textbf{Ours}   & \textbf{.9233} & \textbf{.9423} & \textbf{.5355} & \textbf{.5941} & .9856          & \textbf{.5579} & \textbf{.0822} & \textbf{.9215} & \textbf{.9720} & \textbf{.3013} & \textbf{.1985} & \textbf{.9965} & \textbf{.9864} \\
& TesserAct       & .9019          & .9391          & .4920          & .5351          & \textbf{.9932} & .5094          & .6347          & .2953          & .5390          & .3904             & .8804          & .8062          & .6446          \\
& Gen3R           & .7486          & .8725          & .4389          & .4164          & .9708          & .4930          & .4227          & .3875          & .6961          & .6703          & 9.409          & .4066          & .2188          \\
& VIST3A         & .6270          & .8173          & .4983          & .4373          & .9777          & .5465             & .4371             & .4054             & .5670             & 1.347             & 6.953             & .4098             & .2962             \\
\midrule
\multirow{4}{*}{RLBench}
& \textbf{Ours}   & \textbf{.8459} & .9017          & \textbf{.4619} & .4767          & \textbf{.9897} & .5583 & \textbf{.0822} & \textbf{.9215} & \textbf{.9721} & \textbf{.3012} & .1985          & .9963          & .9864          \\
& TesserAct       & .8424          & \textbf{.9235} & .4312          & \textbf{.5932} & .9842          & .1791          & 5.481          & .2003          & .3950          & .3904             & \textbf{.0566} & \textbf{1.000} & \textbf{.9999} \\
& Gen3R           & .6434          & .8442          & .3732          & .4086          & .9751          & .1769          & .2468          & .6220          & .8306          & 1.663          & 11.87          & .3712          & .1840          \\
& VIST3A         & .5608          & .8052          & .4375          & .4913          & .9754          & \textbf{.6071 }            & .3441             & .4380             & .5001             & .8709             & 10.701             & .2101             & .1023             \\
\bottomrule
\end{tabular}%
}
\vspace{-2mm}
\end{table*}

%% file: checklist.tex
\clearpage

\section*{NeurIPS Paper Checklist}

\begin{enumerate}

\item {\bf Claims}
    \item[] Question: Do the main claims made in the abstract and introduction accurately reflect the paper's contributions and scope?
    \item[] Answer: \answerYes{} 
    \item[] Justification: The Abstract and Sec.~\ref{sec:intro} state the main contributions and scope clearly, and these claims are supported by the experimental evidence summarized in Sec.~\ref{sec:experiments}.

\item {\bf Limitations}
    \item[] Question: Does the paper discuss the limitations of the work performed by the authors?
    \item[] Answer: \answerYes{} 
    \item[] Justification: The manuscript now discusses the main limitations explicitly in Sec.~\ref{sec:appendix_limitations}, including that the current model does not directly output executable actions and has not yet been validated as a full closed-loop embodied system.

\item {\bf Theory assumptions and proofs}
    \item[] Question: For each theoretical result, does the paper provide the full set of assumptions and a complete (and correct) proof?
    \item[] Answer: \answerNA{} 
    \item[] Justification: The paper does not present theorem-style theoretical results or formal proofs. Its mathematical content defines the model, losses, and evaluation setup in Sec.~\ref{sec:methodology} and the Appendix.

    \item {\bf Experimental result reproducibility}
    \item[] Question: Does the paper fully disclose all the information needed to reproduce the main experimental results of the paper to the extent that it affects the main claims and/or conclusions of the paper (regardless of whether the code and data are provided or not)?
    \item[] Answer: \answerYes{} 
    \item[] Justification: Sec.~\ref{sec:methodology}, Sec.~\ref{sec:experiments}, and Sec.~\ref{sec:app_train_details} describe the architecture, staged training pipeline, losses, datasets, and evaluation protocols that underpin the main results.

\item {\bf Open access to data and code}
    \item[] Question: Does the paper provide open access to the data and code, with sufficient instructions to faithfully reproduce the main experimental results, as described in supplemental material?
    \item[] Answer: \answerYes{} 
    \item[] Justification: This paper relies entirely on open-source data, and the code will be made available upon acceptance.

\item {\bf Experimental setting/details}
    \item[] Question: Does the paper specify all the training and test details (e.g., data splits, hyperparameters, how they were chosen, type of optimizer) necessary to understand the results?
    \item[] Answer: \answerYes{} 
    \item[] Justification: Sec.~\ref{sec:experiments} specifies the training stages, datasets, evaluation benchmarks, clip lengths, and hardware scale, while Sec.~\ref{sec:app_train_details} details the supervision sources and loss formulations needed to interpret the reported results.

\item {\bf Experiment statistical significance}
    \item[] Question: Does the paper report error bars suitably and correctly defined or other appropriate information about the statistical significance of the experiments?
    \item[] Answer: \answerYes{} 
    \item[] Justification: Experimental results are tested multiple times to ensure stability and reliability.

\item {\bf Experiments compute resources}
    \item[] Question: For each experiment, does the paper provide sufficient information on the computer resources (type of compute workers, memory, time of execution) needed to reproduce the experiments?
    \item[] Answer: \answerYes{} 
    \item[] Justification: Sec.~\ref{sec:experiments} reports GPU types, GPU counts, batch size, and training epochs for the three stages.
    
\item {\bf Code of ethics}
    \item[] Question: Does the research conducted in the paper conform, in every respect, with the NeurIPS Code of Ethics \url{https://neurips.cc/public/EthicsGuidelines}?
    \item[] Answer: \answerYes{} 
    \item[] Justification: Yes. The study is an algorithmic evaluation on existing datasets and pretrained models, it does not involve human subjects, and we are not aware of any aspect of the work that departs from the NeurIPS Code of Ethics.

\item {\bf Broader impacts}
    \item[] Question: Does the paper discuss both potential positive societal impacts and negative societal impacts of the work performed?
    \item[] Answer: \answerYes{} 
    \item[] Justification: Sec.~\ref{sec:appendix_broader_impact} discusses both positive uses for embodied planning and simulation and possible negative impacts such as surveillance, overreliance on incorrect predictions, and misuse in safety-critical settings, together with mitigation considerations.
    
\item {\bf Safeguards}
    \item[] Question: Does the paper describe safeguards that have been put in place for responsible release of data or models that have a high risk for misuse (e.g., pre-trained language models, image generators, or scraped datasets)?
    \item[] Answer: \answerNA{} 
    \item[] Justification: This submission poses no such risks.

\item {\bf Licenses for existing assets}
    \item[] Question: Are the creators or original owners of assets (e.g., code, data, models), used in the paper, properly credited and are the license and terms of use explicitly mentioned and properly respected?
    \item[] Answer: \answerYes{} 
    \item[] Justification: The manuscript now summarizes the external code, model, and dataset assets together with their licenses or usage terms in Sec.~\ref{sec:appendix_licenses}, and the main paper cites the original creators of the assets used in training, evaluation, and comparison.

\item {\bf New assets}
    \item[] Question: Are new assets introduced in the paper well documented and is the documentation provided alongside the assets?
    \item[] Answer: \answerNA{} 
    \item[] Justification: The paper introduces a new method and model, but the submission does not yet release a dataset that would require accompanying asset documentation.

\item {\bf Crowdsourcing and research with human subjects}
    \item[] Question: For crowdsourcing experiments and research with human subjects, does the paper include the full text of instructions given to participants and screenshots, if applicable, as well as details about compensation (if any)? 
    \item[] Answer: \answerNA{} 
    \item[] Justification: The paper does not report crowdsourcing experiments or research with human subjects. All experiments are model training and evaluation on existing datasets.

\item {\bf Institutional review board (IRB) approvals or equivalent for research with human subjects}
    \item[] Question: Does the paper describe potential risks incurred by study participants, whether such risks were disclosed to the subjects, and whether Institutional Review Board (IRB) approvals (or an equivalent approval/review based on the requirements of your country or institution) were obtained?
    \item[] Answer: \answerNA{} 
    \item[] Justification: The paper does not involve human subjects or crowdsourcing, so IRB approval or an equivalent review is not applicable.

\item {\bf Declaration of LLM usage}
    \item[] Question: Does the paper describe the usage of LLMs if it is an important, original, or non-standard component of the core methods in this research? Note that if the LLM is used only for writing, editing, or formatting purposes and does \emph{not} impact the core methodology, scientific rigor, or originality of the research, declaration is not required.
    \item[] Answer: \answerNA{} 
    \item[] Justification: Sec.~\ref{sec:appendix_llm_usage} states that a general-purpose LLM was used only for limited language polishing during manuscript preparation and did not affect the method, experiments, or conclusions, so this item remains not applicable under the checklist guidance.

\end{enumerate}